\documentclass[hidelinks]{article}

\usepackage{PRIMEarxiv}
\usepackage[utf8]{inputenc}
\usepackage[T1]{fontenc}
\usepackage{lmodern}
\usepackage{hyperref}
\usepackage{url}
\usepackage{booktabs}
\usepackage{amsfonts}
\usepackage{amsmath}
\usepackage{microtype}
\usepackage[super,sort&compress]{natbib}
\setcitestyle{super,comma}
\usepackage{graphicx}
\usepackage{placeins}
\usepackage{longtable}
\usepackage{orcidlink}
\usepackage{amsmath}
\usepackage{amssymb}
\usepackage{array}
\usepackage{caption}
\usepackage{xcolor}
\usepackage[most]{tcolorbox}
\usepackage{placeins}
\usepackage{needspace}
\usepackage{algorithm}
\usepackage{algpseudocode}
\usepackage{amsmath}
\usepackage{listings}
\graphicspath{{figures/}}
\usepackage{subcaption}

\newcommand{\code}[1]{\url{#1}}

\DeclareCaptionFont{compact}{\scriptsize\linespread{0.93}\selectfont}
\hypersetup{colorlinks=true,citecolor=blue,linkcolor=blue,urlcolor=blue}
\definecolor{swAgentA}{HTML}{0E806B}
\definecolor{swAgentABack}{HTML}{EAF5F1}
\definecolor{swAgentB}{HTML}{3C78A8}
\definecolor{swAgentBBack}{HTML}{EDF4F9}
\definecolor{swAction}{HTML}{B8702D}
\definecolor{swActionBack}{HTML}{FAF1E7}
\definecolor{swWorld}{HTML}{5D6B68}
\definecolor{swWorldBack}{HTML}{F1F4F3}
\definecolor{swInterpret}{HTML}{8A6A20}
\definecolor{swInterpretBack}{HTML}{FBF7E9}
\definecolor{swCodeBack}{HTML}{F7F8F8}

\newenvironment{swchatleft}[1]{%
  \begin{flushleft}
  \begin{tcolorbox}[
    enhanced,
    width=0.88\linewidth,
    colback=swAgentABack,
    colframe=swAgentA,
    colbacktitle=swAgentABack,
    coltitle=black,
    fonttitle=\sffamily\scriptsize,
    title={#1},
    boxrule=0.55pt,
    arc=2.2mm,
    boxsep=1.2mm,
    left=1.6mm,
    right=1.6mm,
    top=1.0mm,
    bottom=1.0mm,
    before skip=3pt,
    after skip=1pt,
    halign=left]
  \footnotesize
}{%
  \end{tcolorbox}
  \end{flushleft}
}

\newenvironment{swchatright}[1]{%
  \begin{flushright}
  \begin{tcolorbox}[
    enhanced,
    width=0.88\linewidth,
    colback=swAgentBBack,
    colframe=swAgentB,
    colbacktitle=swAgentBBack,
    coltitle=black,
    fonttitle=\sffamily\scriptsize,
    title={#1},
    boxrule=0.55pt,
    arc=2.2mm,
    boxsep=1.2mm,
    left=1.6mm,
    right=1.6mm,
    top=1.0mm,
    bottom=1.0mm,
    before skip=3pt,
    after skip=1pt,
    halign=left]
  \footnotesize
}{%
  \end{tcolorbox}
  \end{flushright}
}

\newenvironment{swchataction}[1]{%
  \begin{flushright}
  \begin{tcolorbox}[
    enhanced,
    width=0.88\linewidth,
    colback=swActionBack,
    colframe=swAction,
    colbacktitle=swActionBack,
    coltitle=black,
    fonttitle=\sffamily\scriptsize,
    title={#1},
    boxrule=0.55pt,
    arc=2.2mm,
    boxsep=1.2mm,
    left=1.6mm,
    right=1.6mm,
    top=1.0mm,
    bottom=1.0mm,
    before skip=3pt,
    after skip=1pt,
    halign=left]
  \footnotesize
}{%
  \end{tcolorbox}
  \end{flushright}
}

\newenvironment{swworldrecord}[1]{%
  \begin{center}
  \begin{tcolorbox}[
    enhanced,
    width=0.96\linewidth,
    colback=swWorldBack,
    colframe=swWorld,
    colbacktitle=swWorldBack,
    coltitle=black,
    fonttitle=\sffamily\scriptsize,
    title={SIMULATOR RECORD - #1},
    boxrule=0.45pt,
    arc=0.6mm,
    boxsep=1.0mm,
    left=1.5mm,
    right=1.5mm,
    top=0.8mm,
    bottom=0.8mm,
    before skip=2pt,
    after skip=2pt,
    halign=left]
  \scriptsize
}{%
  \end{tcolorbox}
  \end{center}
}

\newenvironment{swinterpretation}{%
  \begin{tcolorbox}[
    enhanced,
    colback=swInterpretBack,
    colframe=swInterpret,
    boxrule=0.45pt,
    arc=0.8mm,
    boxsep=1.0mm,
    left=1.5mm,
    right=1.5mm,
    top=1.0mm,
    bottom=1.0mm,
    before skip=4pt,
    after skip=7pt]
  \footnotesize\sffamily Interpretation: \rmfamily
}{%
  \end{tcolorbox}
}

\lstdefinestyle{swjson}{
  basicstyle=\ttfamily\scriptsize,
  backgroundcolor=\color{swCodeBack},
  frame=single,
  rulecolor=\color{swWorld},
  framerule=0.4pt,
  breaklines=true,
  breakatwhitespace=false,
  columns=fullflexible,
  showstringspaces=false,
  keepspaces=true,
  xleftmargin=1.5mm,
  xrightmargin=1.5mm,
  aboveskip=4pt,
  belowskip=6pt
}

\title{SwarmWorld: Stigmergic technological evolution in societies of language-model agents}

\author{
  \orcidlink{0000-0002-0532-9791} Subhadeep Pal\textsuperscript{1,2} \quad 
  \orcidlink{0000-0002-4847-6986} Fiona Y. Wang\textsuperscript{1,3} \quad
  \orcidlink{0000-0002-4173-9659} Markus J. Buehler\textsuperscript{1,2,4,5,\#} \\
  \\
  \textsuperscript{1}Laboratory for Atomistic and Molecular Mechanics (LAMM),\\
  \textsuperscript{2}Department of Civil and Environmental Engineering, \\
  \textsuperscript{3}Department of Biological Engineering, \\
  \textsuperscript{4}Department of Mechanical Engineering, \\
  \textsuperscript{5}Center for Computational Science and Engineering, \\
  Schwarzman College of Computing, \\
  Massachusetts Institute of Technology, Cambridge, MA 02139, USA \\ \\
    \textsuperscript{\#}Corresponding author: 
  \texttt{mbuehler@mit.edu}
}
\date{}

\begin{document}
\maketitle

\begin{abstract}
Collective intelligence can emerge when individuals coordinate through a shared environment, allowing local actions to accumulate into durable social organization. Language-model agents offer a new substrate for this process, yet most multi-agent systems rely on direct conversation, predefined roles, or centralized workflows. It remains unclear whether decentralized agents can build functional technologies and outperform independent search. Here, initially homogeneous LLM agents in SwarmWorld self-organize without assigned roles or recipes into evolving technological societies. Agents explore a spatial environment, process resources, test materials, construct persistent artifacts, and write executable controllers evaluated by a deterministic simulator under unseen disturbances after the agents are removed. SwarmWorld splits cognition from consequence: agents propose architectures and controllers within fixed action and material schemas, while the simulated world determines function. Shared societies develop broader, more resilient technological portfolios than a strong best-of-$N$ isolated-search baseline, although isolated search remains competitive for the strongest artifact. Agents differentiate into exploration, construction, maintenance, and coordination behaviors, transitioning as the world matures. Technologies accumulate through collaborative construction, executable inheritance, and persistent agent-artifact networks, with most reuse beginning through physical observation rather than communication. Explicit cultural mechanisms amplify collaboration and organization, but functional benefits depend on outcome and timescale. Physical stigmergy alone supports capable societies, while interaction drives persistent technological ecologies rather than universally superior individual inventions.
\end{abstract}

\keywords{language-model agents \and multi-agent systems \and stigmergy \and cumulative culture \and scientific discovery \and bio-inspired materials}

\section{Introduction}
\label{sec:introduction}
 
Collective behavior allows groups to achieve outcomes beyond the reach of isolated individuals, from ant foraging and honeybee nest-site selection to quorum-guided decisions in fish schools \cite{bonabeau_self-organization_1997,sumpter_principles_2006,goss_self-organized_1989,seeley_quorum_2004,ward_quorum_2008}. Such organization need not require a central coordinator. Local feedback can amplify useful behavior, task allocation can adapt to changing needs, and persistent environmental modifications can coordinate later activity through stigmergy \cite{grasse1959,theraulaz_brief_1999,gordon_organization_1996}. These mechanisms do more than aggregate simultaneous actions: they allow one individual's activity to alter the information and opportunities available to others. They therefore motivate a central question for artificial collectives: can decentralized agents do more than search in parallel by constructing a shared, cumulative substrate for future action?

Two computational lineages frame this question. Swarm-intelligence methods translate decentralized interaction into search and control: ant colony optimization reinforces useful paths, particle swarm optimization shares individual and population experience, and artificial bee colony methods balance exploration and exploitation \cite{dorigo_ant_1996,kennedy_particle_1995,karaboga_powerful_2007}. The same principles extend to swarm robotics and materials search, including particle-swarm crystal-structure prediction in CALYPSO \cite{brambilla_swarm_2013,wang_crystal_2010}. A parallel lineage made the evolving world itself the object of computation. Cybernetics and system dynamics emphasized feedback within stateful systems \cite{wiener1948cybernetics,forrester1969urban}, while cellular automata showed how repeated local rules generate persistent global structure \cite{vonneumann1966automata,gardner1970life}. Interactive simulations such as \textit{Hamurabi} and \textit{SimCity} made intervention in evolving systems an object of experimentation, whereas Boids and Sugarscape populated simulated environments with autonomous agents whose local behavior produced collective organization \cite{ahl1978basic,wright1989simcity,reynolds1987flocks,epstein1996growing}. Recent transformer models of cellular dynamics further connect learned sequence modeling to this local-rule tradition \cite{berkovich_lifegpt_2025,berkovich_automatagpt_2026}. These lineages supply complementary ingredients---distributed coordination on one hand and persistent, consequential worlds on the other---but generally retain fixed representations, hand-designed policies, or predefined objectives. LLM-guided robotic swarms begin to relax those restrictions by allowing agents to reason and communicate within collective tasks \cite{strobel_llm2swarm_2024,jimenez-romero_multi-agent_2025,jiang_exploring_2025}.

Multi-agent reinforcement learning made some forms of coordination, competition, and tool use learnable rather than prescribed \cite{lowe2017multiagent,baker2020emergent}; language-model agents add general reasoning, memory, communication, and program synthesis to this foundation. Generative Agents showed how memory, reflection, and planning can support emergent social behavior in a simulated town, while Project Sid reported specialization, collective rule formation, and cultural transmission at larger scale \cite{park2023generative,altera2024projectsid}. GovSim isolates cooperation over shared resources, and AgentSociety extends generative-agent simulation beyond 10,000 agents \cite{piatti2024govsim,piao2025agentsociety}. TerraLingua moves closer to a persistent LLM ecology: agents create and revise textual artifacts that outlive their authors and acquire branching cultural lineages \cite{paolo_terralingua_2026}. Other systems make capabilities themselves cumulative. Voyager stores reusable executable skills, GenSwarm generates coordinated multi-robot policies, and DiscoveryWorld grounds hypothesis formation and experimentation in interactive scientific tasks \cite{wang2023voyager,ji_genswarm_2026,jansen2024discoveryworld}. Yet scale, memory, and communication do not make collective advantage automatic. Local perception still challenges coordination, while rapid consensus can suppress exploration even as individual reasoning improves \cite{ruan_benchmarking_2025,zomer_unraveling_2026}.

A parallel body of work applies LLMs and multi-agent systems to scientific discovery by coupling reasoning to knowledge representations, code, simulation, and analysis tools \cite{buehler_mechgpt_2024, buehler_accelerating_2024, buehler_generative_2024}. ProtAgents assigns specialized agents to protein design, structure prediction, simulation, and retrieval; SciAgents links specialized reasoning through knowledge graphs; and AtomAgents connects multimodal agents to atomistic simulation for alloy design \cite{ghafarollahi_protagents_2024,ghafarollahi_sciagents_2025,ghafarollahi_automating_2025}. Sparks and SparksMatter organize hypothesis generation, computational testing, and iterative refinement, whereas CASCADE emphasizes the acquisition and exchange of reusable scientific skills \cite{ghafarollahi_sparks_2025,ghafarollahi_autonomous_2026,huang_cascade_2025}. Decentralized protein-design swarms, PharmaSwarm, and MusicSwarm explore collective organization in sequence search, drug discovery, and long-horizon creative production \cite{wang_swarms_2025,song_llm_2025,buehler_musicswarm_2026}; virtual laboratories and ScienceClaw $\times$ Infinite extend the unit of analysis toward communities that compete, transmit research lineages, or build on persistent scientific artifacts \cite{loo_agentic_2026,wang_autonomous_2026}. These systems establish substantial autonomy; the protein-design swarm additionally reports experimental validation of generated designs. Most nevertheless prescribe roles, workflows, tool access, evaluation structures, or bounded interaction patterns rather than asking what technological organization emerges among initially equivalent agents.

What remains missing is a controlled test that combines four properties: initially equivalent agents, a shared world that retains their modifications, executable technologies whose function is evaluated independently of agent claims, and a matched isolated-search baseline that measures whether interaction changes capability rather than merely increasing the number of samples. TerraLingua provides a close comparison for persistent cultural accumulation, but its principal artifacts are textual and its endpoints concern ecological survival and interpreted artifact properties. The distinction is consequential: persistent information can demonstrate cultural accumulation, whereas persistent technology can also be subjected to an external functional assay. Technological descent should therefore have two independently observable consequences---later agents can inherit and modify earlier constructions, and the resulting artifacts can succeed or fail under dynamics that do not depend on an LLM's description of their value. Scientific-agent systems provide functional evaluation, but commonly within designed decompositions or workflows. The unresolved question is not simply whether LLM agents can communicate, specialize, or create artifacts, but whether their interaction can produce a functionally stronger technological ecology than the same computational population can discover independently.

\begin{figure}[ht]
    \centering
    \includegraphics[width=\textwidth]{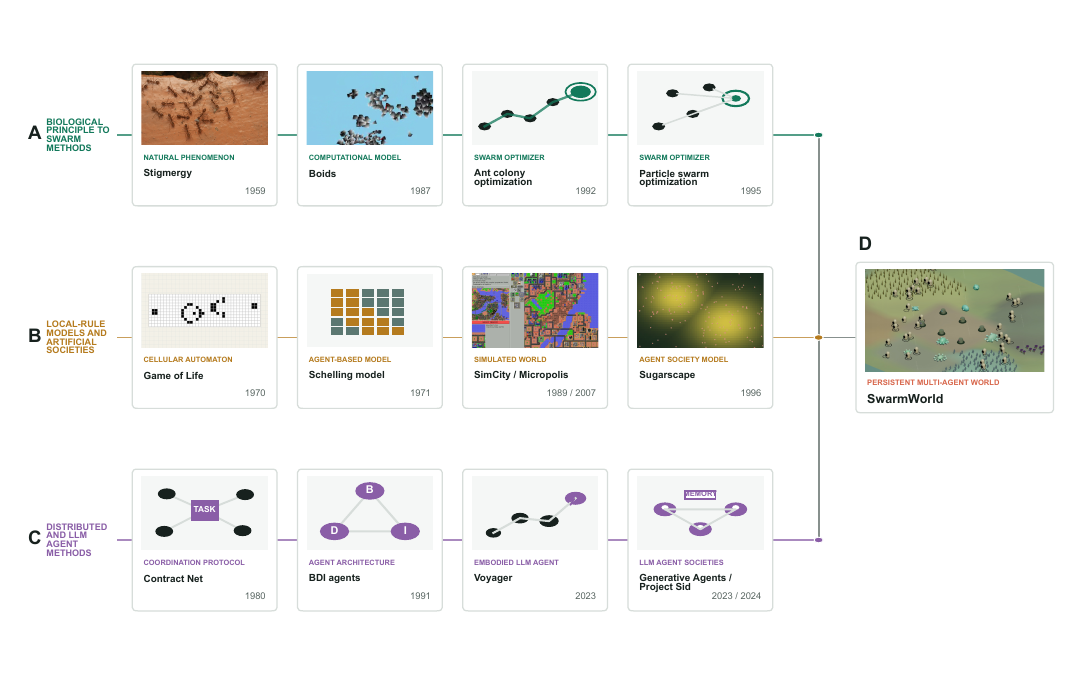}
    \caption{Conceptual lineages leading to SwarmWorld; connectors indicate conceptual convergence rather than direct descent between every adjacent item. (A) Biological collectives and swarm-computing methods contribute decentralized coordination, feedback, and stigmergy. (B) Local-rule models and artificial societies contribute persistent worlds in which local actions alter shared state and generate population-level organization. (C) Distributed-agent and LLM systems contribute explicit agent state, memory-guided planning, embodiment, communication, and reusable executable skills. (D) SwarmWorld combines these elements in a materially constrained multi-agent world where agents build persistent artifacts and inherit or modify their programs. The resulting technological ecology can be evaluated after the agents are removed. \textit{Image credits.} Ant photograph by Kiran Vati K, \href{https://commons.wikimedia.org/wiki/File:Ants_on_the_hunt_for_food.jpg}{CC BY 4.0}, cropped and color-adjusted. Boids still from a video by Andrew Magill, \href{https://commons.wikimedia.org/wiki/File:Boids_OpenGL_example.webm}{CC BY 2.0}, with a frame extracted, cropped, and color-adjusted. Game of Life glider-gun diagram by Bryan.burgers, \href{https://commons.wikimedia.org/wiki/File:Game_of_life_glider_gun.svg}{public domain}, placed on a grid background. Micropolis screenshot based on the original software by Will Wright and Maxis Software/Electronic Arts, the Unix port by Don Hopkins and DUX Software, screenshot by Tomhannen, and edit by bayo; \href{https://commons.wikimedia.org/wiki/File:Micropolis_-_big_city.png}{GPLv3 or later with Section 7 additional terms}, cropped and color-adjusted. The Sugarscape view, method schematics, and SwarmWorld rendering were created by the authors.}
    \label{fig:historical-lineage-tree}
\end{figure}

To address this gap, we introduce SwarmWorld, a controlled environment that combines initially homogeneous LLM agents with a persistent, materially constrained world (Figure~\ref{fig:historical-lineage-tree}). Without assigned roles, predefined recipes, or a technology catalog, agents explore locally, process source--sink-accounted resources, test materials, construct persistent artifacts, and author executable controllers. This architecture enforces a proposal--consequence separation: ideas, messages, and designs are agent-authored claims, whereas the deterministic simulator alone determines what can be built and what functions. Every technology is spatially situated, reads and acts on local world state, and continues to execute between agent decisions. Because later agents can encounter, inherit, and modify these constructions, successful work becomes part of the environment rather than remaining only in a transcript. Because the agents are removed during held-out evaluation, artifact performance can also be measured independently of its creator's interpretation.

We isolate communication, cross-agent program inheritance, and physical stigmergy through controlled ablations, and compare every society with an endpoint-wise best-of-$N$ envelope of matched isolated agents receiving the same scheduled decision opportunities. This makes swarm advantage falsifiable: interaction must improve capability beyond what the same computational population achieves through independent search. Across populations of 50--200 agents and complementary long-horizon experiments, shared worlds produced broader and more resilient technological portfolios and self-organized differentiation between exploratory and technology-centered behavior. Explicit culture generated collaboration and executable inheritance but did not improve every endpoint, while isolated search could still retain the strongest individual artifact. The contribution is therefore a bounded swarm advantage: interaction chiefly supports the accumulation of a diverse, persistent technological ecology, not universally superior individual inventions.
 
\section{Results and Discussion}

SwarmWorld provides a setting in which collective intelligence can be evaluated through multiple channels including agent behavior, communication and the technologies that a population leaves behind (in a shared world). This design allows us to ask how interaction changes the discovery, accumulation, inheritance, and functional robustness of technology, and whether these effects can be distinguished from the gains expected from parallel independent search. The results trace this process from controlled interaction mechanisms and emergent behavioral differentiation to executable technological lineages, diffusion through persistent artifacts, and the long-horizon organization and robustness of the resulting agent-artifact ecology.

\subsection{Research Design}

To test these questions, we performed two paired studies that provide complementary evidence (for a glossary of key terms, see Section~\ref{glossary_section}). The population-scaling study ran four mechanism-resolved conditions for 800 ticks at $N=50$, 100, and 200, with four matched world seeds per cell and eight held-out disturbance schedules. The long-horizon study followed $N=100$ societies for 3,200 ticks under full culture, no explicit culture, and an endpoint-wise best-of-100 independent-search envelope. Frozen states were evaluated at ticks 400, 800, 1,600, 2,400, and 3,200. The world seed is the unit of replication throughout; agent trajectories, artifacts, time samples, and disturbance schedules are nested observations. With four paired seeds, the analysis emphasizes effect sizes, paired consistency, and mechanisms rather than asymptotic population-level inference.

Notably, the main result is not a universal swarm advantage. Shared physical worlds consistently produced broader and more resilient portfolios than isolated search, and explicit culture generated measurable collaboration, cross-agent code descent, network densification, and behavioral reorganization. However, the isolated envelope could retain the strongest single artifact, no-explicit-culture societies sometimes outperformed full culture, and the apparent benefit of culture depended on both time and endpoint. The figures therefore connect the simulation interface to behavior, recorded technological lineage, held-out function, and structural robustness.

\subsection{A controlled shared world separates physical, cultural, and independent search mechanisms}

\begin{figure}[htbp]
    \centering
    \includegraphics[width=0.97\textwidth,height=0.64\textheight,keepaspectratio]
    {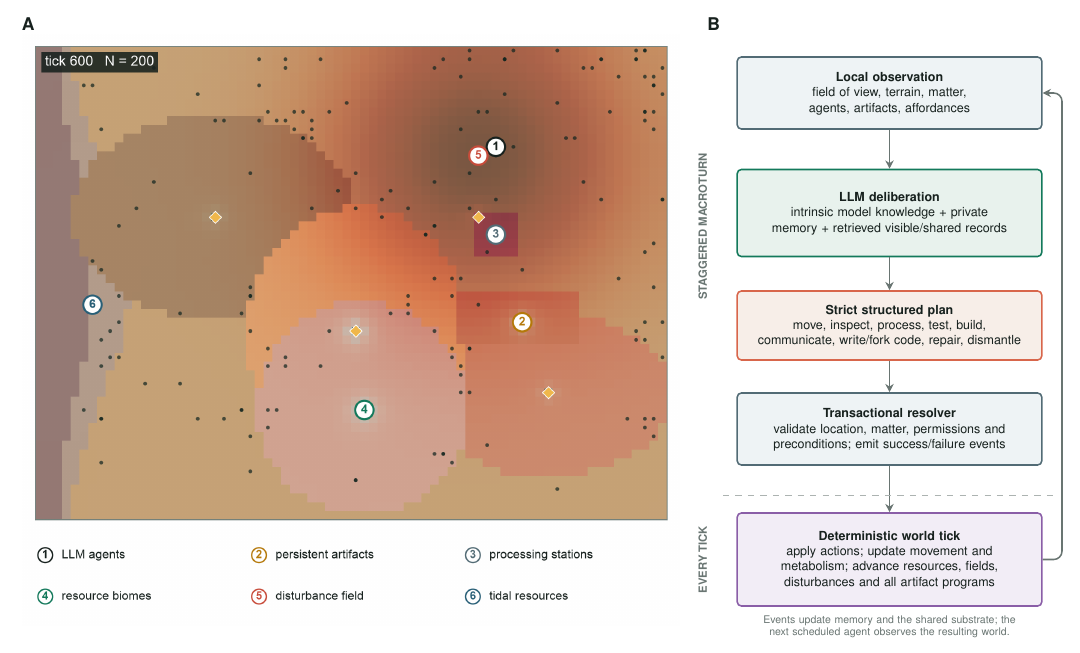}
    \caption{Simulated world and agent-environment algorithm. (A) Authoritative state at tick 600 from a representative $N=200$ society. Numbered callouts identify language-model agents, persistent agent-built artifacts, fixed processing stations, resource biomes, an active disturbance field, and tidal resources. The map is the state on which actions are resolved, not a decorative visualization. Agents can move through the terrain, gather and transform matter, inspect or operate artifacts, and encounter local environmental changes produced by both disturbances and other agents. (B) One staggered macroturn begins with a local observation and retrieved private memory, continues through language-model deliberation and a strict structured plan, and ends with transactional validation of actions, targets, permissions, and material preconditions. The deterministic world tick then advances motion, metabolism, resources, fields, disturbances, and every installed artifact program. Resulting events enter agent memory and the shared substrate before the next scheduled observation. The agents decide what to try, but the world decides what actually happens; durable artifacts let one agent's successful work become part of another agent's future environment.}
    \label{fig:world-algorithm}
\end{figure}

The simulation couples open-ended language-model decisions to a deterministic physical substrate (Figure~\ref{fig:world-algorithm}). A complementary three-dimensional rendering of a representative world (seed 3202) illustrates the spatial separation of resource biomes, processing foundries, agents, and persistent agent-built artifacts (Figure \ref{fig:biofoundry_field}). Each agent receives only a local observation and retrieved memory, then emits a schema-constrained plan whose individual actions are checked and resolved transactionally. Movement, metabolism, fields, material transport, disturbances, and installed artifact programs continue on every world tick, including ticks without a model call. This separation is important: the model supplies design and action choices, while the simulator determines whether those choices are legal and what consequences they have. Every SwarmWorld artifact is spatially situated: it occupies coordinates, reads only local sensors, and influences agents and fields only where it stands. Persistent artifacts can therefore become part of the environment encountered by later agents rather than remaining text in a shared transcript.

\begin{figure}[htbp]
    \centering
    \includegraphics[width=0.97\textwidth,height=0.64\textheight,keepaspectratio]
    {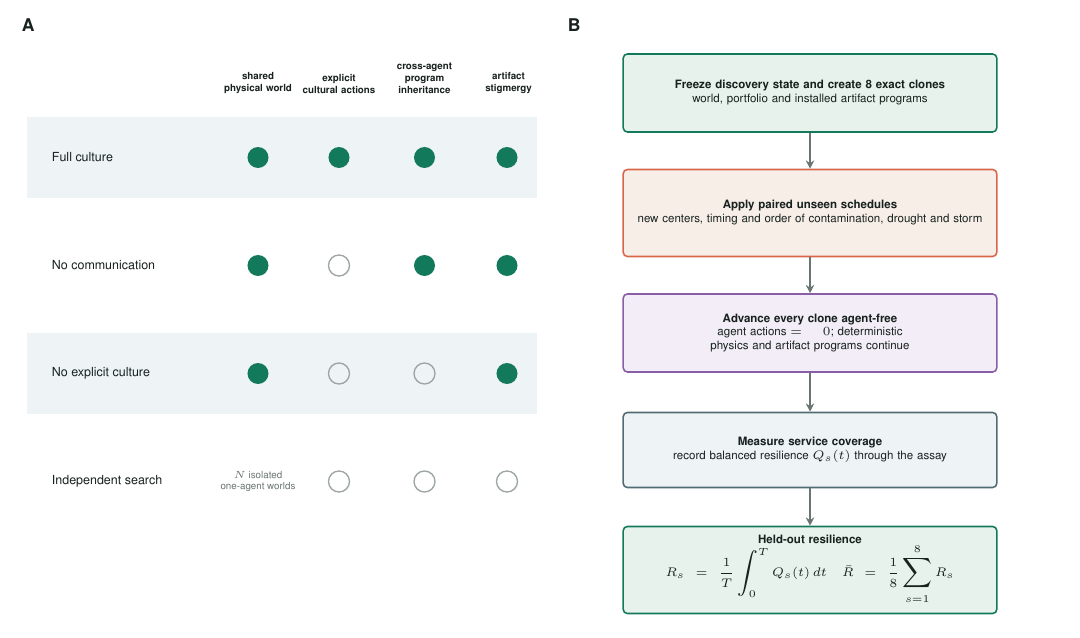}
    \caption[Mechanism-resolved conditions and held-out assay]{Mechanism-resolved conditions and the agent-free held-out resilience assay. (A) Rows define the four interventions and columns identify the mechanisms available in each. A filled circle means the mechanism is present. Full culture combines a shared physical world, explicit cultural actions, cross-agent program inheritance, and artifact stigmergy. No communication removes nearby messages, public records, teaching, trading, task claims, and publication-dependent composition while preserving the shared world and executable inheritance. No explicit culture also removes cross-agent program forking and measured skill inheritance, leaving agents able to coordinate only through persistent artifacts and environmental changes. Independent search consists of $N$ isolated one-agent worlds and therefore has none of the three collective mechanisms. (B) At a discovery checkpoint, the complete state is frozen into eight exact clones. Each clone receives a paired unseen schedule of contamination, drought, and storm, with new centers, timings, and orderings. Agents take no actions during evaluation, but the world physics and installed artifact programs continue. Balanced service coverage $Q_s(t)$ is integrated over assay time and averaged across schedules to produce held-out resilience. The assay asks whether the technology left behind can protect the habitat under new stresses after its inventors have gone.}
    \label{fig:experimental-assay}
\end{figure}

The four experimental conditions remove mechanisms without changing the scientific task (Figure~\ref{fig:experimental-assay}). Full culture provides a shared world, explicit messages and records, cross-agent executable-program inheritance, and artifact-mediated stigmergy. No communication removes direct communication and publication-dependent composition but retains shared artifacts and program inheritance. No explicit culture additionally removes cross-agent program forking and measured skill inheritance, leaving only physical stigmergy. Independent search replaces the shared society with $N$ isolated one-agent worlds and reports an endpoint-wise best-of-$N$ envelope. This is a deliberately strong control because its winner can differ across endpoints and checkpoints. The held-out assay further separates discovery from evaluation: agents are removed, the frozen technological state is cloned into eight paired unseen disturbances, and only deterministic physics and installed programs continue.

\begin{figure}[htbp]
    \centering
    \includegraphics[width=0.97\textwidth,height=0.64\textheight,keepaspectratio]
    {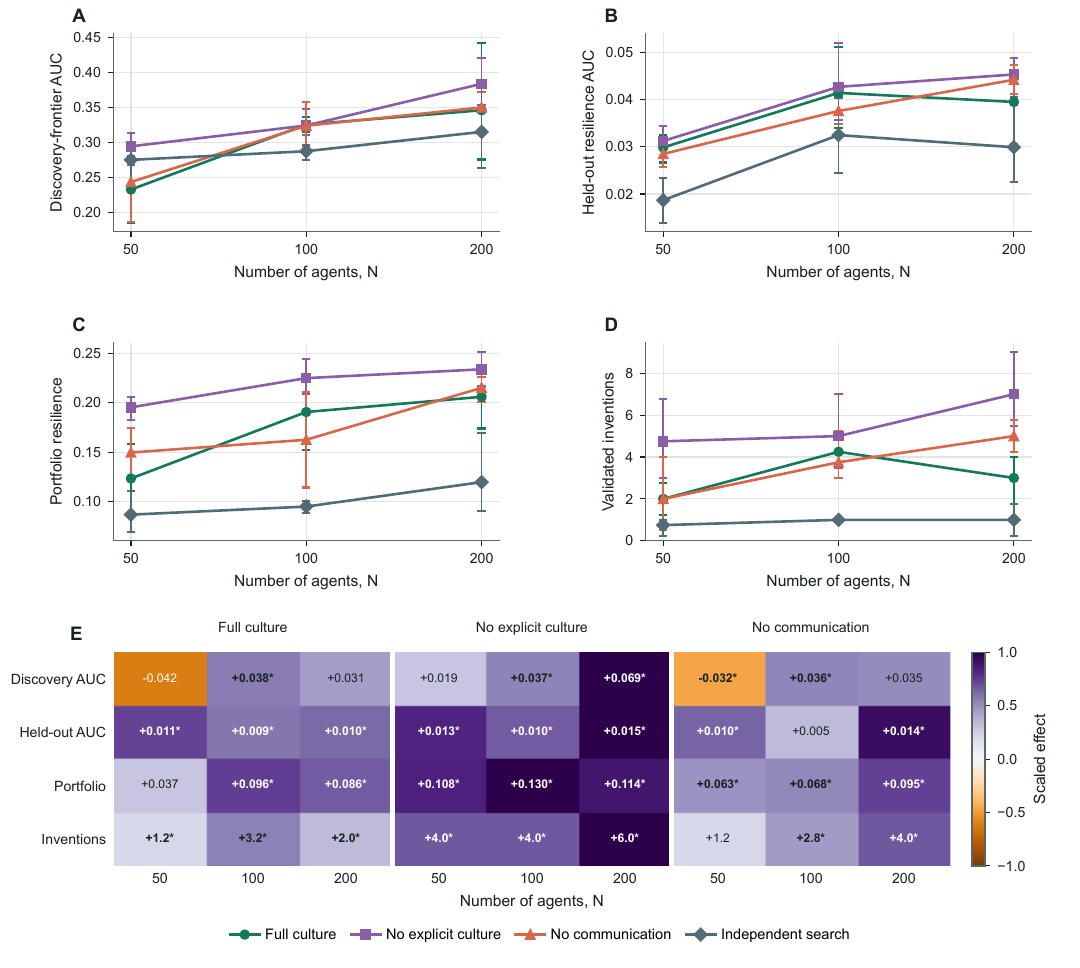}
    \caption[Population scaling and collective effects]{Population scaling and mechanism-resolved collective effects over 800 discovery ticks. Points and heavy lines in panels A-D are means across four matched world seeds; error bars are paired seed-bootstrap 95\% intervals. (A) Discovery-frontier AUC integrates the running best measured artifact performance, rewarding discoveries that occur early and remain on the frontier. (B) Held-out resilience AUC measures agent-free service coverage under eight unseen disturbance schedules. (C) Portfolio resilience quantifies the breadth and redundancy of functional service across the final artifact collection. (D) Validated inventions count technologies that pass the simulator's objective validation criteria. (E) Each cell is a shared-world condition minus the paired endpoint-wise independent-search envelope. Printed values are raw paired effects, asterisks mark intervals excluding zero, and color is normalized within each outcome row because the four outcomes have different units. The shared world yields the most consistent gains in held-out and portfolio resilience and in invention count, whereas discovery AUC depends on condition and scale. Societies become better at building useful collections of technologies, but more communication does not automatically make them better at finding the single fastest early winner.}
    \label{fig:collective-results}
\end{figure}

Population scaling produced a mechanism-dependent rather than monotonic result (Figure~\ref{fig:collective-results}). Discovery-frontier AUC generally increased with population, but the condition ranking changed with $N$. At $N=50$, full culture and no communication trailed the independent envelope on discovery AUC; at $N=100$, all three shared-world conditions exceeded it; at $N=200$, no explicit culture produced the largest paired discovery gain, $+0.069$. The more robust endpoints were more consistent. Held-out resilience exceeded the isolated envelope for nearly every shared-world cell, portfolio resilience showed positive paired effects throughout, and validated inventions increased strongly, reaching a mean paired gain of six inventions for no explicit culture at $N=200$. We reserve the term ``invention'' for artifacts that clear the full validation gate of tested materials, a complete design, an installed agent-authored program, threshold performance, and behavioral novelty (Section~\ref{glossary_section}); an invention is therefore a validated, functioning, situated technology rather than a proposal alone.
Thus, access to a shared physical substrate was broadly beneficial, while adding explicit cultural machinery did not uniformly improve short-horizon performance.

\subsection{Unassigned agents differentiate and create executable technological culture}

\begin{figure}[htbp]
    \centering
    \includegraphics[width=0.97\textwidth,height=0.64\textheight,keepaspectratio]
    {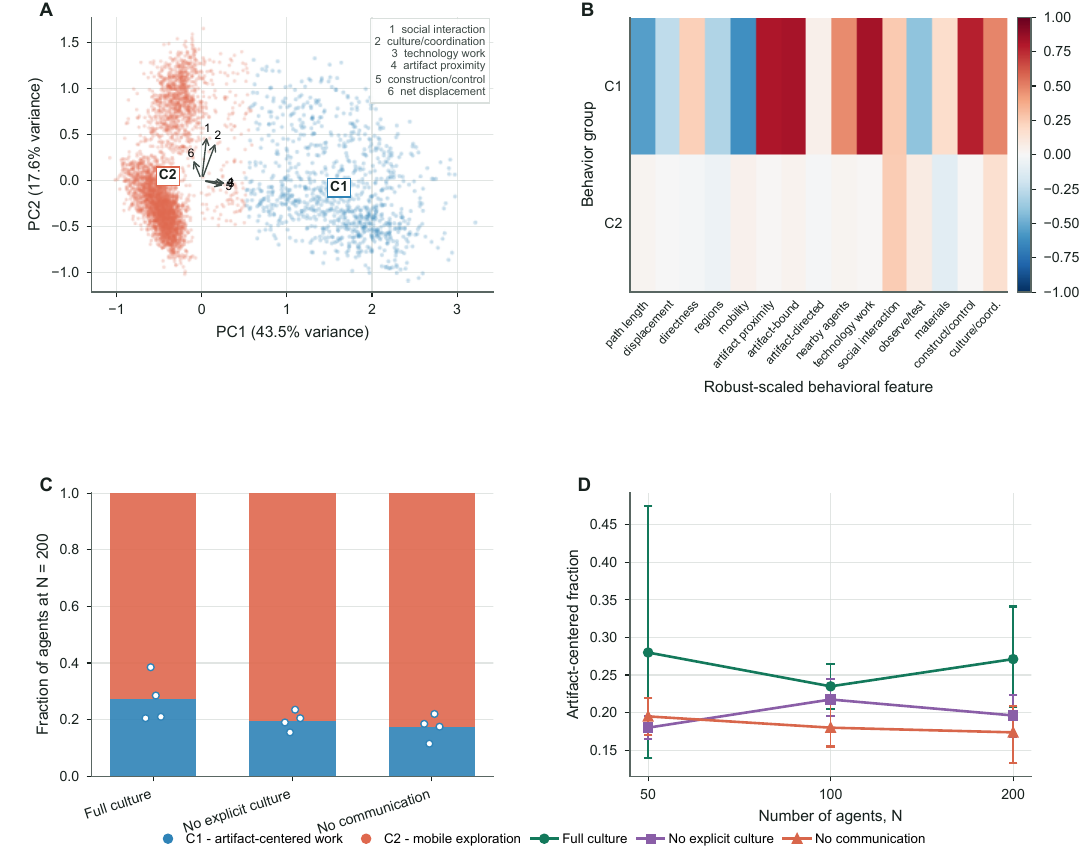}
    \caption[Data-driven behavioral phenotypes]{Data-driven behavioral phenotypes discovered in the 800-tick scaling study. Each point represents one complete agent trajectory, and the two-group k-means classification is learned from 15 robust-scaled behavioral features without access to condition, population, seed, identity, or an assigned role. (A) PCA gives an auditable linear projection. The numbered arrows are the six largest loading vectors and point toward increasing social interaction, cultural coordination, technology work, artifact proximity, construction/control, and net displacement. Clustering is performed in the full feature space rather than in this two-dimensional display. (B) Cluster-average standardized signatures identify C1 as artifact-centered work and C2 as mobile exploration; red indicates above-corpus values and blue indicates below-corpus values. (C) Stacked bars show the mean $N=200$ composition, with four seed fractions overlaid. (D) The artifact-centered fraction is followed across $N=50$, 100, and 200 for each shared-world condition. Initially identical agents divide into a smaller group that stays near and works on technology and a larger group that continues exploring, and explicit culture shifts more agents toward the technology-centered mode.}
    \label{fig:emergent-organization}
\end{figure}

We find that behavioral differentiation arose without role prompts or condition labels (Figure~\ref{fig:emergent-organization}). A two-cluster model fit to 15 robust-scaled features separated artifact-centered work from mobile exploration. The first phenotype combined artifact proximity, artifact-bound motion, construction, control, and cultural coordination; the second retained broader movement and lower artifact engagement. At $N=200$, the artifact-centered fraction averaged approximately 27\% under full culture, 20\% without explicit culture, and 17\% without communication. These fractions are not task assignments. They are post hoc descriptions of complete trajectories, and the between-condition comparison uses seed-level proportions rather than treating agents as independent replicates.

\begin{figure}[htbp]
    \centering
    \includegraphics[width=0.97\textwidth,height=0.64\textheight,keepaspectratio]
    {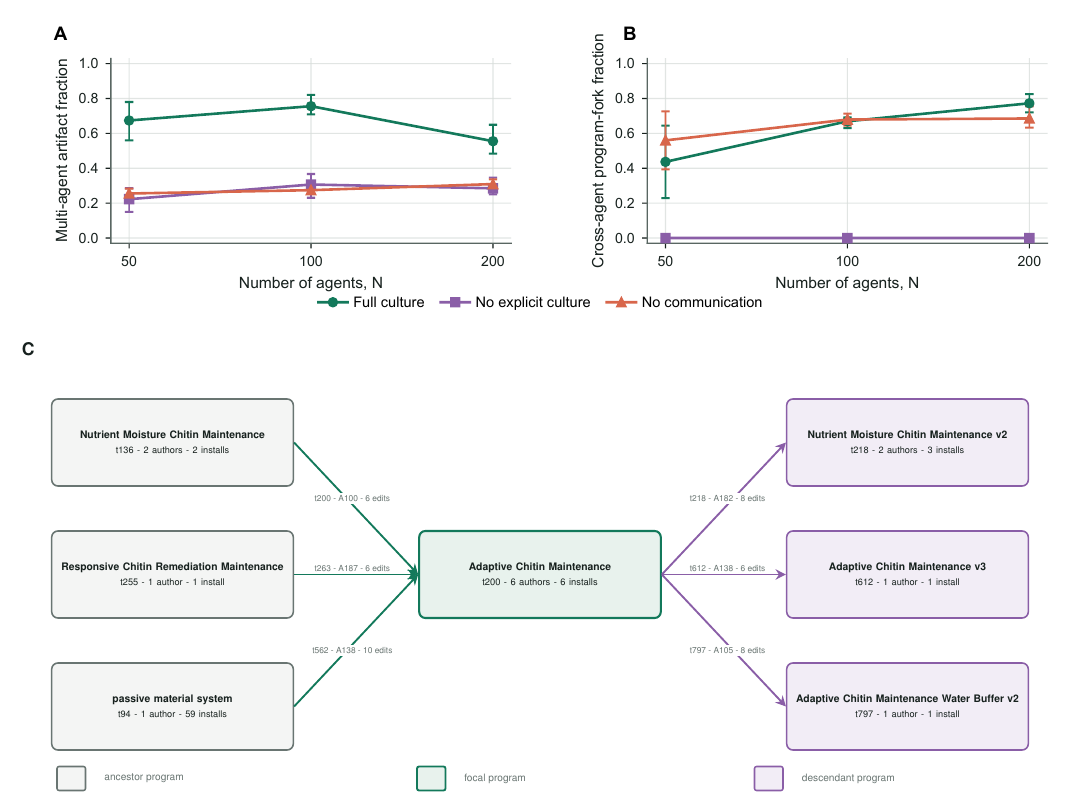}
    \caption[Executable technological culture]{Executable technological culture. Seed means and 95\% seed-bootstrap intervals are shown across population size. (A) Fraction of constructed artifacts with recorded contributions from multiple agents. Full culture produces the largest collaborative fraction at every population, while appreciable collaboration without explicit culture demonstrates that agents can still meet around and modify persistent objects. (B) Fraction of eligible program forks whose child author differs from the parent author. Cross-agent descent grows with population under full culture and remains possible without direct communication because executable inheritance is still available in that intervention; it is zero by construction when explicit culture and cross-agent inheritance are disabled. (C) Focused content-addressed program lineage centered on \texttt{Adaptive Chitin Maintenance}. Gray boxes are ancestor programs, the green box is the focal program, and purple boxes are descendants. Each node records first tick, number of authors, and installations; each arrow records fork tick, author, and instruction-level edit count. Agents did not merely discuss ideas: they reused, edited, installed, and propagated one another's executable environmental controllers, producing attributable technological descent.}
    \label{fig:executable-culture}
\end{figure}

The cultural record also entered physical construction and executable code (Figure~\ref{fig:executable-culture}). Under full culture, 67\%, 76\%, and 56\% of artifacts at $N=50$, 100, and 200, respectively, recorded contributions from more than one agent, substantially exceeding the corresponding ablations. Cross-agent program forking remained common whenever program inheritance was available, including the no-communication condition; it was exactly absent when the mechanism was disabled. The focused lineage around \texttt{Adaptive Chitin Maintenance} shows several independently authored ancestor programs converging on a six-author focal program, followed by multiple descendants with recorded instruction edits. These are content-addressed forks and installations, not semantic similarity between descriptions.

\begin{figure}[htbp]
    \centering
    \includegraphics[width=0.97\textwidth,height=0.9\textheight,keepaspectratio]
    {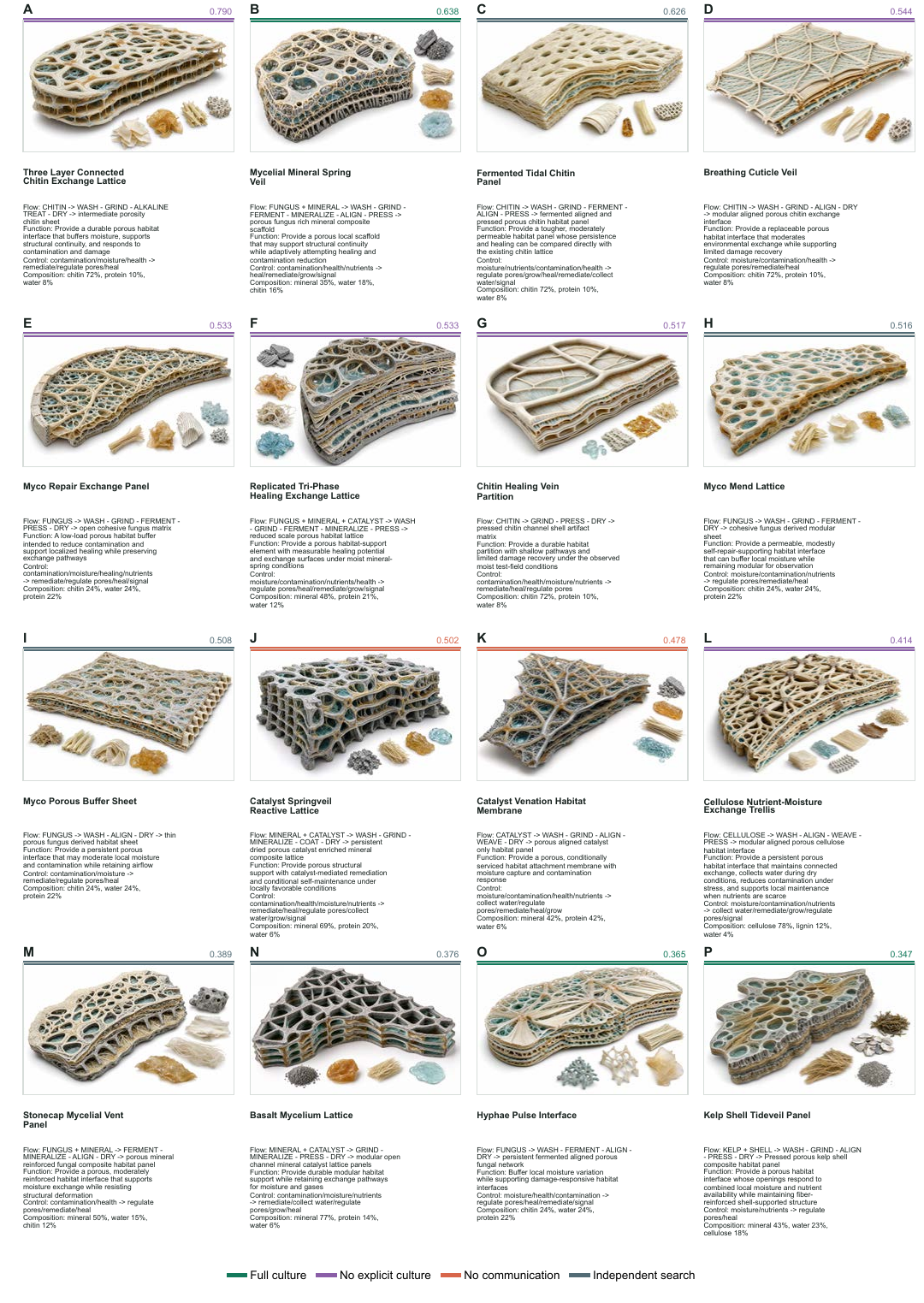}
    \caption[Sixteen agent-invented technologies]{Sixteen ranked, deduplicated agent-invented technologies from the 800-tick study. The colored rule above each panel identifies its experimental condition and the number at upper right is lifetime-peak simulator performance. Panels A-D show the Three Layer Connected Chitin Exchange Lattice, Mycelial Mineral Spring Veil, Fermented Tidal Chitin Panel, and Breathing Cuticle Veil. Panels E-H show the Myco Repair Exchange Panel, Replicated Tri-Phase Healing Exchange Lattice, Chitin Healing Vein Partition, and Myco Mend Lattice. Panels I-L show the Myco Porous Buffer Sheet, Catalyst Springveil Reactive Lattice, Catalyst Venation Habitat Membrane, and Cellulose Nutrient-Moisture Exchange Trellis. Panels M-P show the Stonecap Mycelial Vent Panel, Basalt Mycelium Lattice, Hyphae Pulse Interface, and Kelp Shell Tideveil Panel. Each portrait combines a complete form, partial cutaway, and dominant component specimens inferred from the recorded architecture, geometry, composition, process sequence, and controller. Text below each image reports the input-process-output flow, agent-authored functional claim, sensor-to-actuator control signature, and leading material fractions. The society explored multiple technological families rather than converging on one prewritten object; the images communicate the proposed mechanisms, while the quantitative score comes only from the simulator. See Section~\ref{sec:performance-ranked-tech-selection} for details on how these representative technologies were selected.}
    \label{fig:technology-gallery}
\end{figure}

The constructed portfolio was diverse in recorded feedstocks and simulator-defined functions (Figure~\ref{fig:technology-gallery}). The 16 displayed technologies are ranked and deduplicated representatives drawn across all four conditions rather than templates supplied in the prompt. Their recorded lifetime-peak simulator scores range from 0.790 to 0.347 and span chitin exchange lattices, mycelial mineral veils, tidal panels, cuticle-like membranes, repair structures, catalyst networks, cellulose trellises, and kelp-shell composites. The accompanying mechanism renderings are generated from the recorded geometry, composition, recipe, and controller. They therefore visualize the agents' proposed technological identity, but they are not literal meshes from the simulator and should not be interpreted as experimental validation of real material performance.

\subsection{Movement, world modification, and provenance connect search to invention}

\begin{figure}[htbp]
    \centering
    \includegraphics[width=0.97\textwidth,height=0.64\textheight,keepaspectratio]
    {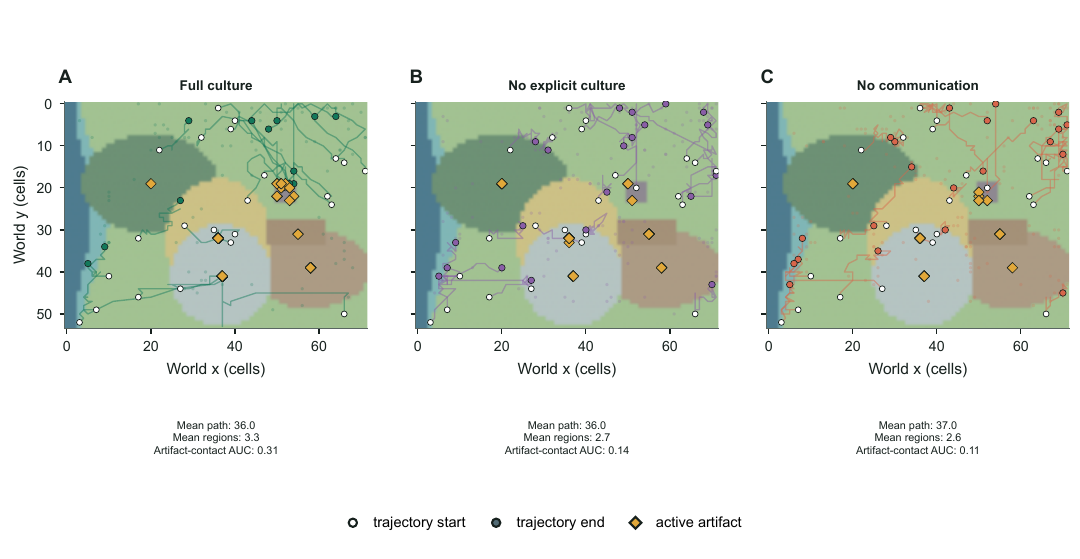}
    \caption[Representative movement trajectories]{Representative $N=200$ movement trajectories from matched seed-3202 societies. For readability, 24 deterministically selected complete paths are shown in each panel over the identical terrain. White circles mark trajectory starts, colored circles mark endpoints, and gold diamonds mark artifacts active at tick 800. (A) Full culture produces paths that repeatedly intersect the central artifact ecology; all-agent summaries report mean path length 36.0 cells, 3.3 regions visited, and artifact-contact AUC 0.31. (B) Without explicit culture, mean path length remains 36.0 cells, but region coverage falls to 2.7 and artifact-contact AUC to 0.14. (C) Without communication, mean path length is 37.0 cells, region coverage is 2.6, and artifact-contact AUC is 0.11. The inset values use all 200 agents, not only the displayed paths. The agents travel similar total distances, but full culture makes their movement more likely to intersect the shared technological infrastructure.}
    \label{fig:trajectory-atlas}
\end{figure}

Matched spatial trajectories show that the shared-world mechanisms changed where agents worked, not merely how far they traveled (Figure~\ref{fig:trajectory-atlas}). In the representative matched seed-3202 episodes shown, mean path length remained nearly constant at 36 to 37 cells across the three $N=200$ conditions. Artifact-contact AUC was 0.31 under full culture, compared with 0.14 without explicit culture and 0.11 without communication, and full-culture agents visited more regions on average. For this representative seed, the difference therefore reflects how movement was organized around constructed technology, not a trivial increase in total locomotion.

\begin{figure}[htbp]
    \centering
    \includegraphics[width=0.97\textwidth,height=0.64\textheight,keepaspectratio]
    {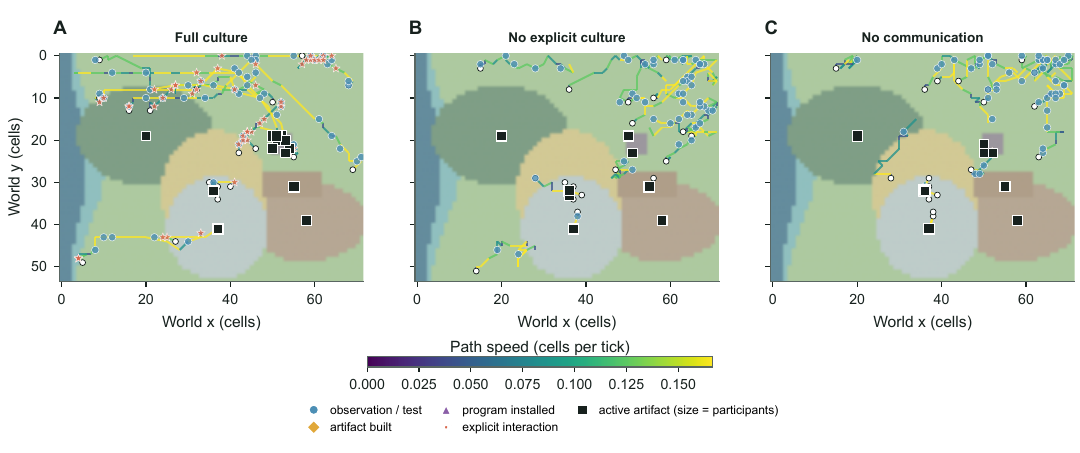}
    \caption[Speed-colored trajectories linked to actions]{Movement linked to scientific work and interaction in representative $N=200$ societies. Eighteen focal agents per condition are selected by round-robin ranking on distance, artifact work, social interaction, and successful actions so that the display spans the observed behavioral repertoire. Segment color gives instantaneous path speed in cells per tick. Blue circles mark observations or tests, gold diamonds mark artifact construction, purple triangles mark program installation, coral stars mark explicit interaction, and black squares mark active artifacts; square area increases with the number of distinct participants. (A) Full culture combines broad observation with dense construction, installation, and explicit social events around shared artifacts. (B) No explicit culture retains observation and physical artifact work but lacks the direct cultural channel. (C) No communication shows the corresponding spatially mediated activity when messages are unavailable. These are not simply tracks of where agents wandered: they show when travel became measurement, building, code deployment, or coordination, and where repeated work turned artifacts into local activity hubs.}
    \label{fig:speed-action-atlas}
\end{figure}

Event-linked trajectories reveal what agents were doing along those paths (Figure~\ref{fig:speed-action-atlas}). Observations and tests were distributed broadly, whereas construction, program installation, and repeated artifact interaction localized around a smaller number of shared sites. Explicit interaction markers appear only where the condition permits them. Artifact marker size records the number of distinct participants, providing a direct visual bridge between spatial hubs and multi-agent work. The resulting pattern is not a swarm motion benchmark in isolation; it is a map of how locomotion carries agents between measurement, fabrication, computation, and social exchange.

\begin{figure}[htbp]
    \centering
    \includegraphics[width=0.97\textwidth,height=0.64\textheight,keepaspectratio]
    {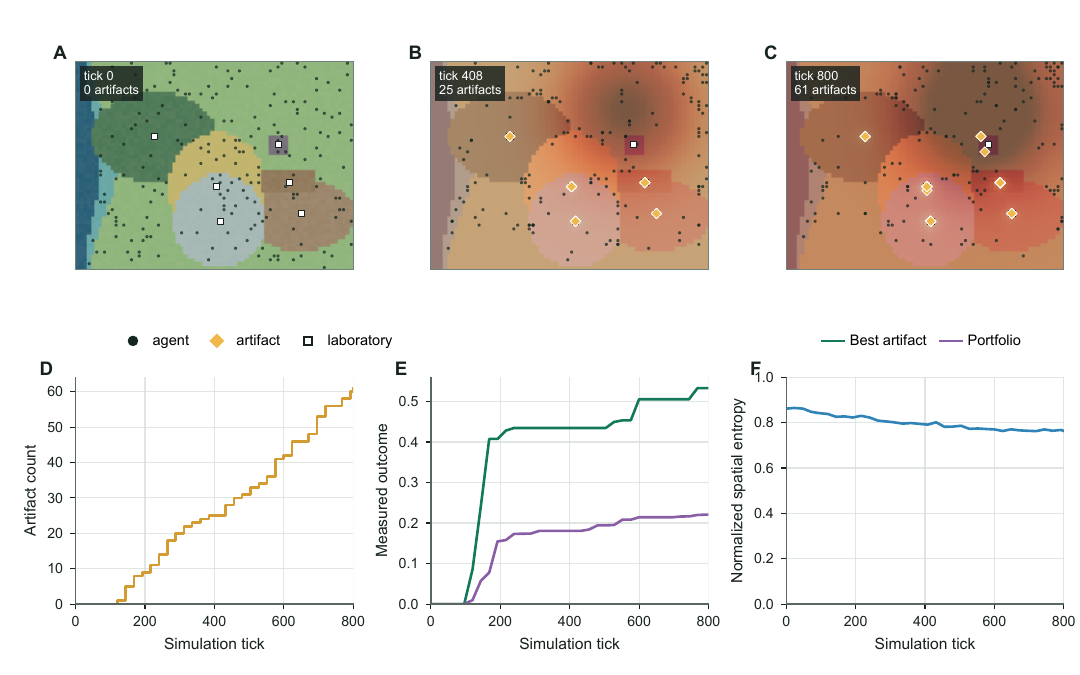}
    \caption[Within-run evolution of a shared biofoundry]{Within-run evolution of a representative no-explicit-culture society with $N=200$ and seed 3202. (A-C) Authoritative world states nearest ticks 0, 400, and 800; the stored snapshots occur at ticks 0, 408, and 800. Agents are black circles, artifacts are gold diamonds, and processing laboratories are open squares. The artifact count increases from 0 to 25 to 61 as the disturbance field and local activity pattern change. (D) Stepwise accumulation of persistent artifacts across the episode. (E) Running best artifact performance and portfolio resilience. The best-object curve rises rapidly after the first construction wave, whereas portfolio resilience grows more slowly as the society adds complementary technologies. (F) Normalized spatial entropy of agent occupancy on a fixed $10\times10$ grid declines from roughly 0.86 to 0.76, indicating moderate concentration rather than complete aggregation. Agents progressively convert an initially empty landscape into a persistent technological habitat; the first strong object appears quickly, but a useful portfolio and stable spatial organization take longer to assemble.}
    \label{fig:world-evolution}
\end{figure}

One representative society makes the temporal coupling between environmental change and technological accumulation visible (Figure~\ref{fig:world-evolution}). The simulation can additionally be inspected through an interactive interface that couples the evolving spatial world to agent state, society-level dynamics, and the underlying knowledge-lineage graph, enabling individual discoveries to be traced from observations and evidence through programs and downstream artifacts (Figure \ref{fig:biofoundry_ui}). The no-explicit-culture, $N=200$, seed-3202 world grew from no artifacts at tick 0 to 25 near tick 400 and 61 at tick 800. Best-artifact performance rose sharply once construction began, whereas portfolio resilience improved more gradually as additional artifacts accumulated. Spatial entropy declined only modestly, indicating concentration around productive sites without complete collapse into a single location. This trajectory is illustrative rather than inferential; the paired aggregate outcomes are those in Figure~\ref{fig:collective-results}.

\begin{figure}[htbp!]
    \centering
    \includegraphics[width=0.97\textwidth,height=0.74\textheight,keepaspectratio]
    {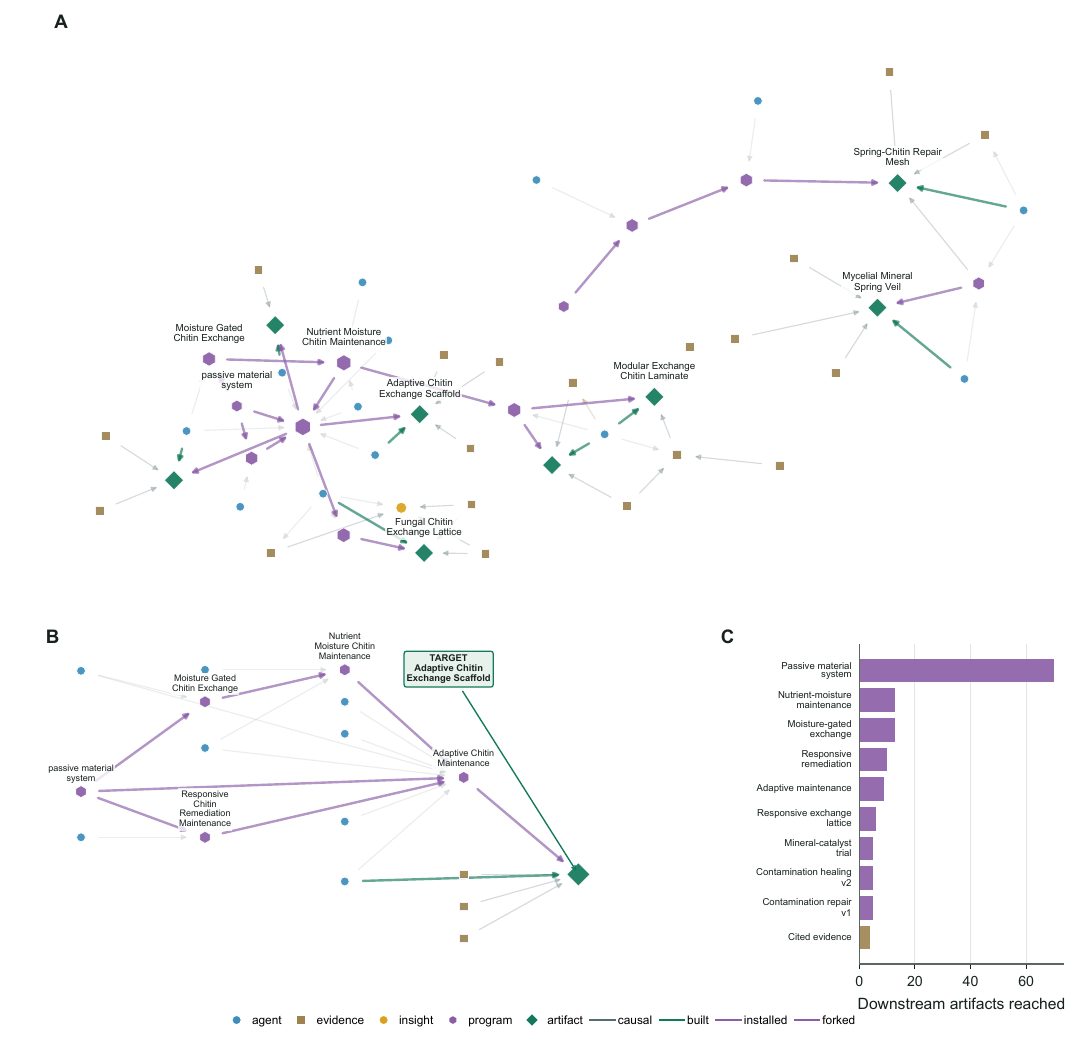}
    \caption[Recorded knowledge lineage]{Recorded knowledge lineage from evidence to technology. (A) Society-scale directed ancestry of eight high-performing active artifacts. Circles denote agents, squares denote evidence, small gold nodes denote insights, purple hexagons denote executable programs, and green diamonds denote artifacts. Typed arrows encode authorship or observation, recorded parent references, construction, installation, and program forking; node area increases with betweenness centrality. Labels are selectively placed to preserve readability. (B) Focused ancestry around \texttt{AdaptiveChitinExchangeScaffold}. Multiple agents, evidence records, precursor artifacts, and inherited programs converge on the target artifact; arrow direction is from recorded precursor to downstream record, program, or artifact. (C) Evidence, insight, and program nodes ranked by the number of distinct downstream artifacts they reach. The passive material system has the largest reach, followed by several maintenance and exchange records. Technologies are not isolated model responses: recorded observations and executable ideas become reusable building blocks that pass through several agents and contribute to multiple later artifacts.}
    \label{fig:knowledge-lineage}
\end{figure}

The recorded provenance graph identifies how evidence, programs, and precursor artifacts were reused (Figure~\ref{fig:knowledge-lineage}). The society-scale view contains multi-step paths from observations and authored insights through executable programs to eight high-performing artifacts. The focused ancestry of \texttt{AdaptiveChitinExchangeScaffold} combines several agents, multiple precursor technologies, and inherited programs. Downstream-reach ranking identifies a small set of reusable knowledge hubs, led by the passive material system, rather than only prolific message authors. Because edges are generated from recorded authorship, construction, installation, fork, and causal-parent events, the graph supports an attributable lineage claim; it does not infer causality from embedding similarity.

\begin{figure}[htbp]
    \centering
    \includegraphics[width=0.97\textwidth,height=0.8\textheight,keepaspectratio]
    {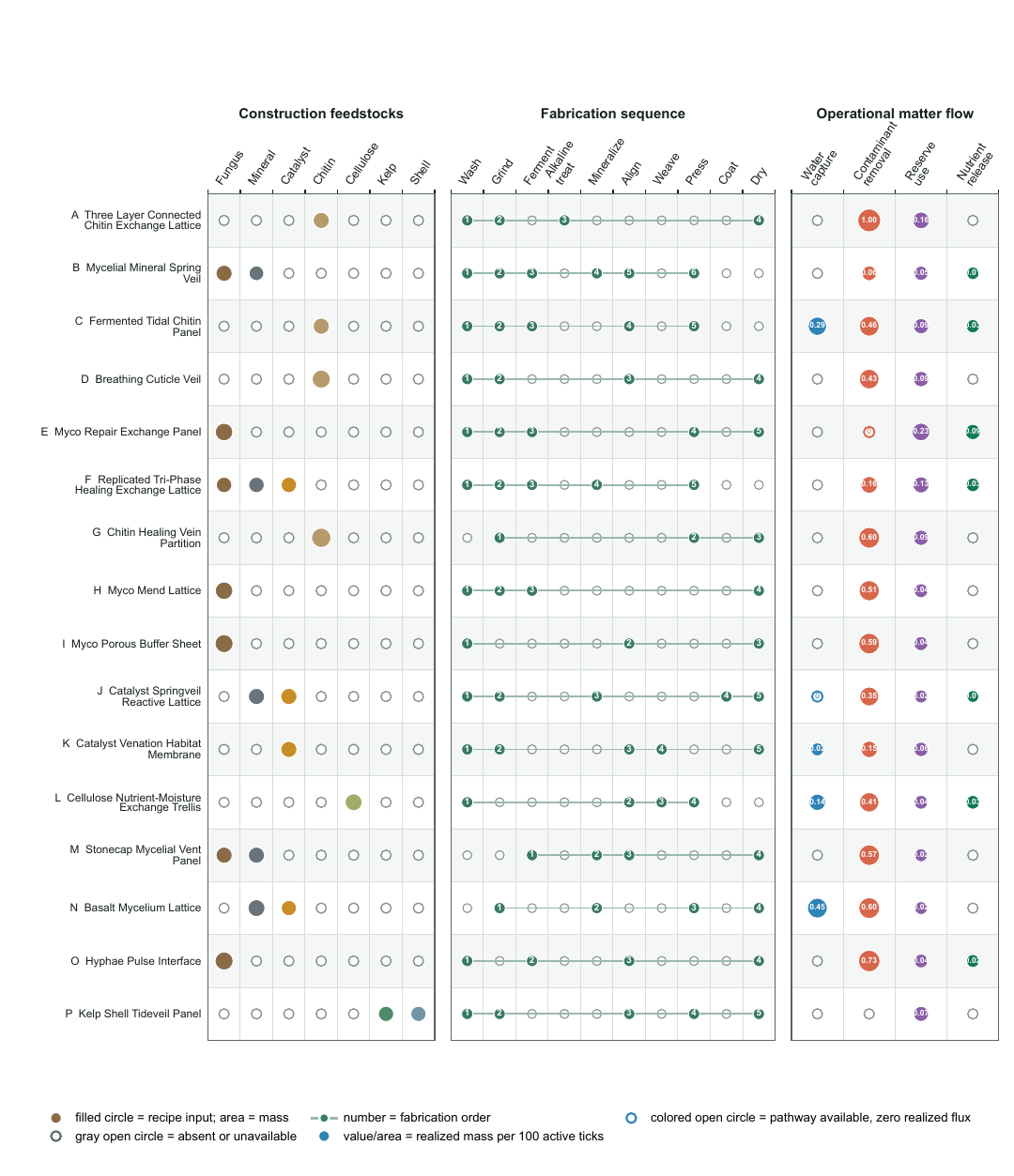}
    \caption[Construction and operational matter pathways]{Construction and operational matter pathways for the 16 technologies in Figure~\ref{fig:technology-gallery}, retained in the same A-P order. The left matrix reports construction feedstocks. Filled circles identify named recipe inputs and circle area increases with input mass; gray open circles mean that the material is absent. The center matrix reports fabrication sequence. Connected numbered circles provide the exact processing order, distinguishing, for example, wash-grind-ferment routes from mineralize-align-weave routes. The right matrix reports realized operational matter flow per 100 active ticks in the authoritative final snapshot. Filled colored circles and printed values encode the realized rate; colored open circles indicate that a controller pathway exists but had zero realized flux; gray open circles indicate that the pathway is unavailable. Water capture transfers local moisture into artifact storage, contamination removal reduces the environmental field, and embodied reserve supports repair, growth, and nutrient release. Construction feedstocks are consumed once at build time and are not automatically replenished during operation. The figure distinguishes what an artifact is made from, how it was fabricated, and what it actually did after construction; only the final columns are evidence of executed simulator function.}
    \label{fig:material-process-flows}
\end{figure}

The material-process map separates three layers that are often conflated in generative design claims (Figure~\ref{fig:material-process-flows}). Construction feedstocks specify what matter was consumed at build time; ordered fabrication nodes specify how those inputs were processed; operational fluxes specify what the resulting artifact actually moved or consumed during the authoritative simulation. Across the 16 exemplars, agents used distinct combinations of fungal, mineral, catalyst, chitin, cellulose, lignin, kelp, and shell resources, with recipes that included washing, grinding, fermenting, mineralizing, aligning, weaving, pressing, and drying. Realized operation most often involved water capture, contamination removal, and embodied reserve consumption. Available but unrealized pathways are shown separately from unavailable ones, preventing design claims from being mistaken for executed function.

\subsection{Long horizons expose metric-specific cultural benefits and tradeoffs}

\begin{figure}[htbp]
    \centering
    \includegraphics[width=0.97\textwidth,height=0.64\textheight,keepaspectratio]
    {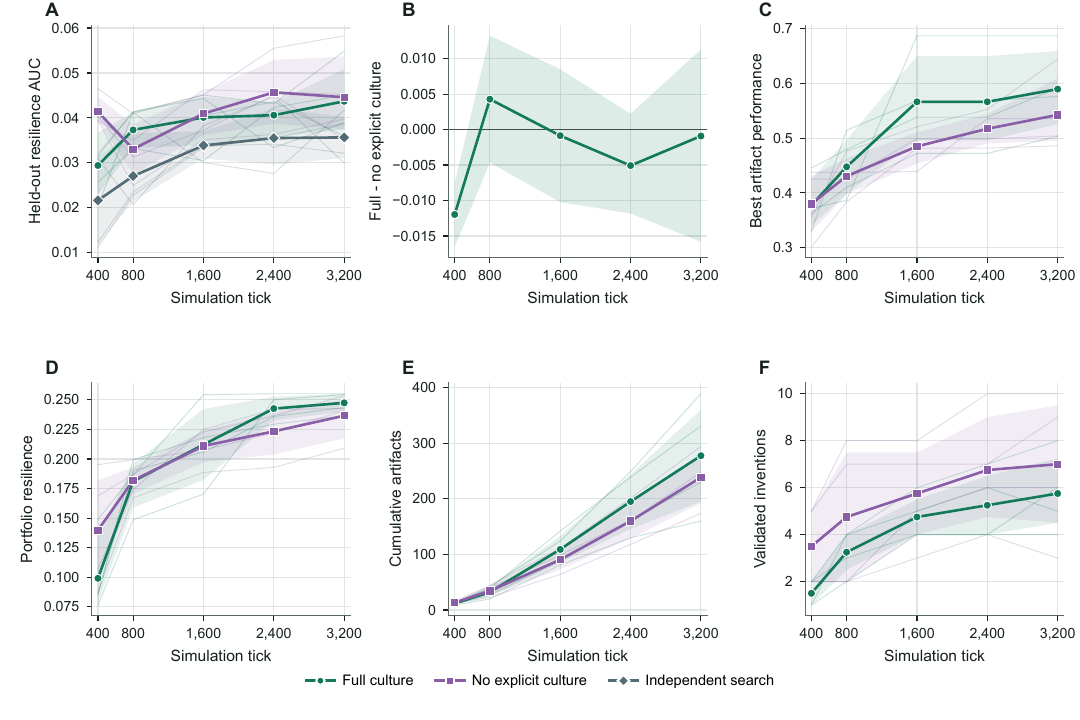}
    \caption[Outcome-dependent cultural crossover]{Outcome-dependent cultural crossover over 3,200 discovery ticks. Four matched $N=100$ seeds are evaluated at five frozen checkpoints. Thin lines are individual world seeds, heavy lines are seed means, and shaded regions are 95\% seed-bootstrap intervals. (A) Agent-free held-out resilience for full culture, no explicit culture, and the endpoint-wise best-of-100 independent-search envelope. Both shared worlds generally remain above the isolated envelope, but their ordering changes. (B) Paired full-minus-no-explicit-culture held-out effect; the mean changes sign and ends near zero. (C) Best active artifact performance crosses in favor of full culture by tick 800. (D) Portfolio resilience crosses near tick 1,600 and remains modestly higher under full culture. (E) Cumulative artifact production also crosses near tick 1,600 and ends at means of 277.5 versus 238.5 artifacts. (F) Validated inventions never cross; no explicit culture remains higher at every checkpoint and ends at 7.0 versus 5.75. Explicit culture helps some capabilities after enough time, but there is no single moment after which it improves everything.}
    \label{fig:long-horizon-crossover}
\end{figure}

Extending discovery to 3,200 ticks did not reveal one universal cultural crossover (Figure~\ref{fig:long-horizon-crossover}). Full culture overtook no explicit culture in mean best-artifact performance by tick 800, and crossed in portfolio resilience and cumulative artifact count near tick 1,600. It never overtook in validated invention count. Held-out resilience changed sign across checkpoints and was effectively tied at tick 3,200. The temporal result therefore rejects a single amortization threshold at which communication suddenly becomes beneficial for every objective. Culture changes the developmental trajectory, but each endpoint responds on its own timescale.

\begin{figure}[htbp]
    \centering
    \includegraphics[width=0.97\textwidth,height=0.64\textheight,keepaspectratio]
    {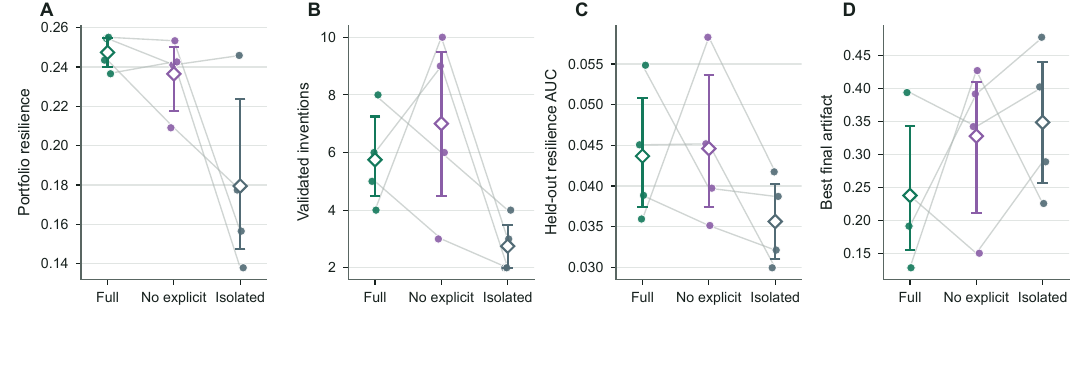}
    \caption[Long-horizon endpoint tradeoffs]{Long-horizon endpoint tradeoffs at tick 3,200 for four matched $N=100$ world seeds. Conditions are full culture, no explicit culture, and an endpoint-wise best-of-100 isolated envelope. Filled points are individual seed outcomes, gray lines preserve within-seed pairing, open diamonds are means, and error bars are 95\% seed-bootstrap intervals. (A) Portfolio resilience favors both shared worlds; full culture averages 0.2474, no explicit culture 0.2365, and isolated search 0.1794. (B) Validated inventions likewise favor shared worlds, with means of 5.75, 7.00, and 2.75. (C) Held-out resilience is highest on average without explicit culture, 0.0446, compared with 0.0356 in the isolated envelope; full culture is similar to no explicit culture at the endpoint. (D) Best final artifact reverses the ordering: the isolated envelope averages 0.3488 versus 0.2380 for full culture. Societies win by maintaining several complementary technologies, while isolated agents can still win a contest defined only by the strongest single artifact.}
    \label{fig:long-horizon-tradeoff}
\end{figure}

The final checkpoint clarifies what shared societies and isolated search optimize (Figure~\ref{fig:long-horizon-tradeoff}). Full culture achieved mean portfolio resilience 0.2474 and no explicit culture 0.2365, both above the isolated envelope at 0.1794. The shared worlds produced 5.75 and 7.00 validated inventions, respectively, versus 2.75 in isolated search. No explicit culture also improved held-out resilience to 0.0446 versus 0.0356 for the isolated envelope. Yet the isolated control retained the strongest final single artifact: 0.3488 versus 0.2380 under full culture. Shared worlds therefore created a broader technological ecology, whereas independent parallel search remained competitive for record-setting single-object optimization.

\begin{figure}[htbp]
    \centering
    \includegraphics[width=0.97\textwidth,height=0.64\textheight,keepaspectratio]
    {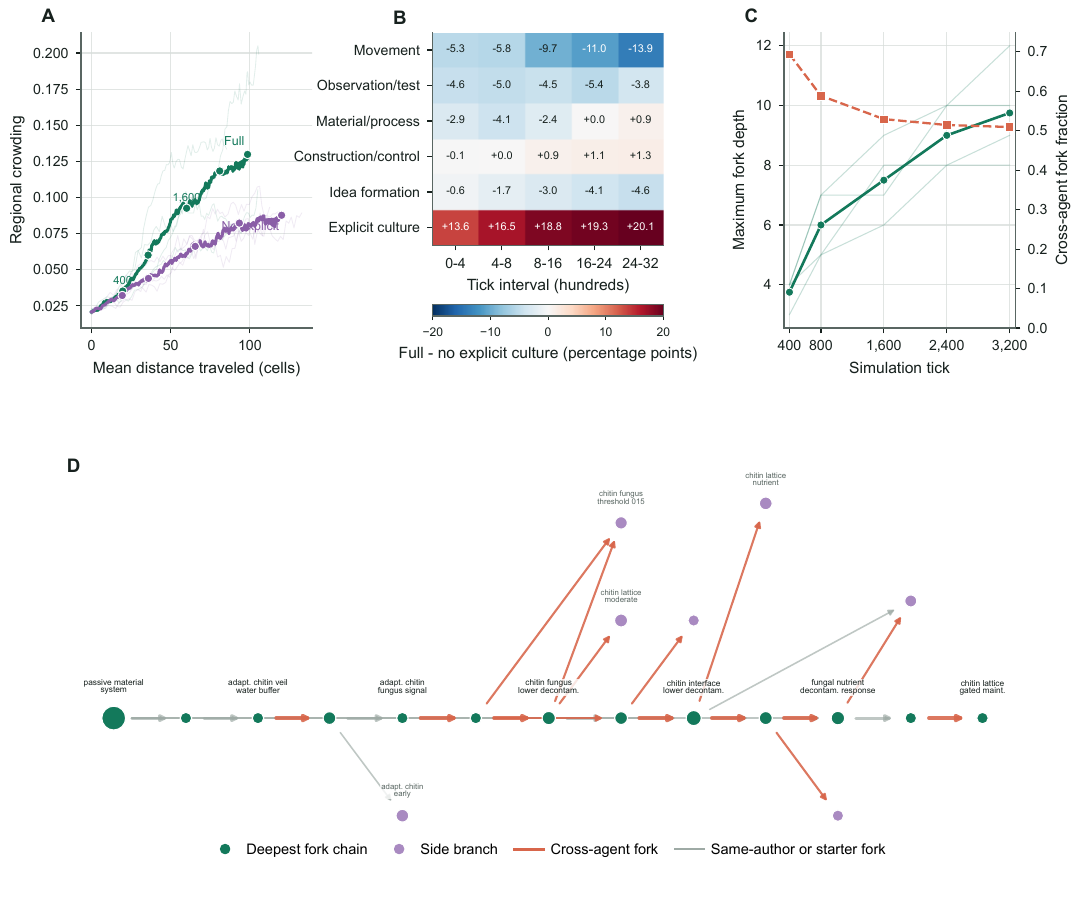}
    \caption[Spatial self-organization and executable lineage]{Cultural self-organization couples spatial hubs to executable code descent. (A) Phase portrait of mean cumulative path length and local regional crowding for full culture and no explicit culture. Light trajectories are the four matched seeds, heavy trajectories are seed means, numbered checkpoints indicate ticks 400, 800, 1,600, 2,400, and 3,200, and arrows point forward in time. Full culture ends with shorter paths and greater crowding, consistent with localized technological hubs. (B) Full-minus-no-explicit-culture allocation of active actions in five time windows, in percentage points. Movement and observation decline relative to the ablation, while explicit cultural activity and construction/control increase. (C) Mean maximum executable-program fork depth rises from 3.75 at tick 400 to 9.75 at tick 3,200, while approximately half of eligible forks remain cross-agent. (D) Exact deepest content-addressed lineage from full culture, seed 3301. Green nodes form the 12-edge longest path, purple nodes are deterministic one-hop branches, coral arrows are cross-agent forks, gray arrows are same-author or starter forks, and node area increases with installation count. The society gradually stops roaming as widely, concentrates around shared infrastructure, and builds a many-generation inheritance system for executable technology.}
    \label{fig:dynamics-lineage}
\end{figure}

The long-horizon dynamics reveal how explicit culture changes the use of time and space (Figure~\ref{fig:dynamics-lineage}). Full-culture agents traveled less by the endpoint, a mean path length of 98.5 cells versus 120.0 without explicit culture, while regional crowding rose to 0.1298 versus 0.0877. By ticks 2,400 to 3,200, full culture allocated 13.9 percentage points less activity to movement and 20.1 points more to explicit cultural actions; construction/control was 1.3 points higher. At the same time, mean executable lineage depth reached 9.75 and roughly half of eligible forks crossed author boundaries. The deepest representative genealogy contains 12 fork edges. Thus, spatial localization was accompanied by an increasingly deep executable culture rather than simple inactivity.

\begin{figure}[htbp]
    \centering
    \includegraphics[width=1.\textwidth,height=0.8\textheight,keepaspectratio]
    {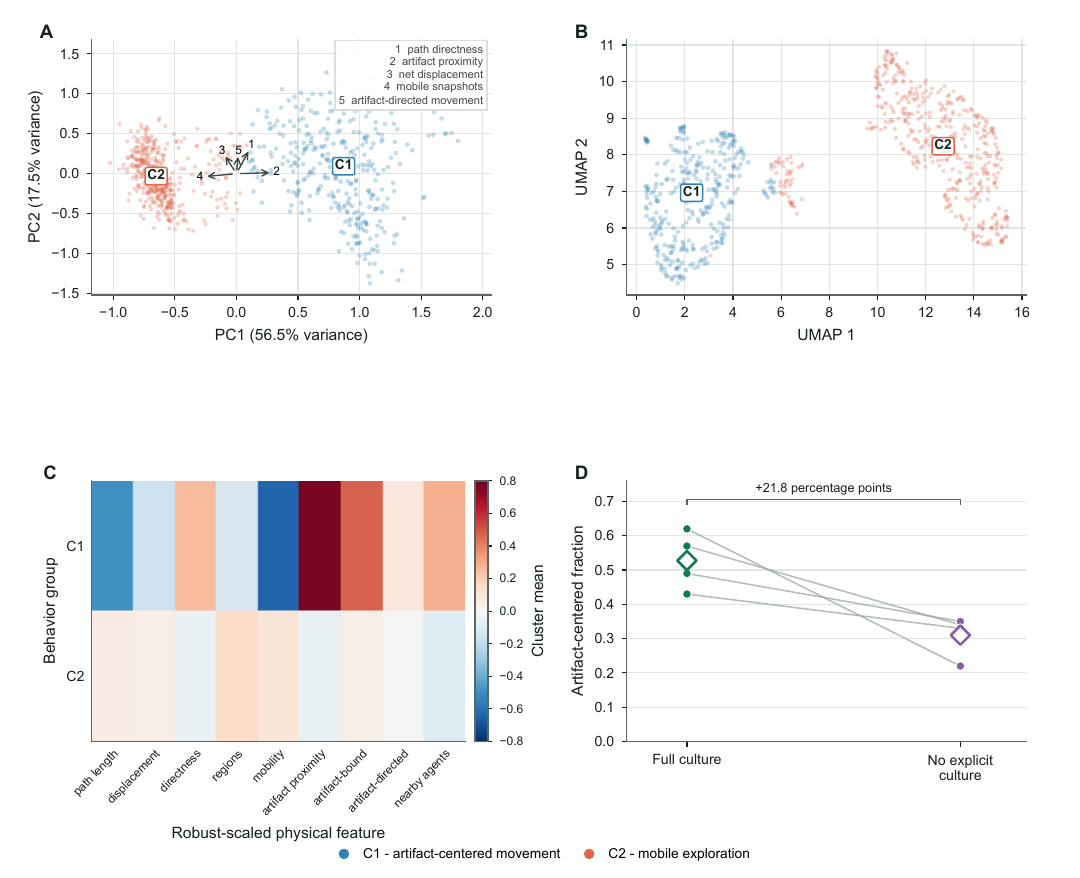}
    \caption[Long-horizon physical behavior phenotypes]{Explicit culture reorganizes physical behavior around shared artifacts. Each point in panels A and B is a complete 3,200-tick trajectory from the eight $N=100$ shared-world episodes. A two-group k-means model is fit in the original nine-dimensional robust-scaled physical feature space without condition, seed, identity, communication, cultural-action, or technology-work labels. (A) PCA gives an auditable linear display; numbered arrows point toward increasing values of the five strongest loading features. (B) UMAP shows nonlinear neighborhood structure, but its axes have no direct behavioral meaning. (C) Cluster means identify C1 as artifact-centered movement, characterized by greater artifact proximity and artifact-bound motion with shorter and less frequent travel, and C2 as mobile exploration. (D) Thin lines pair the four seeds, filled circles are seed fractions, and open diamonds are means. Full culture increases the C1 fraction from 31.0\% to 52.8\%, a paired gain of 21.8 percentage points; the 95\% seed-bootstrap interval is 12.0 to 33.5 points and the clustering silhouette is 0.472. Explicit culture changes where agents physically spend their lives, not only what they say.}
    \label{fig:long-horizon-behavior}
\end{figure}

A second label-blind model provides a complementary characterization of how the cultural intervention was associated with physical behavior (Figure~\ref{fig:long-horizon-behavior}). The clustering used only nine movement and artifact-proximity features from all 800 complete trajectories in the eight shared-world episodes; explicit messages, cultural actions, condition labels, and technology-work counts were excluded. One phenotype combined high artifact proximity and artifact-bound movement with shorter, less frequent travel, while the other represented mobile exploration. Full culture placed 52.8\% of agents in the artifact-centered phenotype versus 31.0\% without explicit culture, a paired difference of 21.8 percentage points with a seed-bootstrap 95\% interval from 12.0 to 33.5 points. This provides a descriptive behavioral signature of self-organization that does not depend on labeling utterances as roles.

\subsection{Technological networks become modular, persistent, diffusive, and selectively vulnerable}

\begin{figure}[htbp]
    \centering
    \includegraphics[width=1.\textwidth,height=0.9\textheight,keepaspectratio]
    {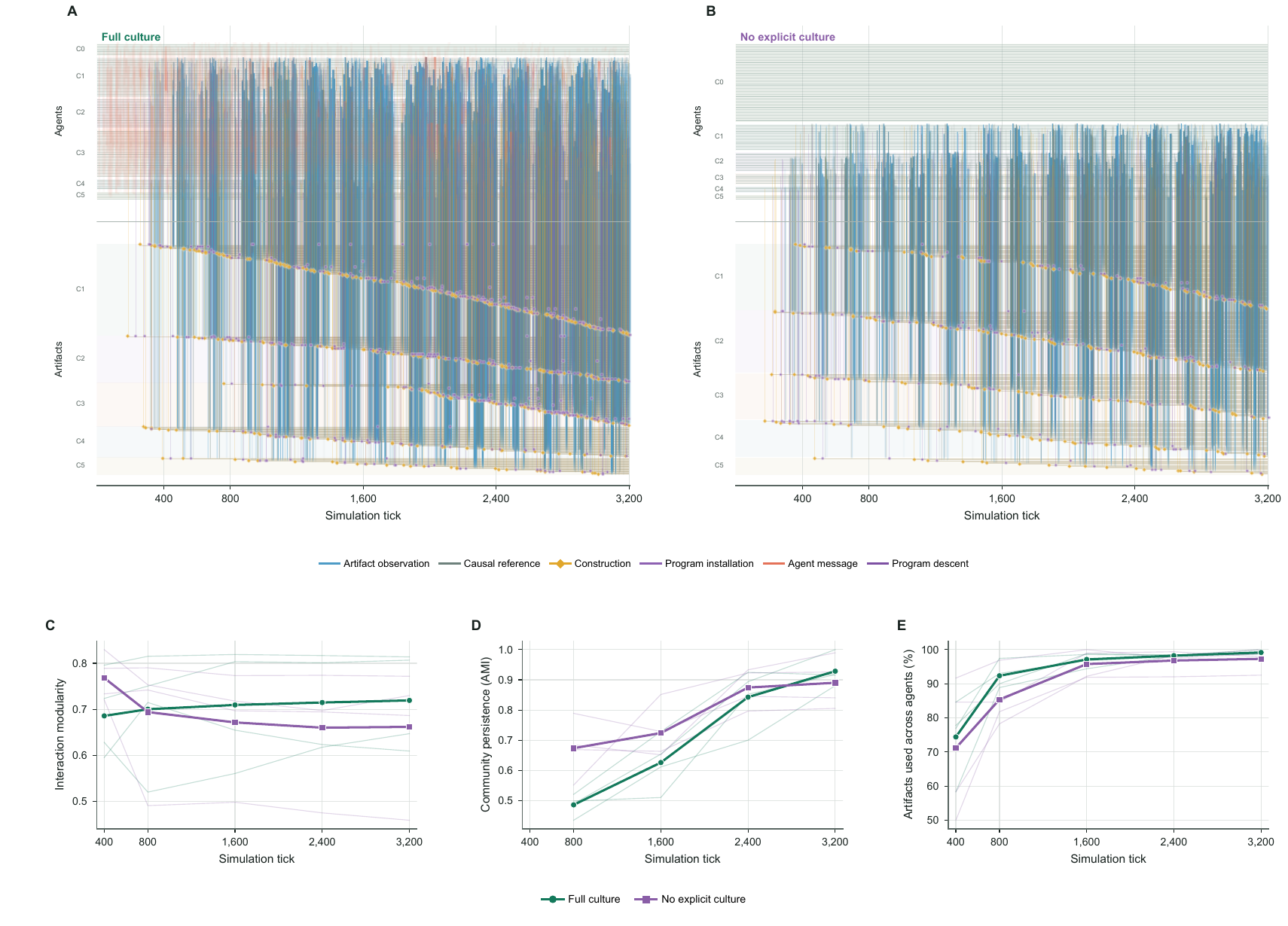}
    \caption[Society-scale temporal circuitry]{A technological society develops persistent temporal circuitry. (A, B) Complete recovered interaction histories for matched $N=100$, seed-3301 full-culture and no-explicit-culture societies over 3,200 ticks. Horizontal tracks in the upper band are agents and tracks in the lower band are artifacts; each artifact track starts at construction. Vertical strokes connect entities at the exact event tick. Blue encodes observation, gray-green recorded parent reference, gold construction, purple program installation or executable descent, and coral explicit messaging. Repeated observations and messages are display-binned with exact counts retained; construction and executable events remain at exact ticks. Tracks are ordered by final data-derived community, not assigned role. The full-culture example records 389 artifacts, 3,924 delivered-message recipient edges, and 358 cross-artifact program-descent events; the matched ablation records 248 artifacts and lacks the disabled event types. (C) Interaction modularity remains substantial, indicating local technological neighborhoods. (D) Adjusted mutual information between successive community assignments rises, showing that those neighborhoods acquire memory. (E) The percentage of artifacts touched by a noncreator approaches saturation in both conditions. A dense web of recurring relationships grows over time, and explicit culture adds message and code-inheritance pathways without replacing physical artifact-mediated coordination.}
    \label{fig:temporal-circuitry}
\end{figure}

Complete event histories show the society assembling a persistent temporal circuit (Figure~\ref{fig:temporal-circuitry}). In the representative full-culture seed, 389 artifact tracks become linked to 100 agent tracks by observations, recorded parent references, construction, program installation, 3,924 delivered-message recipient edges, and 358 cross-artifact program-descent events. The matched no-explicit-culture world contains 248 artifacts and, by intervention, neither explicit messages nor cross-artifact program descent. Across four seeds, both conditions retained moderate interaction modularity, successive community assignments became increasingly persistent, and more than 95\% of artifacts were eventually used by a noncreator. Explicit culture therefore added a denser social and executable layer to an artifact-mediated structure that was already capable of broad cross-agent uptake.

\begin{figure}[htbp]
    \centering
    \includegraphics[width=0.97\textwidth,height=0.64\textheight,keepaspectratio]
    {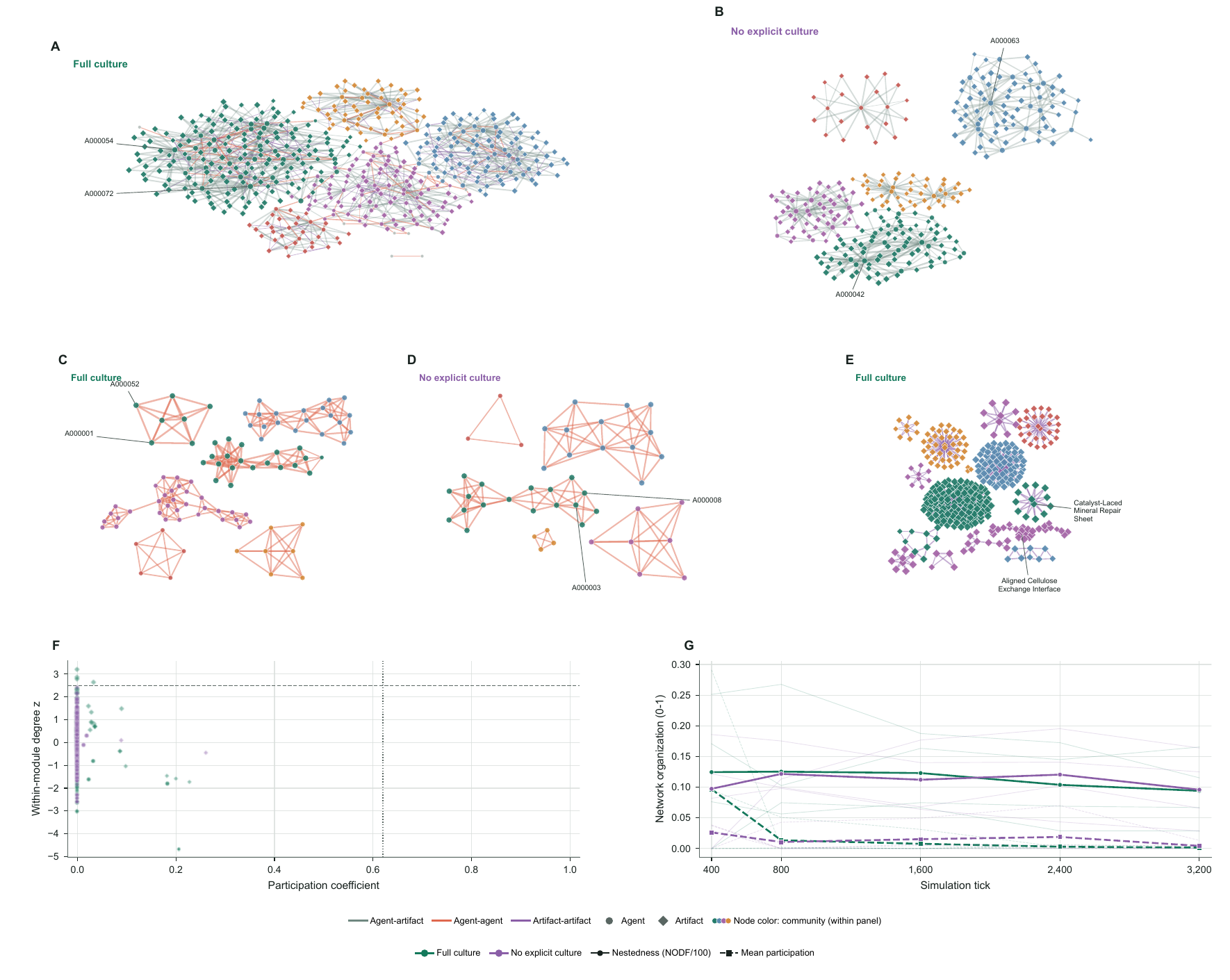}
    \caption[Network anatomy of technological neighborhoods]{Network anatomy of the long-horizon technological society. (A, B) Deterministic visualization backbones of matched $N=100$, seed-3301 networks accumulated over 3,200 ticks. The underlying graph links agents to artifacts through observation, recorded parent reference, construction, contribution, programming, repair, and dismantling. The display retains each node's two strongest physical ties and a maximum-weight spanning tree for each component, then adds the strongest local social-exchange and executable-lineage ties. This is a reproducible readability filter, not a significance threshold; complete graphs are retained in GraphML. Full culture produces 489 displayed nodes and 1,156 edges, versus 293 nodes and 514 edges without explicit culture. Circles are agents, diamonds are artifacts, edge color denotes interaction layer, and node area increases with interaction count. Node color denotes community within a panel; colors do not identify matched communities across conditions. Labels beginning with \texttt{A} are stable agent identifiers. (C, D) Degree-conditioned agent coordination projections. An agent pair is validated only when it shares at least two artifacts and its overlap exceeds a hypergeometric null conditioned on both agents' artifact degrees, after Benjamini--Hochberg correction at $q\leq0.05$. Full culture yields 1,100 validated pairs among 92 agents, compared with 261 pairs among 45 agents without explicit culture. The display keeps each agent's four strongest local surprise links, while GraphML retains all validated pairs. (E) The analogous artifact co-use projection joins technologies used by unexpectedly overlapping agent populations, not technologies with merely similar names. Full culture contains 18,563 validated pairs among 385 artifacts; 710 strongest local links are displayed across 15 components. (F) Node-role cartography. Participation coefficient measures cross-community mixing, and within-module z-score measures local centrality. Reference lines at 0.62 and 2.5 are heuristic thresholds from prior work rather than universal classification boundaries. All nodes have participation below 0.26, and only four full-culture artifacts exceed the hub reference line. (G) Replicated temporal comparison across four matched seeds and five checkpoints. Thin lines are seeds and heavy lines are means; solid curves show binary NODF nestedness divided by 100 and dashed curves show mean agent participation. At tick 3,200, mean nestedness is 0.094 under full culture and 0.096 without explicit culture, while mean participation is 0.0016 and 0.0044. Explicit culture creates a larger and more statistically connected technological network, but it does not merge the society into one centralized hierarchy. Both conditions form specialized local neighborhoods, with explicit culture adding more pathways among them.}

    \label{fig:network-science}
\end{figure}

Figure~\ref{fig:network-science} asks a more specific question than whether the interaction network is large: what kind of organization did the agents construct? The analysis begins from the complete typed event history and forms a weighted bipartite graph linking agents to artifacts through observation, recorded parent reference, construction, contribution, programming, repair, and dismantling. Repeated events are compressed logarithmically so that recurrence matters without allowing one frequently sampled relationship to dominate. Panels A and B show deterministic visualization backbones, not the full graphs and not a significance test. For each node, the two strongest physical ties are retained, and a maximum-weight spanning tree preserves the connected structure of every component; the strongest local social-exchange and executable-lineage ties are then added. The matched full-culture world contains 489 participating nodes and 1,156 displayed backbone edges, including 105 agent-agent and 218 artifact-artifact edges. The no-explicit-culture backbone contains 293 participating nodes and 514 displayed edges, all physical by intervention. Thus, the visual difference reflects both more constructed artifacts and the additional social and executable layers created by explicit culture.

Panels C-E then replace descriptive edge selection with a degree-controlled statistical question. Two agents are joined only when they share at least two artifacts and their observed overlap exceeds a hypergeometric null conditioned on both agents' artifact degrees, after Benjamini--Hochberg correction over all possible pairs~\cite{benjamini1995controlling}. The full-culture seed contains 1,100 validated agent pairs involving 92 agents, compared with 261 pairs involving 45 agents without explicit culture. The displayed projections retain each agent's four strongest local surprise links, yielding 270 versus 116 visible edges; all validated edges remain in the exported graph. Panel E reverses the projection: two artifacts are joined when they are used by unexpectedly overlapping agent populations. In full culture, 18,563 artifact pairs among 385 artifacts pass the same corrected test, and the display retains 710 strongest local links. These null-corrected projections show that the denser full-culture pattern is not explained only by some agents or artifacts having higher degree.

The remaining panels distinguish expansion from global integration. In panel F, participation coefficient measures how evenly a node's weighted ties cross communities, whereas within-module z-score measures whether it is unusually central inside its own community. The reference lines at participation 0.62 and within-module z-score 2.5 are heuristic role-cartography thresholds from prior work, not universal classification boundaries~\cite{guimera2005cartography}. Every displayed node has participation below 0.26, and only four full-culture artifacts exceed the hub reference line. Panel G extends the comparison across four matched seeds and five checkpoints. By tick 3,200, mean NODF nestedness is nearly identical under full culture and no explicit culture, 0.094 and 0.096 after scaling to 0--1, and mean agent participation is also close to zero, 0.0016 and 0.0044. Explicit culture therefore expands the number and variety of statistically supported coordination pathways without producing one centralized or globally mixed society. The emergent architecture is instead a larger mosaic of locally coherent technological neighborhoods connected by a small number of hubs and bridges.

\begin{figure}[htbp]
    \centering
    \includegraphics[width=0.97\textwidth,height=0.64\textheight,keepaspectratio]
    {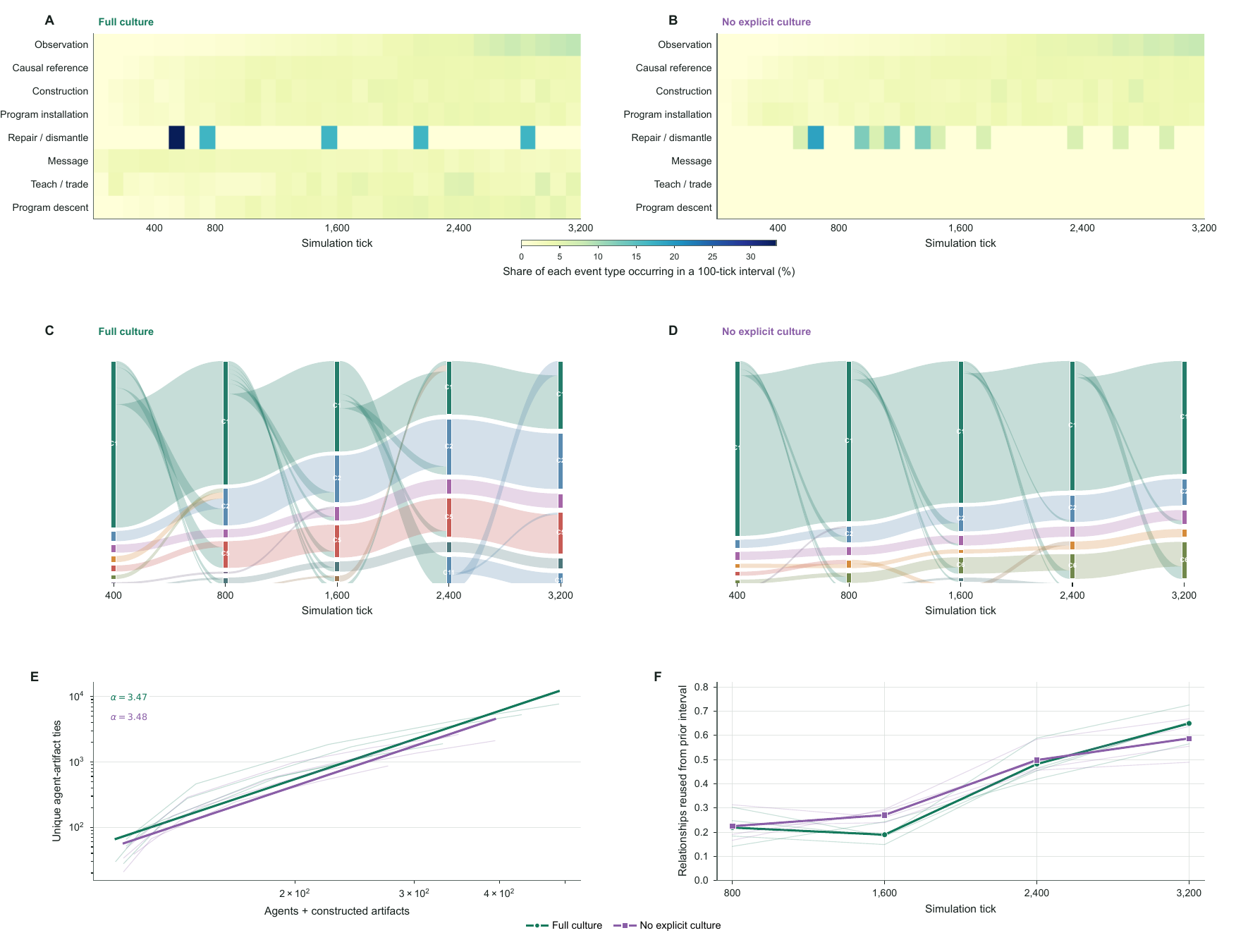}
    \caption[Dynamic assembly of technological networks]{Technological interaction networks assemble, differentiate, and retain relationship memory. (A, B) Temporal allocation of eight event classes across 32 nonoverlapping 100-tick intervals under full culture and no explicit culture. Each row is normalized to sum to 100\%, so color locates when that event type occurs rather than comparing absolute counts between rows. (C, D) Alluvial maps follow all 100 agents in matched seed 3301 among Louvain communities at ticks 400, 800, 1,600, 2,400, and 3,200. Community identities are propagated by maximum-overlap matching, and ribbon width is the exact number of transitioning agents. (E) Cumulative unique agent-artifact ties plotted against agents plus constructed artifacts on log-log axes. Thin lines are seeds, heavy lines are means, and the fitted descriptive exponents are $\alpha=3.47$ for full culture and $\alpha=3.48$ without explicit culture. The exponent describes densification with network size, not growth per unit time. (F) Fraction of interval ties reused in the next interval, rising to means of 0.650 and 0.588. Both societies develop stable neighborhoods, but full culture produces roughly twice as many distinct agent-technology relationships by tick 3,200 and retains more of them from one period to the next.}
    \label{fig:dynamic-network-assembly}
\end{figure}

The network changes through both expansion and memory (Figure~\ref{fig:dynamic-network-assembly}). Event-class heat maps show early repair and dismantling pulses and a later redistribution toward observation, recorded parent use, construction, teaching, and program descent. Community alluvial diagrams reveal substantial early reassignment followed by larger persistent streams. Cumulative unique agent-artifact ties grow superlinearly with the total number of agents plus artifacts, with descriptive log-log exponents of 3.47 and 3.48. The similar exponents do not imply equal network size: by tick 3,200, full culture has a mean 4,031.3 unique ties versus 2,027.3 without explicit culture. Relationship reuse from one interval to the next reaches 0.650 and 0.588, respectively. Explicit culture therefore enlarges the realized relational substrate, while both conditions acquire repeated local interaction patterns.

\begin{figure}[htbp]
    \centering
    \includegraphics[width=0.97\textwidth,height=0.64\textheight,keepaspectratio]
    {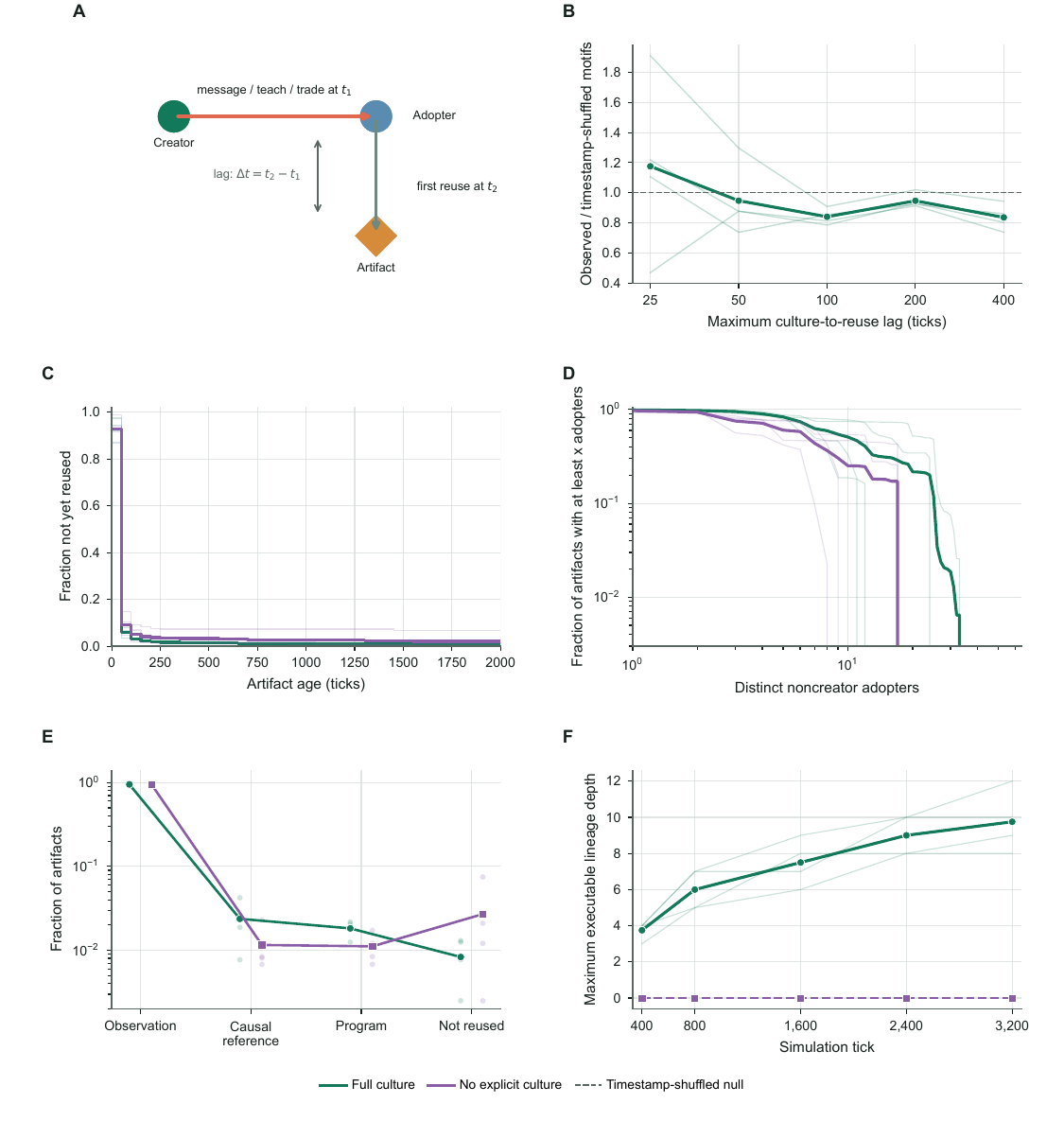}
    \caption[Technology diffusion and adoption pathways]{Technology diffusion through the shared world. (A) Temporal motif tested under full culture. A creator messages, teaches, or trades with the eventual first adopter at $t_1$, and that agent first reuses the artifact at $t_2$. The double-headed bracket denotes the elapsed lag $\Delta t=t_2-t_1$; it is not an additional interaction edge. (B) Observed motif frequency divided by a 200-replicate timestamp-shuffled null that preserves directed dyads and the global activity schedule. The dashed line is parity. The mean ratio is 1.175 at a 25-tick lag but remains below one at 50 to 400 ticks, so direct transmission is not consistently enriched. (C) Kaplan--Meier fraction of artifacts not yet reused, with unreused artifacts right-censored at tick 3,200~\cite{kaplan1958nonparametric}. (D) Complementary cumulative distribution of distinct noncreator adopters. The full-culture intervention produces earlier and broader reuse. (E) First-reuse channels; approximately 95\% of adoption begins with physical observation in both conditions, while recorded parent reference and programming are much rarer immediate channels. (F) Maximum executable lineage depth grows to a mean 9.75 under full culture and remains zero where cross-agent program descent is disabled. The explicit-culture intervention is associated with faster and broader reuse, but most agents encounter technology through the shared world rather than receiving a direct handoff from the inventor.}
    \label{fig:cultural-diffusion}
\end{figure}

Technology diffusion was faster and broader under full culture, but it did not follow a simple inventor-to-adopter message cascade (Figure~\ref{fig:cultural-diffusion}). Across seeds, 99.3\% of full-culture artifacts and 96.9\% of no-explicit-culture artifacts were reused by a noncreator. Median time to first reuse among reused artifacts was 5 versus 8 ticks, and mean adoption breadth was 13.53 versus 7.49 noncreator agents. Approximately 95\% of first reuse occurred through direct physical observation in both conditions. Direct creator-to-adopter contact exceeded a timestamp-shuffled null only weakly at the shortest 25-tick window and fell below parity for longer windows. Culture therefore appears to alter the society-wide informational and physical network, after which agents commonly discover technology through the world rather than receiving it directly from its inventor.

Finally, structural knockout assays distinguish distributed redundancy from hub dependence (Figure~\ref{fig:structural-robustness}). Randomly removing half of the agents left 98.3\% of full-culture artifacts and 95.2\% of no-explicit-culture artifacts connected to at least one surviving agent. In contrast, high-degree removal reduced access to 59.6\% and 73.9\%, while broker removal reduced it to 62.9\% and 68.4\%. The largest connected component showed the same ordering. Full culture therefore distributed participation broadly enough to tolerate random dropout, yet concentrated sufficient traffic in high-degree and high-betweenness agents to create targeted vulnerabilities. These are topological measurements on the recorded network; they do not demonstrate physical service, adaptation, or recovery after removing agents from a live simulation.

\subsection{Granular roles recur as dynamic states rather than fixed identities}

The complete trajectories support a more granular analysis than the two broad worker-explorer phenotypes (Figure~\ref{fig:dynamic-roles}). Every shared-world trajectory was divided into nonoverlapping 200-tick windows. This produced 22,400 agent-windows from 20 episodes: twelve N=200, 800-tick societies under full culture, no explicit culture, and no communication, plus eight N=100, 3,200-tick societies under full culture and no explicit culture. The fit used a deterministic episode-balanced sample of 16,000 windows so that every condition-seed world contributed equally. Thirteen movement, spatial-context, task-action, and cultural-interaction features were robust-scaled; condition, study, population, seed, identity, and time were withheld.

Model selection first recovered the familiar two broad modes with silhouette 0.467, then resolved each parent independently into two occupied submodes. Post hoc profiles identified four recurring states: constructor/operator, artifact-local caretaker, cultural coordinator, and mobile surveyor. This hierarchy is not inferred from the arrows or apparent islands in the PCA display. It is fit in the original 13-dimensional feature space, requires every child to contain at least 5\% of its parent, and remains stable when replicate families are withheld: mean adjusted Rand index is 0.999 for the broad split and 0.921 and 0.980 for the two conditional splits.

The resulting sequences reveal behavioral succession. In the 800-tick, N=200 full-culture study, the mean constructor/operator fraction rises from 0 in the first 200-tick window to 0.240 in the last, while the cultural-coordinator fraction falls from 0.814 to 0.560. In the 3,200-tick, N=100 full-culture study, constructors rise from 0.008 to 0.535 and coordinators fall from 0.695 to 0.292. No-explicit-culture societies also develop more constructors, but end at 0.252 while mobile surveyors remain the majority at 0.633. Full culture is also more behaviorally fluid: mean window-to-window switching is 0.270 versus 0.102 in the 800-tick full and no-explicit-culture conditions, and 0.244 versus 0.138 over 3,200 ticks. The same agents can therefore change activities as the technological ecology matures; the roles are not permanent social classes.

The cultural-coordinator state must be interpreted carefully because two of its defining features are disabled by the ablations. An anti-circularity sensitivity therefore removes explicit culture/coordination and social-contact features before fitting. The remaining 11 physical and task features independently support three modes with silhouette 0.551: mobile observation/testing, stationary artifact-proximal construction, and artifact-local movement with material work. Thus, granular physical/task differentiation is not merely a relabeling of whether communication was allowed. Agent-window assignments remain descriptive, and paired condition contrasts use the four simulation seeds. Recurrence across the two studies is not attributed to horizon alone because population, seed family, and decision schedule also change.

\subsection{The collective architecture transfers to a distinct volcanic materials world}

\begin{figure}[]
    \centering

    \includegraphics[
        width=0.9\textwidth,
        height=0.42\textheight,
        keepaspectratio
    ]{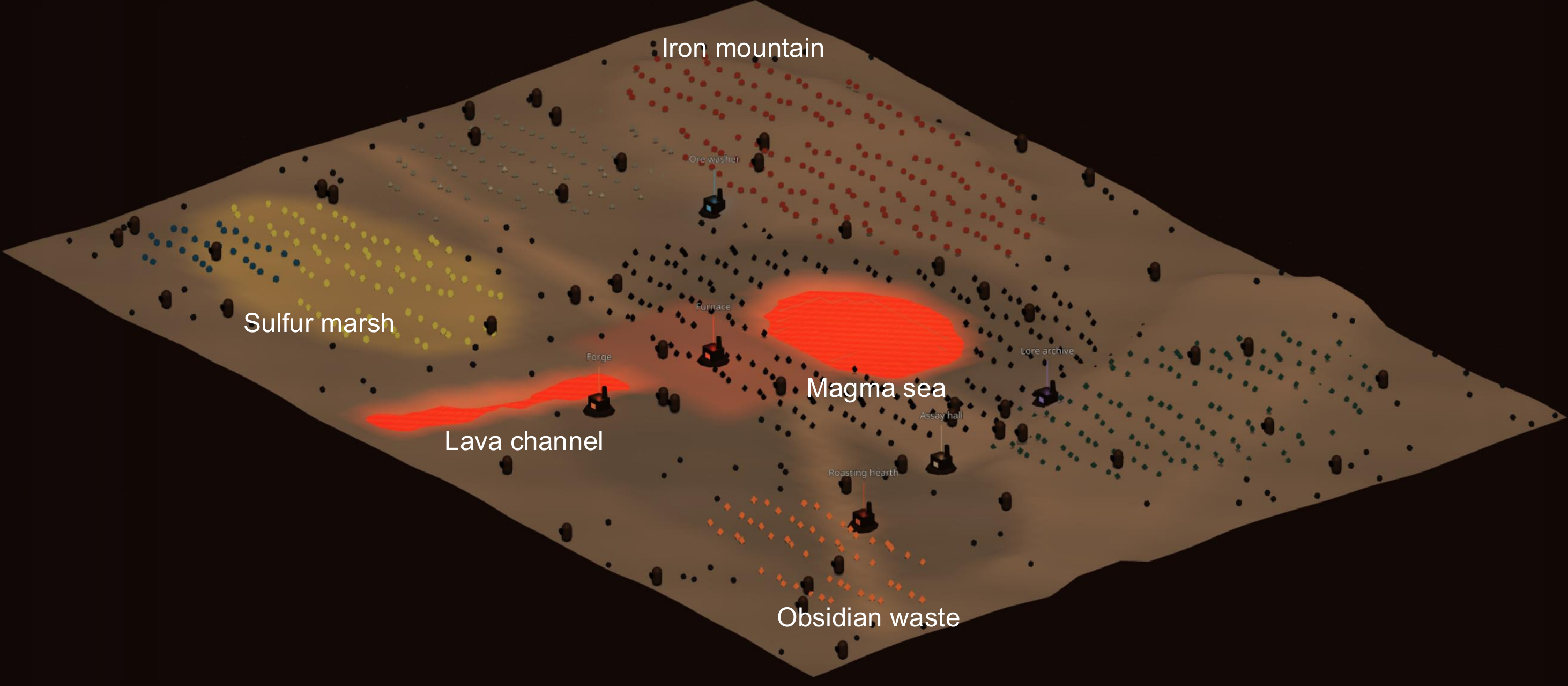}

    \vspace{0.4cm}

    \includegraphics[
        width=0.97\textwidth,
        height=0.6\textheight,
        keepaspectratio
    ]{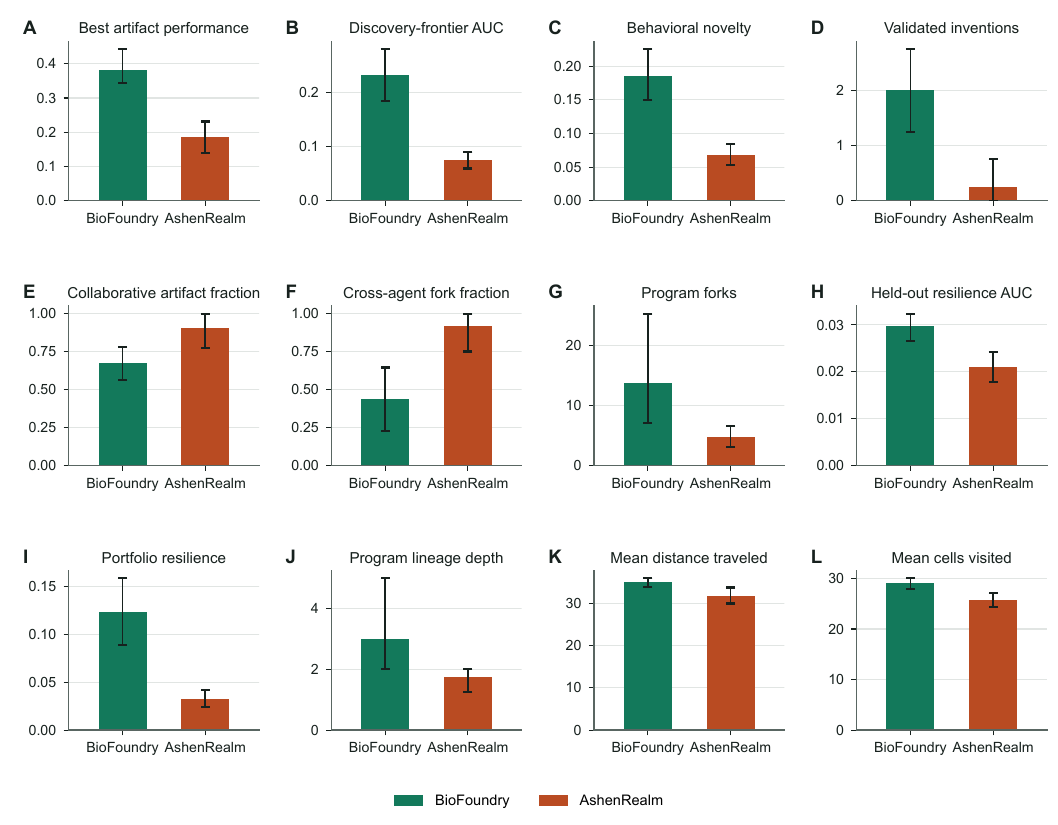}

    \caption{Zoomed-in view of a representative SwarmWorld AshenRealm environment. The rendering shows the $72\times54$-cell world for seed 1703, highlighting the spatial substrate through which agents explore, gather resources, fabricate materials, and construct persistent technologies. Distinct environmental regions include magma sea and resource biomes such as the iron mountain, lava channel, sulfur marsh, and obsidian waste. Cross-world transfer of collective technological behavior from BioFoundry to AshenRealm. Twelve endpoints summarize societies operating in the original BioFoundry and the independently specified volcanic AshenRealm environment: (A) best-artifact performance, (B) discovery-frontier AUC, (C) behavioral novelty, (D) validated inventions, (E) collaborative-artifact fraction, (F) cross-agent program-fork fraction, (G) number of program forks, (H) held-out resilience AUC, (I) portfolio resilience, (J) program-lineage depth, (K) mean distance traveled, and (L) mean number of cells visited. AshenRealm occupies a different functional regime, with lower performance and resilience measures, but retains multi-agent construction, executable inheritance, lineage formation, and broad spatial exploration.}
    \label{fig:ashenrealm_comparison}
\end{figure}

We next asked whether the organizational phenomena observed in BioFoundry depended on its particular material ecology. We therefore replaced the terrain, resources, processing pathways, environmental fields, and functional objectives with a distinct volcanic materials world (AshenRealm), while retaining the same underlying agent--world interaction architecture, persistent-artifact mechanism, and provenance framework. The $72\times54$-cell environment spatially separates lava channels, obsidian wastes, sulfur marshes, iron mountains, magma seas, ash plains, a forge enclave, and a proving ground (Figure \ref{fig:ashenrealm_comparison}). Agents must locate and transform volcanic feedstocks through metallurgical operations, construct persistent technologies, and maintain function under spatially varying thermal, ash, and seismic hazards. Thus, AshenRealm changes both the accessible design space and the physical consequences of construction while preserving the mechanisms through which agents explore, interact, and modify a shared world.

The resulting societies did not simply reproduce the numerical behavior of BioFoundry. AshenRealm operated in a different functional regime, with lower absolute best-artifact performance, discovery-frontier AUC, behavioral novelty, resilience, and invention counts, but retained substantial multi-agent construction, cross-agent program reuse, executable lineage formation, and spatial exploration (Figure~\ref{fig:ashenrealm_comparison}). Label-blind behavioral analysis likewise recovered artifact-centered work and mobile exploration, together with an additional stationary/low-activity state (Figure~\ref{fig:behavior_phenotype_ashen_world}). The recurrence of these organizational modes under a different resource topology and consequence layer suggests that the observed differentiation is not specific to the original BioFoundry landscape. Because the two worlds encode different resources, objectives, and functional scales, however, their raw performance values should not be interpreted as a matched comparison of task difficulty.

\begin{figure}[htbp]
    \centering
    \includegraphics[width=0.97\textwidth,height=0.70\textheight,keepaspectratio]
    {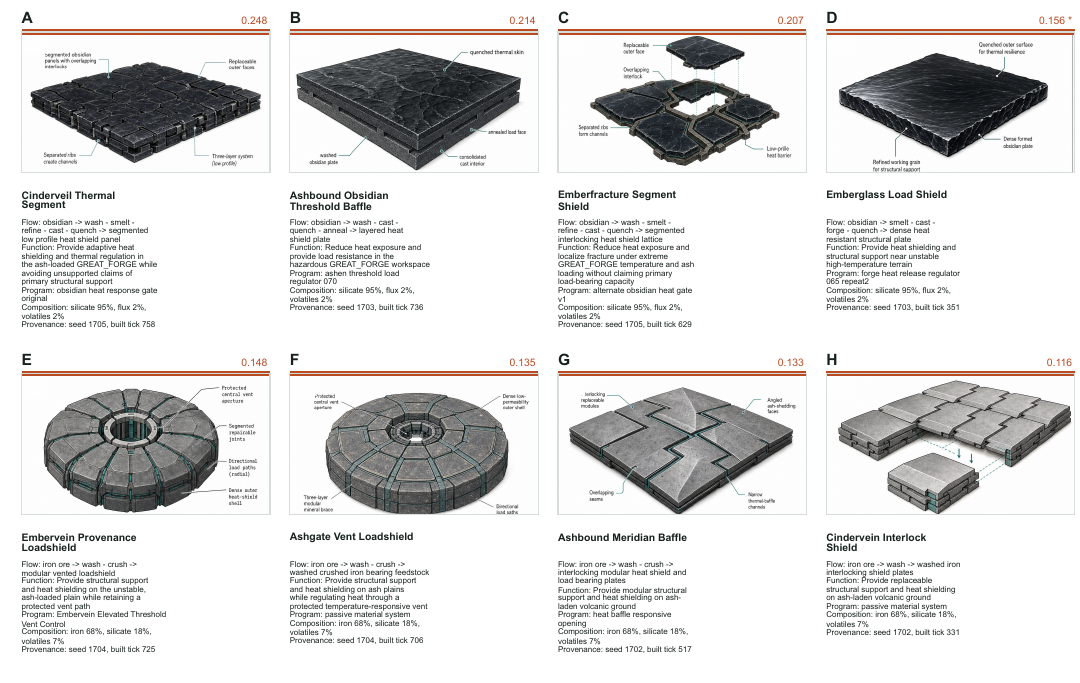}
    \caption{Representative agent-invented technologies in AshenRealm with full communication between agents (N=50). Eight performance-ranked technologies drawn across AshenRealm world seeds illustrate the material and functional design space generated by the agents. (A--D) The Cinderveil Thermal Segment, Ashbound Obsidian Threshold Baffle, Emberfracture Segment Shield, and Emberglass Load Shield are predominantly obsidian-based thermal and structural systems with lifetime-peak simulator performances of 0.248, 0.214, 0.207, and 0.156, respectively. (E--H) The Embervein Provenance Loadshield, Ashgate Vent Loadshield, Ashbound Meridian Baffle, and Cindervein Interlock Shield are iron-rich load-bearing and heat-management systems with performances of 0.148, 0.135, 0.133, and 0.116. Each panel reports the recorded material-processing pathway, agent-authored functional description, installed controller program, dominant composition, world seed, and construction tick. The renderings visualize the recorded architecture and proposed operating mechanism; they are not literal simulator meshes or experimentally manufactured structures. All quantitative performance values are obtained from the deterministic simulator.}    
\label{fig:ashenrealm_tech}
\end{figure}

The technologies themselves changed with the world. Performance-ranked AshenRealm artifacts include obsidian thermal segments, threshold baffles, fracture-localizing shields, and iron-based loadshields and vented interlocks, with lifetime-peak simulator performance ranging from 0.116 to 0.248 (Figure~\ref{fig:ashenrealm_tech}). The top four artifacts are predominantly obsidian based, and employ multi-step processing (see Figure \ref{fig:matter_pathway_ashen_world}). These designs arose across multiple world seeds and combined distinct processing histories, compositions, functional claims, and executable control programs. In a representative trajectory, persistent technology accumulated from no artifacts initially to 2 at tick 400 and 20 at tick 800, while best-artifact and portfolio performance increased in stages and the population became progressively more spatially localized (Figure~\ref{fig:world_evolution_ashen_world}). Together, these results provide a transfer test of the SwarmWorld substrate: the same geospatial and persistent-world architecture can support collective construction, executable inheritance, and behavioral differentiation under a materially distinct discovery problem.

\subsection{The transferable world architecture supports sequence-defined protein biomaterials}

\begin{figure}[htbp]
    \centering
    \includegraphics[width=0.97\textwidth,height=0.80\textheight,keepaspectratio]
    {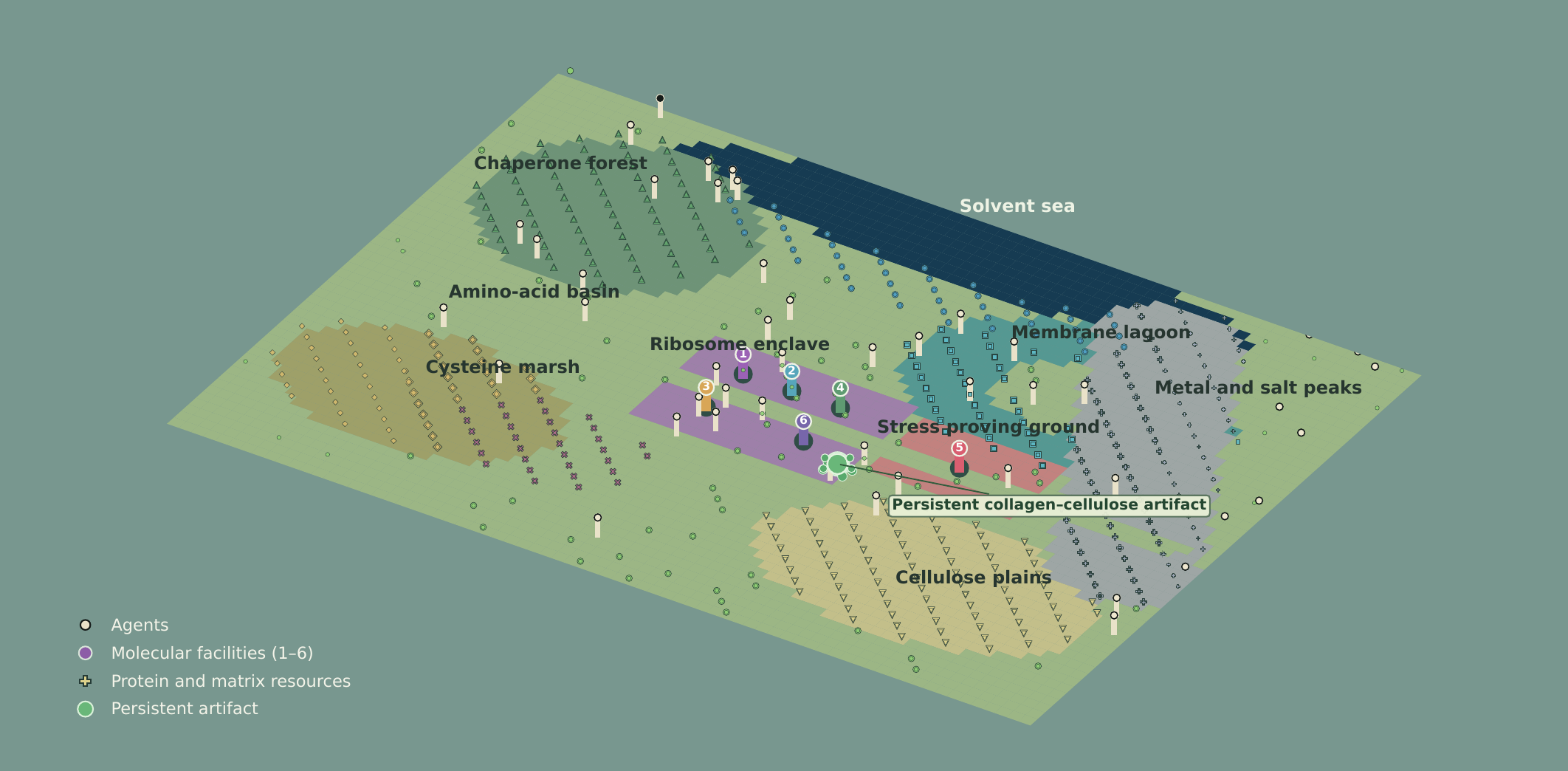}
    \caption[Protein Realms spatial protein--biomaterial world]{Production state of the Protein Realms no-communication society at tick 800 ($72\times54$ traced cells, $N=50$, seed 3801). The four-sided isometric plane is the complete traced rectangular terrain, containing a solvent sea, amino-acid basin, membrane lagoon, chaperone forest, cysteine marsh, cellulose plains, metal-and-salt peaks, ribosome enclave, and stress proving ground. Patterned glyphs mark a deterministic spatial sample of the trace's molecular-feedstock cells for legibility. Numbered markers identify the ribosome array, purification column, matrix loom, folding chamber, stress assay, and sequence archive; white circles are final agent positions. The green marker is the persistent collagen--cellulose installation constructed at tick 741. Separating feedstocks, processing facilities, and the proving ground forces designs to move through an executable spatial workflow rather than being scored directly from language.}
    \label{fig:protein-realms-world}
\end{figure}

AshenRealm established that the persistent-world architecture could support collective construction after replacing BioFoundry's material ecology, processing routes, and hazards. We next asked whether the same architecture could reach a design space organized around protein sequences and biological matrices rather than minerals and metallurgy. We therefore applied the declarative world builder to create Protein Realms, a $72\times54$ molecular landscape in which geography constrains protein--matrix design (Figure~\ref{fig:protein-realms-world}). This is a catalog-conditioned feasibility world rather than unrestricted de novo sequence generation. The package exposes 12 score-blind, versioned variants spanning collagen-like, silk--elastin-like, resilin-like, and mussel-adhesive families, including literature-inspired repeat motifs~\cite{pdb1a3j_collagen,teng2009selp,li2011resilin,das2015mussel}. Agents receive sequence, family, and permitted parent-lineage information, but not the hidden normalized property, matrix-affinity, or stress-resistance profiles. These hidden profiles are fixed simulator priors rather than biochemical predictions or new measurements. Agents must instead explore spatially separated amino-acid, membrane, chaperone, cysteine, cellulose, and mineral regions; collect peptide precursor, lipid, chitin, cellulose, mineral, buffer, chaperone, and crosslinking resources; and locate the ribosome array, purification column, folding chamber, matrix loom, sequence archive, and stress assay.

A legal Protein Realms design pairs one catalog sequence with at least one matrix and a peptide fraction between 10 and 60\% by mass. Agents select an ordered processing route, fabricate a quarter-scale microbatch, test it at the spatially separate proving ground, and rebuild the exact tested recipe as a persistent installation. The deterministic consequence layer evaluates the protein profile, matrix composition, sequence--matrix affinity, protein fraction, and executed process state rather than the artifact's name or claimed function. Installed biomaterials can then be repaired but not dismantled, so their function and degradation remain part of the evolving world.

\begin{figure}[htbp]
    \centering
    \includegraphics[width=0.97\textwidth,height=0.80\textheight,keepaspectratio]
    {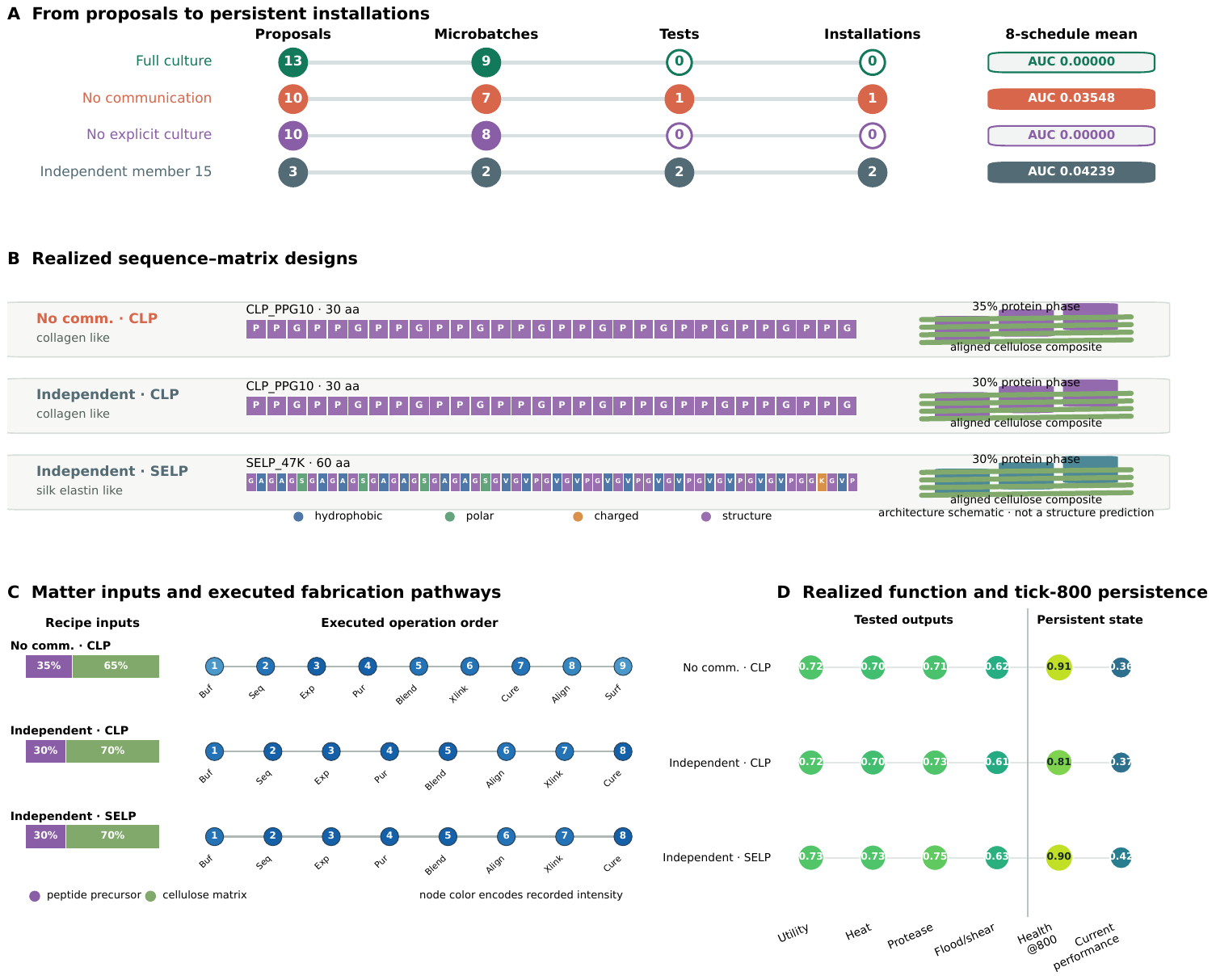}
    \caption[Protein Realms preliminary outcomes]{Trace-grounded designs and outcomes from the single-seed, 800-tick Protein Realms pilot. (A) Stage counts from proposals through microbatches, tests, and persistent installations across all four conditions; endpoint labels give mean resilience AUC across eight frozen, agent-free disturbance schedules. Independent member 15 is the selected member of the predeclared best-of-50 isolated-search envelope and represents one 16-call world, not another shared $N=50$ society. (B) The three realized installations. Colored sequence strips reproduce the exact retained catalog sequences by residue class; the adjacent protein-in-cellulose drawings are explanatory architecture schematics, not molecular-structure predictions. (C) Recorded recipe inputs and executed fabrication pathways. Bar widths give input mass fractions, numbered nodes preserve operation order, and node color encodes recorded intensity. (D) Realized consequences: normalized tested utility and heat, protease, and flood/shear resistance, followed by health and current performance at tick 800. These are simulator outputs from fixed catalog priors and executed recipes, not biochemical measurements. Only no communication and the selected independent member completed the hidden spatial assay gate; full culture and no explicit culture therefore remain visible in panel A but have no invented design rows. The pilot demonstrates an executable sequence-to-persistent-material workflow but does not support inferential comparison among conditions.}
    \label{fig:protein-realms-results}
\end{figure}

This workflow produced three trace-grounded persistent biomaterials within 800 ticks (Figure~\ref{fig:protein-realms-results}). The no-communication society installed a collagen-like CLP\_PPG10 phase in a cellulose-rich matrix at tick 741; at tick 800 it retained health 0.906 and performance 0.363 and achieved mean held-out resilience AUC 0.03548 across eight frozen disturbance schedules. Selected independent member 15 produced CLP\_PPG10 and silk--elastin-like SELP\_47K cellulose composites at ticks 388 and 738. Their retained records expose the actual sequence, material fractions, processing order, tested properties, and persistent state. SELP\_47K was the strongest installed design, with utility 0.729; normalized heat, protease, and flood/shear resistance of 0.726, 0.754, and 0.634; and tick-800 health 0.904 and performance 0.418.

Full culture and no explicit culture generated 13 and 10 proposals, respectively, but did not complete a valid assay within the horizon. The pilot used $N=50$, discovery seed 3801, a fixed decision schedule, and an 800-call budget per condition, and all 53 traces completed without provider errors and replayed deterministically. The independent endpoint is the selected best of 50 isolated 16-call worlds rather than another shared $N=50$ society. Within these descriptive, single-seed bounds, Protein Realms demonstrates that the same spatial and persistent-world architecture used for BioFoundry and AshenRealm can execute sequence selection, protein--matrix processing, assay-gated construction, continued material persistence, and agent-free evaluation in a sequence-defined biomaterials domain.

\section{Conclusion}

The experiments support a bounded form of swarm advantage: interaction is most valuable when performance depends on building and maintaining a technological ecology, not when the objective is only to find one record-setting object. Across the 800-tick scaling study, shared worlds produced consistent gains in held-out resilience, portfolio resilience, and validated inventions relative to the endpoint-wise independent-search envelope (Figure~\ref{fig:collective-results}). At 3,200 ticks, both shared-world conditions retained stronger portfolios and more validated inventions, while isolated search retained the strongest single final artifact (Figure~\ref{fig:long-horizon-tradeoff}). The central distinction is therefore between collective coverage and individual optimization. A society can be better prepared across several needs without containing the single best specialist technology.

Emergence appeared in organization as well as outcome. No agent was assigned to be an explorer, builder, maintainer, or cultural broker, yet label-blind analyses recovered artifact-centered and mobile-exploration phenotypes (Figures~\ref{fig:emergent-organization} and~\ref{fig:long-horizon-behavior}). Under explicit culture, the artifact-centered long-horizon fraction increased by 21.8 percentage points, movement declined relative to the ablation, regional crowding increased, and executable lineage depth continued to grow (Figures~\ref{fig:dynamics-lineage} and~\ref{fig:long-horizon-behavior}). Multi-agent construction and cross-agent program forking further show that the emergent organization entered persistent objects and executable controllers rather than remaining a pattern in language alone (Figure~\ref{fig:executable-culture}).

The shared physical world acted as an external memory and transmission medium. Agents with similar total path lengths differed strongly in artifact contact (Figure~\ref{fig:trajectory-atlas}), and event-linked paths show movement becoming observation, construction, installation, and coordination around persistent sites (Figure~\ref{fig:speed-action-atlas}). Recorded provenance connects evidence and programs to downstream artifacts (Figure~\ref{fig:knowledge-lineage}), while the material-process analysis separates proposed form from executed construction and operational flux (Figure~\ref{fig:material-process-flows}). Most importantly, technology reuse was nearly universal and faster under full culture, but approximately 95\% of first adoption occurred through physical observation and direct inventor-to-adopter social contact was not consistently enriched against the shuffled null (Figure~\ref{fig:cultural-diffusion}). Culture therefore operated diffusely: explicit exchange reshaped the society-wide substrate, after which much transmission occurred through encounters with the world itself.
In evolutionary terms, variation occurred in design and code space through authorship, forking, and edits, while simulator dynamics imposed differential functional consequences at the locations where technologies ran.
The emerging picture is therefore closer to technological evolution than to conventional multi-agent coordination: intelligence is distributed not only across agents, but across agents, persistent artifacts, executable lineages and a world whose accumulated modifications reshape the search space encountered by subsequent activity.

The mature society was neither a globally synchronized collective nor a collection of independent agents. It formed persistent, modular technological neighborhoods with a limited set of local hubs and cross-community connectors (Figures~\ref{fig:temporal-circuitry} and~\ref{fig:network-science}). Agent-artifact ties densified superlinearly, full culture produced roughly twice as many cumulative ties, and interval-to-interval relationship reuse rose in both shared worlds (Figure~\ref{fig:dynamic-network-assembly}). This organization created distributed redundancy: random removal of half the agents left nearly all artifacts connected to at least one survivor. It also created concentration: targeted removal of high-degree agents or brokers caused much larger losses (Figure~\ref{fig:structural-robustness}). Self-organization thus produced both robustness and a recognizable failure mode.

Longer time did not convert every cultural mechanism into a universal advantage. The crossover depended on the measured outcome: best-artifact performance favored full culture by about tick 800, portfolio resilience and artifact production crossed near tick 1,600, invention count never crossed, and held-out resilience showed no persistent full-culture advantage (Figure~\ref{fig:long-horizon-crossover}). No explicit culture remained highly capable because artifact stigmergy preserved a powerful form of decentralized coordination. This result is scientifically important because it identifies the shared world, rather than communication alone, as a principal source of collective capability. It also shows that adding cultural channels changes what the population learns to organize around, but can introduce coordination costs and structural dependence without guaranteeing better function on every endpoint.

The claims should remain proportional to the evidence. Inference is based on four matched world seeds per condition, one model and prompting configuration, and simulator-defined material and environmental functions. The technology portraits in Figure~\ref{fig:technology-gallery} are mechanism visualizations rather than experimentally manufactured geometries, and the knockout assay in Figure~\ref{fig:structural-robustness} measures graph topology rather than physical recovery after agents disappear from a running world. Within those boundaries, the study establishes a reproducible experimental path for studying language-model swarms as evolving societies: measure not only final answers, but also movement, durable artifacts, executable inheritance, recorded provenance, diffusion, network memory, and failure under controlled intervention.

The proposal-consequence separation also charts the path outward, because the consequence layer can in principle be replaced while retaining the same measurement framework, extending earlier work that emphasized the importance of the builder-breaker philosophy~\cite{wang2026selfrevisingdiscoverysystemsscience}. A first step keeps the world \textit{in silico} and raises its fidelity: coupling recipes, geometries, and controllers to established atomistic and continuum solvers would move simulator-defined function toward real material behavior, extending physics-aware agentic design~\cite{ghafarollahi_automating_2025,ghafarollahi_autonomous_2026,wang2026selfrevisingdiscoverysystemsscience,ghafarollahi_protagents_2024}. A second step grounds sensing and disturbance in reality: held-out schedules drawn from measured environmental records, and artifact controllers executed on embedded hardware against live sensor streams, would test whether stigmergic societies remain capable when the world is measured rather than generated. The third step is embodiment: robotic platforms and autonomous laboratories can supply the construction, testing, and observation primitives that the simulator currently resolves~\cite{brambilla_swarm_2013,strobel_llm2swarm_2024,ji_genswarm_2026}, so that persistent artifacts become physical objects in shared workspaces and stigmergy operates through the actual environment; our experimentally validated protein-design swarm illustrates one route from agent-swarm design to experimental testing~\cite{wang_swarms_2025,wang2026selfrevisingdiscoverysystemsscience}, and distributed artifact exchange across laboratories offers a route to scale it~\cite{wang_autonomous_2026}. Because the independent-search envelope, the agent-free held-out assay, provenance, executable lineage, and knockout analysis provide a common evaluation template across consequence layers, each of these substitutions can be evaluated against the same falsifiable criteria established here. Heterogeneous model populations, agent mortality and reproduction, resource economies, and human participants acting in the same persistent world define complementary axes along which the boundaries of the swarm advantage can be mapped.

\section{Materials and Methods}

\subsection{Overall algorithm}

SwarmWorld separates cognition from physical execution (see Algorithm~\ref{alg:swarmworld}). At fixed, staggered macroturns, each agent receives only its local observation, private memory, and whatever shared records are permitted by the experimental condition. The LLM returns a strictly validated research-state update and a plan of up to $L$ atomic actions. These actions enter a queue, and only one action per agent is attempted during each simulator tick. Consequently, agents can formulate multistep strategies without requiring an LLM call at every movement or physical operation.

The deterministic simulator checks every attempted action against spatial, material, energetic, and treatment-specific constraints before changing the world. It then advances environmental fields and source--sink-accounted resources and executes every installed artifact program. Successful and failed outcomes become new experience, while communication, publication, program inheritance, and artifact visibility depend on the experimental condition. Neither roles nor recipes are assigned by the simulator, model weights remain fixed, and observer-computed performance metrics are not supplied to the agents as a global reward.

At each declared evaluation checkpoint, the world and its technological portfolio are frozen and all agents are removed. Identical copies are exposed to paired, previously unseen disturbance schedules, during which only deterministic physics and installed artifact programs operate. This measures whether the accumulated technology remains functional without continued LLM intervention.

\begin{algorithm}[htbp]
\caption{SwarmWorld discovery and agent-free evaluation}
\label{alg:swarmworld}
\begin{algorithmic}[1]
\Require Population $N$; discovery horizon $T$; macroturn interval $m$;
plan limit $L$; condition capabilities $\mathcal{C}$; evaluation checkpoints
$\mathcal{H}$; held-out disturbance schedules $\mathcal{D}$
\Ensure Event trace, discovery metrics, and held-out resilience measurements
\State $W\gets\Call{GenerateWorld}{s_{\mathrm{disc}}}$
\State $\{x_i,M_i,Q_i,\phi_i\}_{i=1}^{N}
  \gets\Call{InitializeAgents}{N,W}$
\Comment{$x_i$: agent state; $M_i$: private memory; $Q_i$: action queue;
$\phi_i$: macroturn phase}
\State $\mathcal{A}\gets\emptyset$; $\mathcal{P}\gets\emptyset$;
$\mathcal{K}\gets\emptyset$
\Comment{$\mathcal{A}$: artifacts; $\mathcal{P}$: programs;
$\mathcal{K}$: permitted shared records}
\For{$t=0,\ldots,T-1$}
  \State $\mathcal{I}_t\gets\{i:(t-\phi_i)\bmod m=0\}$
  \Comment{fixed, treatment-invariant macroturn schedule}
  \ForAll{$i\in\mathcal{I}_t$ \textbf{in parallel}}
    \State $o_i\gets\Call{ObserveLocalWorld}{x_i,W,\mathcal{A},\mathcal{C}}$
    \State $r_i\gets\Call{RetrieveContext}
      {M_i,\mathcal{K},\mathcal{P},o_i,\mathcal{C}}$
    \State $(z_i,\widehat Q_i)\gets
      \Call{LLMStructuredPlan}{o_i,r_i,\mathcal{C},L}$
    \Comment{$z_i$: research-state update; $\widehat Q_i$: proposed action plan}
    \State $\widehat Q_i\gets\Call{ValidateSchema}{\widehat Q_i,\mathcal{C}}$
    \State $M_i\gets\Call{StoreResearchState}{M_i,z_i}$
    \State $Q_i\gets\widehat Q_i$
    \Comment{commit the validated plan as a bounded action queue}
  \EndFor
  \ForAll{active agents $i$}
    \If{$Q_i\neq\emptyset$}
      \State $u_i\gets\Call{PopFront}{Q_i}$
    \Else
      \State $u_i\gets\mathrm{WAIT}$
    \EndIf
  \EndFor
  \State $E_t^{\mathrm{act}}\gets
    \Call{ResolveAndApply}
      {\{u_i\}_{i=1}^{N},W,\mathcal{A},\mathcal{P},\mathcal{K},\mathcal{C}}$
  \Comment{enforce spatial, material, energetic, permission, and action constraints}
  \State $(W,E_t^{\mathrm{env}})\gets\Call{AdvanceEnvironment}{W,t}$
  \State $(W,\mathcal{A},E_t^{\mathrm{art}})\gets
    \Call{ExecuteArtifactPrograms}{W,\mathcal{A},\mathcal{P},t}$
  \Comment{persistent controllers execute on every simulator tick}
  \State $E_t\gets E_t^{\mathrm{act}}\cup E_t^{\mathrm{env}}
    \cup E_t^{\mathrm{art}}$
  \State $(\{M_i\},\mathcal{K})\gets\Call{CommitEvents}{E_t,\mathcal{C}}$
  \State \Call{AppendTrace}
    {$t,W,\{x_i,M_i,Q_i\}_{i=1}^{N},\mathcal{A},\mathcal{P},\mathcal{K},E_t$}
  \If{$t+1\in\mathcal{H}$}
    \State $F_{t+1}\gets\Call{FreezeAndRemoveAgents}{W,\mathcal{A},\mathcal{P}}$
    \ForAll{$d\in\mathcal{D}$}
      \State $F_{t+1,d}\gets\Call{CloneAndStress}{F_{t+1},d}$
      \State $R_{t+1,d}\gets\Call{EvaluateAgentFree}{F_{t+1,d}}$
      \Comment{only deterministic physics and installed programs continue}
    \EndFor
  \EndIf
\EndFor
\State \Return trace, discovery metrics, and
$\{R_{h,d}:h\in\mathcal{H},\ d\in\mathcal{D}\}$
\end{algorithmic}
\end{algorithm}

\subsubsection{Agent-to-world interface}

Agents observe a semantic description of their local neighborhood and return a closed, schema-validated research-state update and action plan. Private memory stores recent outcomes, tested recipes, notebook evidence, and the agent's current hypothesis and next checkpoint; treatment-permitted records supply the only symbolic information from other agents. Runtime validation enforces location, ownership, empirical grounding, capacity, and condition permissions, and invalid output becomes a logged safe wait. Representative retained messages and actions and the complete interface specification are provided in Section~\ref{sec:si-agent-communication}.

\subsection{World representation and constructing custom worlds}
\label{section:world-representation}

Each simulation world is represented as a rectangular two-dimensional lattice with typed terrain, resource, facility, and continuous environmental-field layers. A world instance is generated as
\[
W_s = G(s;\boldsymbol{\theta}),
\]
where $s$ is a random seed and $\boldsymbol{\theta}$ specifies dimensions, spatial-generation rules, resource capacities and renewal rates, environmental dynamics, facility-placement constraints, and disturbance processes. Generation is deterministic for fixed $s$ and $\boldsymbol{\theta}$. Candidate layouts are validated against structural invariants, including sufficient walkable area, availability of required resource classes, nonoverlapping facilities, and reachable interaction regions. Invalid candidates are rejected and regenerated deterministically. The resulting arrays and complete generator manifest are recorded in the simulation trace, allowing the initial world to be reconstructed exactly.

The world builder also supports declarative scenario packages that replace or extend terrain, resource, facility, processing, field-dynamics, and service definitions without modifying the simulation loop. Packages are specified through validated YAML documents and mapped onto the simulator's stable typed interfaces; package-authored executable code is not loaded. Alternative procedural generators may therefore be used provided that they produce the same authoritative state representation and satisfy the declared physical invariants. Visual presentation is kept separate from simulation state: terrain meshes, object assets, textures, lighting, particles, and shader parameters may be assigned by a renderer, but they cannot alter physics, resource quantities, agent observations, action legality, or measured outcomes. Thus, the same simulated world may be rendered at different levels of visual fidelity while remaining scientifically identical. Further details are provided in Section~\ref{sec:extended-world-building}.

\subsection{Materials, artifacts, and interaction mechanisms}

Agents collect source--sink-accounted feedstocks, formulate and test processing recipes, and build persistent material systems whose measured behavior is determined by numerical material state, geometry, environment, stored matter, health, and optional bounded controller programs. Communication, publication, teaching, trade, task claims, program reuse, and program forking are independently controlled by the experimental condition, while the shared physical world can retain stigmergic traces in every condition except isolated search. Equations, controller limits, source--sink accounting, and provenance rules are given in Section~\ref{sec:si-materials-artifacts}.

\subsection{Experimental design and analysis}

The completed 800-tick study crossed four interaction conditions, three population sizes ($N=50,100,200$), and four matched discovery seeds, yielding 48 episodes. A separate completed 3,200-tick study compared full culture, no explicit culture, and independent search at $N=100$ over four new seeds. At declared checkpoints, frozen portfolios were evaluated without agents under eight paired held-out disturbance schedules. Full condition definitions, controls, run provenance, and checkpoint designs are provided in Section~\ref{sec:si-experimental-design}.

Discovery, held-out resilience, portfolio breadth and redundancy, validated inventions, behavior, lineage, and network structure were computed from immutable traces and frozen evaluation copies. The independently generated world seed was the unit of inference; conditions were compared within matched population--seed blocks using deterministic paired bootstrap intervals, supplemented by exact sign-flip tests for the long-horizon study. Endpoint definitions and analysis procedures are provided in Section~\ref{sec:si-analysis}, and release validation and reproducibility details in Section~\ref{sec:si-reproducibility}.

\section*{Data and code availability}
Code, exact prompt manifests, versioned protocols, raw model outputs, analysis scripts, figure-generation code, and manuscript sources are available at \url{https://github.com/lamm-mit/SwarmWorld}. Additional data from the experiments conducted as part of this paper is available at \url{https://huggingface.co/datasets/lamm-mit/swarmworld-data}.

\subsection*{Funding}
This work was primarily supported by the U.S. Department of Energy, Office of Science, Office of Advanced Scientific Computing Research and Office of Basic Energy Sciences, Scientific Discovery through Advanced Computing (SciDAC) program under the FORUM-AI project. Additional support was provided by MIT's Generative AI Impact Consortium (MGAIC). 

\section*{Author contributions}
M.J.B. conceived and led the study, developed the methodology, conducted the computational work and primary analysis, interpreted the results, and drafted the manuscript. S.P. and F.Y.W. contributed to data analysis, additional computational experiments, and manuscript writing and editing. All authors reviewed and approved the final manuscript.

\section*{Competing interests}
The authors declare no competing interests.

\bibliographystyle{naturemag}

\bibliography{references}

\newpage

\setcounter{page}{1}
\renewcommand{\thepage}{S\arabic{page}}
\setcounter{section}{0}
\renewcommand{\thesection}{S\arabic{section}}
\setcounter{table}{0}
\renewcommand{\thetable}{S\arabic{table}}
\setcounter{figure}{0}
\renewcommand{\thefigure}{S\arabic{figure}}
\setcounter{equation}{0}
\renewcommand{\theequation}{S\arabic{equation}}

\begin{center}
  {\LARGE \textbf{Supplementary Information}\\[1.5em]
  \textbf{SwarmWorld: Stigmergic technological evolution in societies of language-model agents}\\}
  \vspace{1cm}
  
  {\large
  \orcidlink{0000-0002-0532-9791} Subhadeep Pal\textsuperscript{1,2} \quad 
  \orcidlink{0000-0002-4847-6986} Fiona Y. Wang\textsuperscript{1,3} \quad
  \orcidlink{0000-0002-4173-9659} Markus J. Buehler\textsuperscript{1,2,4,5,\#} \\
  }
  \vspace{0.5cm}
  
  {\normalsize
  \textsuperscript{1}Laboratory for Atomistic and Molecular Mechanics (LAMM)\\
  \textsuperscript{2}Department of Civil and Environmental Engineering, \\
  \textsuperscript{3}Department of Biological Engineering\\
  \textsuperscript{4}Department of Mechanical Engineering, \\
  \textsuperscript{5}Center for Computational Science and Engineering, \\
  Schwarzman College of Computing, \\
  Massachusetts Institute of Technology, Cambridge, MA 02139, USA \\
  \vspace{0.5cm}
  \textsuperscript{\#}Corresponding author: \texttt{mbuehler@mit.edu}
  }
\end{center}

\newpage

\section{Glossary of key terms}
\label{glossary_section}

Table~\ref{tab:glossary} defines the quantities used throughout the paper. Equations are included only where the implementation computes a corresponding mathematical quantity; terms that describe an intervention, object, or event are defined operationally.

\begingroup
\footnotesize
\setlength{\tabcolsep}{4pt}
\renewcommand{\arraystretch}{1.25}
\begin{longtable}{@{}>{\raggedright\arraybackslash}p{0.15\textwidth}>{\raggedright\arraybackslash}p{0.32\textwidth}>{\raggedright\arraybackslash}p{0.22\textwidth}>{\raggedright\arraybackslash}p{0.22\textwidth}@{}}
\caption{Glossary of technical terms, implementation-aligned definitions, plain-language interpretations, and broadly accessible examples.}
\label{tab:glossary}\\
\toprule
Term & Technical definition & Plain-language meaning & Broad example \\
\midrule
\endfirsthead
\multicolumn{4}{@{}l}{Table~\thetable\ continued}\\
\toprule
Term & Technical definition & Plain-language meaning & Broad example \\
\midrule
\endhead
\midrule
\multicolumn{4}{r@{}}{Continued on next page}\\
\endfoot
\bottomrule
\endlastfoot

World tick
& One discrete deterministic state transition, $S_{t+1}=F(S_t,A_t)$, where $S_t$ is the complete world state and $A_t$ is the set of accepted agent actions at tick $t$. Given the state, actions, and seeded disturbance schedule, $F$ advances movement, metabolism, resources, fields, artifacts, and installed programs.
& The basic clock step of the simulated world. The world keeps changing even when no agent is currently asking the language model what to do.
& A video game advances one frame at a time; machines, weather, and moving characters continue between a player's major decisions. \\

Macroturn or decision opportunity
& A scheduled occasion on which one agent receives a local observation and memory context, calls the language model, and proposes a structured multi-action plan. Macroturn phases are staggered and matched across paired conditions.
& One chance for an agent to stop, look around, think, and choose its next actions.
& A field scientist periodically checks instruments and updates the day's plan while the experiment continues running between checks. \\

Persistent artifact
& An agent-constructed world object with recorded creator, contributors, materials, geometry, services, provenance, state, and optionally an executable controller. It remains in the world until modified, retired, dismantled, or otherwise changed by simulator rules.
& A technology that stays behind and can later be found, tested, operated, or modified by someone else.
& A device left on a shared laboratory bench can influence the next researcher even if its builder is absent. \\

Artifact program
& A validated instruction sequence installed on an artifact and executed by the deterministic simulator every tick. Conceptually, $u_{a,t}=\pi_a(o_{a,t})$: program $\pi_a$ reads the artifact's permitted local sensors $o_{a,t}$ and emits permitted control operations $u_{a,t}$.
& Code that lets a built object sense local conditions and act without another language-model call.
& A thermostat repeatedly reads temperature and turns heating on or off after the installer leaves. \\

Full culture
& The intervention containing one shared physical world, explicit cultural actions, cross-agent executable-program inheritance, and artifact stigmergy. Explicit actions include messages, public records, teaching, trading, task claims, and publication-dependent composition.
& Agents can communicate directly, inherit one another's code, and also coordinate indirectly through the things they build.
& A research group shares a laboratory, talks, writes notebooks, and modifies common software and equipment. \\

Artifact stigmergy
& Indirect coordination produced when an agent changes persistent artifacts or the environment and another agent later observes or acts on that changed state. It remains available in the no-explicit-culture condition even when direct cultural channels and cross-agent program inheritance are disabled.
& Agents coordinate through traces in the world rather than through direct conversation.
& Ants coordinate through modified trails; people can likewise coordinate by adding parts or annotations to a shared workbench. \\

Independent-search envelope
& For endpoint $e$, checkpoint $t$, and $N$ isolated one-agent worlds, the control is $Y_{\mathrm{iso}}(e,t)=\max_{i\in\{1,\ldots,N\}}Y_i(e,t)$. The maximizing member may differ across endpoints and checkpoints.
& The comparison gives isolated search its best available result for each question, rather than forcing one solo agent to win every contest.
& A national team may choose one athlete for the sprint and another for the high jump; the envelope records the best specialist in each event. \\

Discovery-frontier AUC
& If $p(t)$ is measured artifact performance, the running frontier is $F(t)=\max_{\tau\leq t}p(\tau)$. The normalized area is $D_T=T^{-1}\int_0^T F(t)\,dt$, implemented by trapezoidal integration over recorded samples.
& A strong invention scores more if it appears early enough to remain the best-known option for much of the experiment.
& Two runners may finish with the same final speed, but the one who led for most of the race has the larger time-averaged frontier. \\

Current service coverage
& For service dimension $k$ at time $t$, $c_k(t)=\max_{a\in\mathcal{A}_t}s_{ak}(t)$ over active, non-retired artifacts. With $\bar c(t)=K^{-1}\sum_k c_k(t)$ and $b(t)=\min_k c_k(t)/\bar c(t)$, balanced coverage is $Q(t)=\bar c(t)[1+b(t)]/2$ when $\bar c(t)>0$.
& The society is judged by the best currently available artifact for each need, with a penalty when one need is badly neglected.
& An emergency kit with food, water, shelter, and medicine is more balanced than one containing excellent food but nothing else. \\

Portfolio resilience
& The discovery-state score uses lifetime-peak service fingerprints, $c_k^{\mathrm{peak}}=\max_a s_{ak}^{\mathrm{peak}}$, and the same balance formula $P=\bar c^{\mathrm{peak}}[1+\min_k c_k^{\mathrm{peak}}/\bar c^{\mathrm{peak}}]/2$. Artifact count alone does not increase the score.
& A portfolio is valuable when its collection has demonstrated strong and balanced coverage across several functions, not merely because it contains many objects.
& A toolbox with a few complementary, proven tools can be more resilient than a warehouse full of duplicate hammers. \\

Held-out resilience
& For unseen schedule $s$, $R_s=T^{-1}\int_0^T Q_s(t)\,dt$, where $Q_s(t)$ is current balanced service coverage during the agent-free assay. The reported value is $R=S^{-1}\sum_{s=1}^{S}R_s$ over $S=8$ paired schedules.
& The frozen technology is tested under new disturbances after all agents are removed; the score measures how well its active functions continue to cover multiple needs.
& A bridge design is evaluated under earthquakes and wind patterns that were not used while designing it, with no engineer allowed to repair it during the test. \\

Validated invention
& The count is $I=\sum_a \mathbf{1}[r_a\land d_a\land g_a\land p_a\geq p^*\land\nu_a\geq\nu^*]$. Artifact $a$ must use a recipe that passed material testing ($r_a$), contain the required design fields ($d_a$), include an agent-authored installed program ($g_a$), exceed the performance threshold $p^*$, and exceed the behavioral-novelty threshold $\nu^*$.
& A named idea is not enough. The artifact must be materially grounded, fully specified, executable, functional, and behaviorally distinct.
& A patent sketch would not count by itself; a device must have tested materials, a complete design, working controls, and a demonstrably new behavior. \\

Artifact-contact AUC
& Let $q(t)$ be the fraction of agents within three cells of an active artifact at recorded time $t$. Contact AUC is $C=(t_m-t_1)^{-1}\int_{t_1}^{t_m}q(t)\,dt$, evaluated by trapezoidal integration.
& The measure records how consistently agents spend time near shared technology, rather than how far they travel in total.
& A museum can measure the fraction of visitors near an exhibit throughout the day, not only the number who entered the building. \\

Behavioral phenotype
& A post hoc cluster of complete agent trajectories. For k-means, assignments minimize $\sum_i\lVert x_i-\mu_{z_i}\rVert^2$ in the declared robust-scaled feature space. Condition, identity, and role labels are excluded from fitting.
& A recurring style of behavior discovered from what agents actually did, not a role assigned in their prompts.
& Travel records might separate commuters who repeatedly visit one workplace from explorers who visit many neighborhoods, without reading their job titles. \\

Recorded lineage and downstream reach
& A directed graph whose edges come from recorded authorship, observation, causal-parent, construction, installation, and fork events. For node $v$, downstream artifact reach is $r(v)=|\{a: v\leadsto a,\ a\text{ is an artifact}\}|$.
& It records which evidence, programs, people, and precursor objects are linked by logged parent and contribution events to later technologies.
& A product genealogy can trace a component specification through several revisions into multiple final devices. \\

Cross-agent program fork and lineage depth
& A fork is cross-agent when the child-program author is not in the recorded parent-author set. Executable-lineage depth is the longest directed path in the content-addressed program-fork directed acyclic graph up to a checkpoint.
& One agent edits another agent's controller, and depth counts how many successive generations of such inherited code accumulate.
& A programmer forks a colleague's repository; later colleagues fork the modified versions, forming a software family tree. \\

Interaction modularity
& Weighted Newman-Girvan modularity, $Q=(2m)^{-1}\sum_{ij}[A_{ij}-k_i k_j/(2m)]\delta(g_i,g_j)$, computed on the weighted bipartite agent-artifact graph using Louvain communities at resolution 1. Event-type weights are multiplied by $\log(1+n)$ for repeated events.
& High modularity means interactions are concentrated within recognizable technological neighborhoods rather than spread uniformly across the whole society.
& A university is modular when laboratory members collaborate mostly within their labs, with fewer links between labs. \\

Community persistence
& Adjusted mutual information between consecutive agent community assignments, $\mathrm{AMI}=[\mathrm{MI}-\mathbb{E}(\mathrm{MI})]/[(H_U+H_V)/2-\mathbb{E}(\mathrm{MI})]$. Adjustment removes similarity expected by chance.
& The measure asks whether the same agents remain grouped together from one checkpoint to the next, allowing community labels themselves to change.
& If project teams retain most of the same members next semester, their community persistence is high even if team numbers are renamed. \\

Participation coefficient and within-module z-score
& For node $i$, $P_i=1-\sum_c(k_{ic}/k_i)^2$ measures how evenly its weighted links span communities. Within its own community, $z_i=(k_i^{\mathrm{own}}-\mu_c)/\sigma_c$ measures how unusually strong its internal connectivity is relative to peers of the same node type.
& Participation distinguishes local specialists from cross-community connectors; the z-score distinguishes ordinary members from local hubs.
& A scientist collaborating only within one lab is a specialist, one collaborating across many labs is a connector, and a highly connected member within one lab is a local hub. \\

NODF nestedness
& Binary nestedness is $100$ times the mean $|N_i\cap N_j|/\min(k_i,k_j)$ over unequal-degree pairs, evaluated for both agent rows and artifact columns. High values mean the smaller neighborhood is largely contained in the larger one.
& Nestedness asks whether specialists mostly use subsets of the technologies used by generalists.
& A small shop is nested within a department store if nearly everything it stocks also appears in the larger store. \\

Densification exponent
& Cumulative unique agent-artifact ties are fit to $E_{\mathrm{cum}}\propto V^{\alpha}$, or $\log E_{\mathrm{cum}}=\alpha\log V+c$, where $V$ is agents plus constructed artifacts. The plotted $\alpha$ is an ordinary least-squares descriptive fit across seeds and checkpoints for one condition.
& If $\alpha>1$, relationships accumulate faster than the number of participating agents and artifacts.
& A growing professional community densifies when adding people and projects creates disproportionately many new collaborations. \\

Relationship reuse
& For consecutive analysis intervals, reuse is $\rho_t=|E_{t-1}\cap E_t|/|E_t|$, where $E_t$ is the set of agent-artifact pairs active in the current interval. It is distinct from the symmetric Jaccard index.
& The measure is the fraction of current working relationships that were already active in the preceding period.
& A shop with many returning customers has high relationship reuse even if it also attracts new customers. \\

Cross-agent adoption and time to first reuse
& An artifact is adopted when a noncreator first observes, causally references, programs, or repairs it; co-construction does not count. Time to first reuse is $T_a=t_a^{\mathrm{first\ noncreator\ reuse}}-t_a^{\mathrm{created}}$. Unreused artifacts are right-censored at tick 3,200.
& Adoption means someone other than the inventor actually engages with the technology, and the delay measures how quickly that happens.
& A neighbor borrowing and using a tool counts as adoption; merely helping build it does not count as later reuse. \\

Structural robustness AUC
& If $L(f)$ is largest-component size after removing fraction $f$ of agents, robustness is $A=f_{\max}^{-1}\int_0^{f_{\max}}L(f)/L(0)\,df$ for $f_{\max}=0.5$. Removal is random, degree-targeted, or betweenness-targeted; the graph is not rewired.
& The score summarizes how much of the recorded network stays connected as agents disappear. It does not measure physical recovery or adaptation after removal.
& An internet topology may tolerate random router failures but fragment quickly if its busiest hubs are deliberately disabled. \\

\end{longtable}
\endgroup


\section{Supplementary Methods}
\label{sec:materials-and-methods}

\subsection{Agent model, communication, and model-to-world interface}
\label{sec:si-agent-methods}

Agents were homogeneous within and across conditions: each used \texttt{gpt-5.6-luna}, temperature 0.7, low reasoning effort, the same system prompt, strict action schema, initial capabilities, inventory limit, and memory budgets. The model endpoint used non-streaming Responses API requests with provider-side response storage disabled. The paper configuration allowed 4,096 output tokens, a maximum of 12 planned actions, a 60,000-character retrieved-context budget, and 64 private memory records. These limits bound context growth but do not assign roles or prescribe a scientific workflow.

An observation contains semantic facts about the agent's local neighborhood rather than global world arrays. It includes visible terrain, resources, facilities, agents, artifacts, environmental measurements, the agent's inventory, pending microbatches, and compact action affordances. A sparse empirical map records only previously observed locations and last-seen ticks. Private memory stores recent outcomes, notebook evidence, tested recipes, and the agent's latest model-authored research state: goal, hypothesis, progress assessment, next observable checkpoint, and collaboration need. Bounded retrieval ranks private and permitted public records using the current research state and empirical reuse signals. The simulator records this state but neither scores nor supplies its content.

The model must return one closed JSON object containing a research-state update and a plan. The provider is asked for strict JSON Schema output, and the decoded response is validated again locally. Runtime checks then enforce state-dependent preconditions such as location, ownership, empirical grounding, capacity, and treatment permissions. Invalid output becomes a logged safe wait and never reaches \texttt{eval}, \texttt{exec}, a shell, or the artifact virtual machine. Retryable transport or service failures are treated as infrastructure events: all queued actions and scheduling bits are preserved, the failed macroturn is retried, and world time does not advance. Nonretryable or structurally invalid model output is retained as a failed decision.

\subsection{World representation and constructing custom worlds}
\label{sec:si-world-overview}
\label{sec:extended-world-building}

Each simulation world is represented as a rectangular two-dimensional lattice with typed terrain, resource, facility, and continuous environmental-field layers. A world instance is generated as
\[
W_s = G(s;\boldsymbol{\theta}),
\]
where $s$ is a random seed and $\boldsymbol{\theta}$ specifies dimensions, spatial-generation rules, resource capacities and renewal rates, environmental dynamics, facility-placement constraints, and disturbance processes. Generation is deterministic for fixed $s$ and $\boldsymbol{\theta}$. Candidate layouts are validated against structural invariants, including sufficient walkable area, availability of required resource classes, nonoverlapping facilities, and reachable interaction regions. Invalid candidates are rejected and regenerated deterministically. The resulting arrays and complete generator manifest are recorded in the simulation trace, allowing the initial world to be reconstructed exactly.

All paper experiments used fixed world scaling: the lattice remained $72\times54$ cells as population increased, so population changed agent density rather than available area. Conditions sharing a discovery seed received the same procedural world and disturbance process. Initial positions were drawn as a deterministic nested permutation of walkable cells, making the first $N$ positions identical when a larger population was truncated to size $N$. Sampling the full permutation at reset prevents population size from shifting later simulator randomness.

\subsubsection{State representation and reproducibility}
The authoritative world state stores integer terrain, resource, and facility layers; floating-point resource mass and capacity layers; and named continuous environmental fields. Principal run-level parameters include grid dimensions, disturbance interval and intensity, resource and field capacities, and the optional \texttt{scenario\_package} path. Every trace records the resolved configuration, engine revision, realized generator manifest, scenario identifier and version, and a SHA-256 hash over \texttt{scenario.yaml}, each referenced package document, and the agent prompt. Replaying a trace therefore detects changes to the active world definition.

\subsubsection{Declarative geometry, resources, and facilities}
A scenario package maps stable internal slots to domain-specific public identifiers
for nine terrains, eight nonempty resources, six nonempty facilities, and ten process
operations. This preserves the observation and replay protocol while allowing the
same simulation engine to express a different materials setting. Grid coordinates are
normalized as $\xi=x/\max(1,w-1)$ and $\eta=y/\max(1,h-1)$. The geometry file first
assigns a base terrain and then applies ordered feature masks; later features overwrite
earlier ones. Implemented masks are circles, rings, ellipses, rectangles, finite-width
ridges, and seeded Bernoulli noise. Facilities are placed from normalized coordinates;
if a requested cell is not walkable, the facility is moved to the nearest walkable
cell. Resource deposits use the same shape operators and are restricted to walkable
cells. For deposit capacity $c$, fractional variation $v$, and initial-fill interval
$[f_0,f_1]$, each selected cell is initialized as
\begin{equation}
C_{xy}=cU(1-v,1+v),\qquad
M_{xy}(0)=C_{xy}U(f_0,f_1),
\end{equation}
and optional renewal follows
\begin{equation}
M_{xy}(t+1)=\min\!\left[C_{xy},M_{xy}(t)+
\rho_{xy}\{C_{xy}-M_{xy}(t)\}\right].
\end{equation}
Harvesting removes mass from this ledger, and later overlapping deposits replace
earlier deposit values at the affected cells.

\subsubsection{Environmental fields and disturbances}
Each named field declares a numerical range, diffusion and decay coefficients,
per-terrain sources, and an initial condition composed of a constant, spatial
gradients, radial Gaussian terms, terrain offsets, and optional seeded Gaussian noise.
Ignoring clipping notation, a noncyclic field is updated in the implemented order
\begin{equation}
F_k(t+1)=(1-\lambda_k)\left[F_k(t)+D_k\Delta_4F_k(t)\right]
+s_k(\tau_{xy}),
\end{equation}
where the four-neighbor Laplacian uses edge-value padding. A field marked as cyclic is
instead overwritten after these operations by a spatially uniform sinusoidal value
defined by its period, night level, and amplitude. Format version 1 requires the
compatibility fields \texttt{temperature}, \texttt{water\_availability},
\texttt{ground\_stability}, \texttt{toxic\_gas}, and \texttt{solar}; additional fields
may be declared and are retained in observations, snapshots, analysis, and rendering.
At every positive tick divisible by the configured disturbance interval, the next
declared disturbance is applied around a seeded walkable center using
\begin{equation}
H_{xy}=\alpha\exp\!\left[-\frac{(x-x_c)^2+(y-y_c)^2}
{2\max\{1,r\min(w,h)\}^2}\right],
\end{equation}
with field-specific deltas and an optional thresholded terrain transformation. Held-out
evaluation resamples disturbance centers and order from an evaluation seed while
leaving the frozen technological state unchanged.

\subsubsection{Package boundary and example}
Scenario packages are YAML data and cannot import Python, execute shell commands, or
modify the active package during an episode. They may define terrain colors and
heights, field-overlay colors, and aliases for common properties and artifact services.
The current implementation does not load package-supplied meshes, textures, or shader
source. Renderer assignments remain presentation-only and cannot alter physics,
resource quantities, observations, action legality, or measured outcomes. Package
loading checks containment, catalog completeness and unique IDs,
cross-references, required compatibility fields, and the default-recipe identifiers.
It does not by itself prove global reachability, facility nonoverlap, deposit
nonemptiness, or numerical field stability; these properties must be established by
scenario-specific tests. The following excerpt illustrates the division between the
run configuration, package manifest, and declarative layer files; a complete package
must additionally define every stable terrain, resource, facility, and operation slot.

\begin{lstlisting}[basicstyle=\ttfamily\scriptsize,breaklines=true]
# run.yaml
world:
  width: 48
  height: 36
  scenario_package: worlds/example_domain
  disturbance_interval: 64
  disturbance_intensity: 0.50
simulation:
  seed: 1701
  max_ticks: 800

# worlds/example_domain/scenario.yaml
format_version: 1
id: example_domain
name: Example Domain
version: 0.1.0
agent_prompt: prompts/agent_instructions.md
documents:
  terrains: terrain.yaml
  geometry: geometry.yaml
  fields: fields.yaml
  resources: resources.yaml
  operations: operations.yaml
  facilities: facilities.yaml
  artifacts: artifacts.yaml
  missions: missions.yaml
  disturbances: disturbances.yaml
  rendering: rendering.yaml
  analysis: analysis.yaml

# geometry.yaml: ordered layers in normalized coordinates
base_terrain: PLAIN
features:
  - terrain: REGION_A
    shape: {kind: ellipse, center: [0.25, 0.30], radius: [0.18, 0.14]}
  - terrain: CORRIDOR
    shape: {kind: ridge, start: [0.10, 0.55], end: [0.90, 0.68], width: 0.025}
  - terrain: WORKSPACE
    shape: {kind: rectangle, bounds: [0.42, 0.48, 0.60, 0.66]}

# resources.yaml: one example deposit
deposits:
  - resource: FEEDSTOCK_A
    shape: {kind: ellipse, center: [0.25, 0.30], radius: [0.17, 0.13]}
    capacity: 3.0
    capacity_variation: 0.15
    initial_fill: [0.65, 1.0]
    regrowth: 0.001

# fields.yaml: required fields use the same structure
fields:
  - id: temperature
    name: Temperature
    range: [0.0, 1.0]
    diffusion: 0.04
    decay: 0.001
    initial:
      constant: 0.25
      noise: 0.01
      radial:
        - {center: [0.75, 0.25], sigma: 0.16, amplitude: 0.35}

# disturbances.yaml
disturbances:
  - id: FIELD_PULSE
    name: Field pulse
    radius: 0.24
    field_deltas:
      temperature: 0.35
      toxic_gas: 0.28
      water_availability: -0.16
\end{lstlisting}

\subsection{Open-ended materials invention}
\label{sec:si-materials-artifacts}

Agents receive the mission to develop bioinspired material systems for environmental resilience. They may harvest only locally present matter, move conserved feedstock between personal inventories and shared depots when allowed, formulate typed processing recipes, operate distributed workstations, fabricate private microbatches, and test those batches. Testing reveals deterministic normalized properties only after fabrication. The action schema provides a representational material vocabulary, but a named input must be grounded by direct observation, personal possession, or exact cited evidence permitted by the treatment; requesting a label cannot override local physical state.

Material properties are transparent game-level surrogates, not SI-calibrated predictions. The evaluator combines composition, ordered processing, hydration, porosity, alignment, crosslinking, and quality into normalized stiffness, toughness, permeability, adhesion, healing, responsiveness, and degradation properties. Agents do not observe the evaluator equation or a global reward. They receive only the outcomes of their own admissible operations and tests.

The engine exposes a single generic artifact class, \texttt{MATERIAL\_SYSTEM}; it does not contain a catalog of membranes, lattices, scaffolds, or preferred biological analogies. An artifact specification contains an agent-authored name, claimed function, architecture, biological inspirations, predicted effects, continuous geometry, a tested material batch, and an optional controller. Text is retained for interpretation and provenance but does not change function. Numeric material state, geometry, local environment, artifact health, stored matter, and program actuation determine behavior.

Closed artifact fluxes enforce source--sink accounting. Water collection removes the same amount from the local field that enters storage; remediation cannot remove more contamination than exists; and growth, healing, repair, and nutrient release consume a bounded embodied reserve. Natural recharge, resource regrowth, disturbances, agent metabolism, and artifact transfers are recorded separately in a flux ledger. The paper configuration enabled closed fluxes and left metabolism, mortality, and replacement disabled.

\subsection{Persistent executable artifacts}

Agents can install a deterministic straight-line controller in a persistent artifact. A controller contains 1--64 instructions over 16 floating-point registers. Named sensors expose local moisture, nutrients, temperature, solar exposure, contamination, artifact health and maturity, storage and reserve, opening fraction, and selected measured material properties. Arithmetic operations include constants, copy, addition, subtraction, multiplication, extrema, and comparisons. Capability scoped actuators request water collection, growth, healing, opening changes, contamination removal, or signal emission.

There are no jumps, loops, calls, imports, dynamic allocation, network access, file access, or strings interpreted as code. Registers are clipped to $[-4,4]$ and each extensive actuator is capped at 0.05 normalized units per tick; the physics layer may further reduce a request to satisfy local capacity and conservation. Programs execute after environmental advancement on every later simulator tick, including agent-free evaluation ticks.

Canonical instruction content determines a program identifier through SHA-256; names, authors, and asserted ancestry cannot spoof identity. When forking is enabled, an agent can use only a program it authored, observed, was taught, or inherited, and the child must change at least one instruction. The registry retains exact parent and child identifiers, authors, installation events, and instruction diffs. A measured skill becomes verified only after inspection records the resulting artifact service. Teaching can transfer this evidence record but cannot create feedstock or empirical material grounding.

\subsection{Interaction channels and recorded provenance}

The full system supports local broadcast or addressed messages, append-only publication, teaching, trade, task claims, shared-depot use, design composition, program reuse, and program forking. Messages and evidence receive stable identifiers; reply and fulfillment identifiers connect requests to later successful actions. Incoming messages wait until the recipient's next fixed macroturn and therefore do not purchase extra model calls.

Epistemic and physical contribution are tracked separately. Material batches retain their recipe, tests, contributing agents, feedstock masses, and causal evidence. Artifacts retain their creator, physical contributors, batch ancestry, specification, and complete program history. Publications and programs retain authorship and parentage. Citing another agent cannot fabricate matter, and co-location alone is not counted as intellectual or physical contribution.

Physical stigmergy does not require a symbolic channel. Harvesting and deposition change resource availability; artifacts remain visible and may continue to alter fields; and later agents can encounter constructions, local gradients, stored matter, damage, or services created by earlier actions. Experimental ablations remove capabilities from both the advertised model schema and the executable engine contract rather than merely asking the model not to use them.

\subsection{Experimental conditions}
\label{sec:si-experimental-design}

The four conditions are defined in Table~\ref{tab:conditions}. The no-communication and no-explicit-culture treatments are distinct: the former retains physically observable executable programs and their possible descent, whereas the latter removes explicit program, skill, authored-text, and mutation-parent access while retaining evidence-labelled physical phenotype and environmental consequences.

\begin{table}[t]
\centering
\caption{Mechanism-resolved experimental conditions.}
\label{tab:conditions}
\begin{tabular}{p{0.21\linewidth}p{0.70\linewidth}}
\toprule
Condition & Available interaction substrate \\
\midrule
Full culture & One shared world with physical stigmergy, messages, publications,
teaching, trade, task claims, shared records, program reuse, and program forking. \\
No communication & One shared world with physical stigmergy. Messaging,
publication, teaching, trade, task claims, and publication-dependent composition are
removed; physically observable program reuse and forking remain available. \\
No explicit culture & One shared world with physical stigmergy. Communication,
program forking, cross-agent sequence inheritance, the skill library, authored
artifact text, and mutation-parent access are removed from the treatment interface. \\
Independent search & $N$ isolated one-agent copies of the seeded world. Member $i$
matches shared agent $i$ in initial position and macroturn phase. Each endpoint is the
maximum over all $N$ members. \\
\bottomrule
\end{tabular}
\end{table}

Independent search is an endpoint-wise envelope, not the trajectory of one selected agent. The isolated winner may differ among discovery AUC, final artifact performance, portfolio resilience, invention count, and held-out evaluation, and may also change between temporal checkpoints. Every isolated member receives the same fixed decision opportunities as its corresponding shared-world agent. Explicit model-call or action budgets, when configured, are partitioned across members. The primary BioFoundry studies used unlimited aggregate budgets and matched scheduled opportunities, whereas the Protein Realms pilot used an 800-call budget per condition, partitioned across the 50 isolated members in independent search. Equal decision opportunity does not imply equal token use because cultural context can lengthen prompts, so token consumption is reported separately.

\subsection{Completed 800-tick population study}
\label{sec:800tick-design}

The primary population study crossed four conditions with $N\in\{50,100,200\}$ and discovery seeds 3201--3204, giving 48 condition--population--seed episodes arranged in matched seed blocks. Discovery lasted 800 ticks. With a macroturn interval of 50, each agent received 16 scheduled decisions, and corresponding shared and isolated members had identical phases. Population size changed density within the fixed $72\times54$ world. All cells used the same model, prompt, strict action schema, configuration hash, engine revision 9, and held-out disturbance seeds 9201--9208.

After discovery, the complete final state was frozen and copied eight times. Agents were removed and each copy advanced for 288 physics ticks under one unseen schedule that changed the location and order of drought, contamination, damage, and resource variation. No model requests or agent actions occurred, but deterministic field dynamics and installed artifact programs continued. Averaging the eight schedules produced one held-out observation for each discovery seed. Artifact knockouts and other portfolio assays likewise used copies and could not feed information back into discovery.

The matrix was assembled from eight recorded study invocations. Their manifests identify engine revision 9 and commits \texttt{83b9e7a}, \texttt{ab68bab}, \texttt{76fa204}, \texttt{a13caee}, and \texttt{738cef6}; all pooled cells have identical configuration, prompt, and action-schema hashes. Three extension manifests record a dirty worktree, which is retained as a provenance caveat together with source-file digests. One $N=200$ full-culture seed had a documented infrastructure deviation. A predeclared rule that did not inspect outcomes replaced only that record with a clean rerun while preserving both manifests. The completed matrix required 89,617 provider calls. Seventeen retryable attempts failed and were repeated without advancing world time (0.019\%). Automated audits pass matched positions, phases, decision opportunities, and engine revision in all 12 population--seed blocks.

\subsection{Completed 3,200-tick long-horizon study}
\label{sec:long-horizon-design}

The long-horizon study was analyzed separately and was not pooled with the 800-tick matrix. It contains 12 episodes in four matched seed blocks: full culture, no explicit culture, and independent search at $N=100$ for discovery seeds 3301--3304. Every episode ran for 3,200 discovery ticks and provided 6,400 scheduled model decisions. Frozen, agent-free copies were evaluated at predeclared ticks 400, 800, 1,600, 2,400, and 3,200 under the same eight held-out seeds 9201--9208. Checkpoint assays performed deterministic local simulation only and did not alter the continuing discovery world. For independent search, every isolated member was assayed and the best-of-100 member was selected separately for each endpoint and checkpoint.

All four matched seed blocks pass the design audit for positions, macroturn phases, decision opportunities, and engine revision. The frozen long-horizon source manifest records the study-summary hash, trace hashes for all shared-world runs, checkpoint tables, and every derived source used in the journal figures. This study tests when explicit cultural channels change technological outcomes; its longer horizon and new seed block do not constitute additional replicates for the population-scaling study.

\subsection{Endpoints and frozen evaluation}
\label{sec:si-analysis}

Discovery-frontier AUC is the time-normalized trapezoidal area under the immutable running maximum of simulator-measured artifact performance. The last frontier value is extended to the declared horizon, so the endpoint rewards both early discovery and sustained improvement. Best final-state artifact performance is reported separately because replacing a controller can lower current function without erasing a historical discovery.

At each agent-free evaluation tick, active artifacts produce a service vector. The portfolio coverage for a service is the maximum current service among active artifacts. Portfolio resilience multiplies mean service coverage by $0.5+0.5b$, where $b$ is the minimum-to-mean coverage ratio, thereby rewarding both functional magnitude and balance without directly rewarding artifact count. Held-out resilience AUC is the mean of this balanced current-service measure over 288 evaluation ticks and then over the eight schedules.

Validated inventions satisfy all predeclared requirements: a tested recipe above the material-utility threshold; a nonempty name, claimed function, architecture, bio-inspiration, and predicted effects; an agent-authored program; lifetime peak artifact performance above threshold; and behavioral novelty above threshold. Supporting mechanism endpoints include multi-agent construction, cross-agent program forking, lineage depth, verified adoption, causal closure, material and program provenance, ecological knockout effects, spatial exploration, artifact proximity, and action allocation. These measurements diagnose how a result arose and do not replace the functional endpoints.

\subsection{Behavioral, lineage, and network analyses}

Behavioral organization was measured after simulation rather than assigned to agents. The 800-tick analysis robust-scaled 15 trajectory and activity features and fit the frozen clustering model without condition, population, seed, or agent-identity labels. The long-horizon physical phenotype analysis used nine movement, coverage, artifact-proximity, artifact-directed-motion, and nearby-agent-exposure features; communication, cultural actions, technology-work rates, and treatment labels were excluded. Principal components and UMAP~\cite{mcinnes_umap_2018} were display methods only; clustering and prediction used the full standardized feature spaces. Candidate cluster counts were compared with the silhouette coefficient~\cite{rousseeuw1987silhouettes}. Agent-level points are descriptive, while treatment comparisons use seed-level phenotype fractions.

Temporal role analysis divided shared-world trajectories into nonoverlapping 200-tick windows. A 13-feature, episode-balanced model was fit without study, condition, population, seed, identity, or time labels, and a separate physical/task-only sensitivity excluded cultural features. Role transitions describe recurring activity states rather than permanent or engine-assigned occupations.

Agent--artifact networks were reconstructed from recorded observation, parent reference, construction, contribution, program installation, repair, dismantling, message, teaching, trade, and executable-descent events. Community assignments used Louvain optimization~\cite{blondel2008fast} on the full weighted networks; simplified backbones were used only for visualization. Agent coordination and artifact co-use projections retained overlaps exceeding a degree-conditioned hypergeometric null after Benjamini--Hochberg correction~\cite{benjamini1995controlling} at $q\leq0.05$ and required at least two shared neighbors. Participation and within-module z-score were interpreted using role-cartography thresholds only as heuristic reference lines~\cite{guimera2005cartography}; binary nestedness used NODF~\cite{almeida-neto_consistent_2008}. Dynamic analyses used 100-tick event intervals and maximum-overlap community matching across checkpoints. Diffusion timing accounted for unreused artifacts with Kaplan--Meier estimates~\cite{kaplan1958nonparametric}, and direct creator--adopter motifs were compared with 200 fixed-seed timestamp shuffles. Structural robustness compared degree- or betweenness-targeted removal with 64 deterministic random removal orders per seed. Structural accessibility after removal is not interpreted as physical function, adaptation, or recovery.

\subsection{Statistical analysis}

The independently generated world seed is the unit of inference. Agents, ticks, artifacts, program forks, behavioral windows, network edges, and held-out schedules are nested observations and are never counted as independent replicates. Conditions are compared only within the same population and matched seed. Each held-out schedule set is averaged before seed-level inference.

Means and 95\% intervals use 20,000 deterministic bootstrap resamples of the four paired seed values. The pseudorandom stream is derived from a stable statistic key so adding another endpoint cannot change an existing interval. Long-horizon paired tables additionally report exact two-sided sign-flip tests. With four nonzero pairs, the smallest attainable two-sided value is 0.125; inference therefore emphasizes effect magnitude, paired consistency, and mechanism rather than dichotomous significance. No agent-level test, multiplicity-adjusted confirmatory claim, or universal scaling law is asserted.

\subsection{Reproducibility and data release}
\label{sec:si-reproducibility}

The source repository retains the engine, configurations, replay and analysis tools, frozen figure inputs, and deterministic journal-figure generators. Figure generation is offline: it verifies source SHA-256 hashes and writes an output manifest containing the Python, Matplotlib, NumPy, font, authoring-code, frozen-input, and output hashes. Previously generated technology portraits are frozen inputs; rebuilding figures does not call a language model or image-generation service.

The paper data builder validates the complete 48-episode 800-tick matrix and 12-episode long-horizon matrix before staging a release. It retains exact pre-run manifests, summaries, authoritative compressed traces, all isolated-member traces, derived analyses, figures, and a file-level SHA-256 inventory. Standalone release scripts reproduce an endpoint summary, stream movement trajectories, summarize trace events and actions, and verify every released file. This release boundary distinguishes the engine revision and Git state that generated each run from the later repository revision used to assemble the manuscript.

\subsection{Performance-ranked selection of semantically distinct technologies}
\label{sec:performance-ranked-tech-selection}

Algorithm~\ref{alg:technology-selection} specifies how technologies are ranked and selected across all designs. 

\begin{algorithm}[htbp]
\caption{Performance-ranked selection of semantically distinct technologies}
\label{alg:technology-selection}
\begin{algorithmic}[1]
\Require Technology records $\mathcal{T}$; target count $K$;
maximum similarity $\tau$; cluster cap $M$; embedding model $E$
\Ensure Selected representative set $\mathcal{S}$

\ForAll{$t \in \mathcal{T}$}
    \State Construct semantic text $x_t$ from the recorded name,
    architecture, function, biological inspiration, materials,
    fabrication process, output form, design principles, and controller
    operations
    \State $\mathbf{z}_t \gets
    E(x_t)/\lVert E(x_t)\rVert_2$
    \Comment{normalized semantic embedding}
    \State $p_t \gets$ recorded lifetime-peak simulator performance
\EndFor

\For{$c \in \{3,\ldots,12\}$}
    \State Cluster $\{\mathbf{z}_t\}$ into $c$ groups using
    average-linkage agglomerative clustering with cosine distance
    \State Compute silhouette score $s_c$
\EndFor
\State $c^\star \gets \arg\max_c s_c$
\State Assign each technology its cluster
$g_t \in \{1,\ldots,c^\star\}$

\State Sort $\mathcal{T}$ by decreasing $p_t$, breaking ties by
technology identifier
\State $\mathcal{S} \gets \varnothing$;
$F \gets \varnothing$;
$n_j \gets 0$ for every cluster $j$

\ForAll{$t \in \mathcal{T}$ in sorted order}
    \State $f_t \gets \operatorname{Hash}(
    \operatorname{NormalizeText}(x_t))$
    \If{$f_t \in F$}
        \State \textbf{continue}
        \Comment{exclude exact semantic-text duplicate}
    \EndIf
    \If{$n_{g_t} \geq M$}
        \State \textbf{continue}
        \Comment{enforce cluster diversity}
    \EndIf
    \If{$\mathcal{S} \neq \varnothing$ \textbf{and}
    $\max_{u\in\mathcal{S}}
    \mathbf{z}_t^{\mathsf{T}}\mathbf{z}_u > \tau$}
        \State \textbf{continue}
        \Comment{exclude semantic near-duplicate}
    \EndIf
    \State $\mathcal{S} \gets \mathcal{S} \cup \{t\}$
    \State $F \gets F \cup \{f_t\}$
    \State $n_{g_t} \gets n_{g_t}+1$
    \If{$|\mathcal{S}|=K$}
        \State \textbf{break}
    \EndIf
\EndFor

\If{$|\mathcal{S}|<K$}
    \State \textbf{raise error}
    \Comment{the declared constraints are infeasible}
\EndIf
\State \Return $\mathcal{S}$
\end{algorithmic}
\end{algorithm}

\begin{figure}[htbp]
    \centering
    \begin{minipage}[t]{0.49\textwidth}
        \textsf{\textbf{A}}\par\vspace{-0.8em}
        \centering
        \includegraphics[width=\linewidth]
        {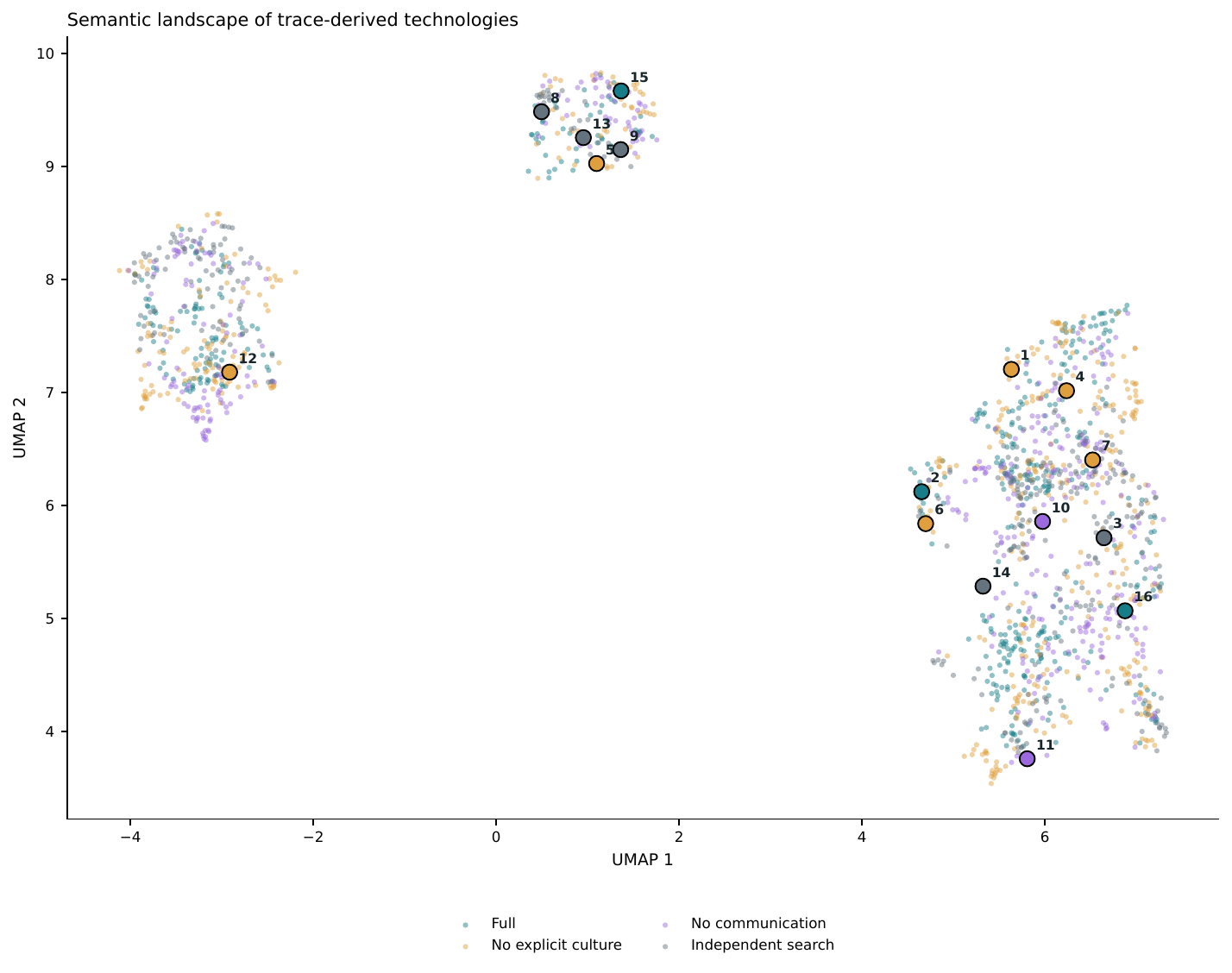}
    \end{minipage}
    \hfill
    \begin{minipage}[t]{0.49\textwidth}
        \textsf{\textbf{B}}\par\vspace{-0.8em}
        \centering
        \includegraphics[width=\linewidth]
        {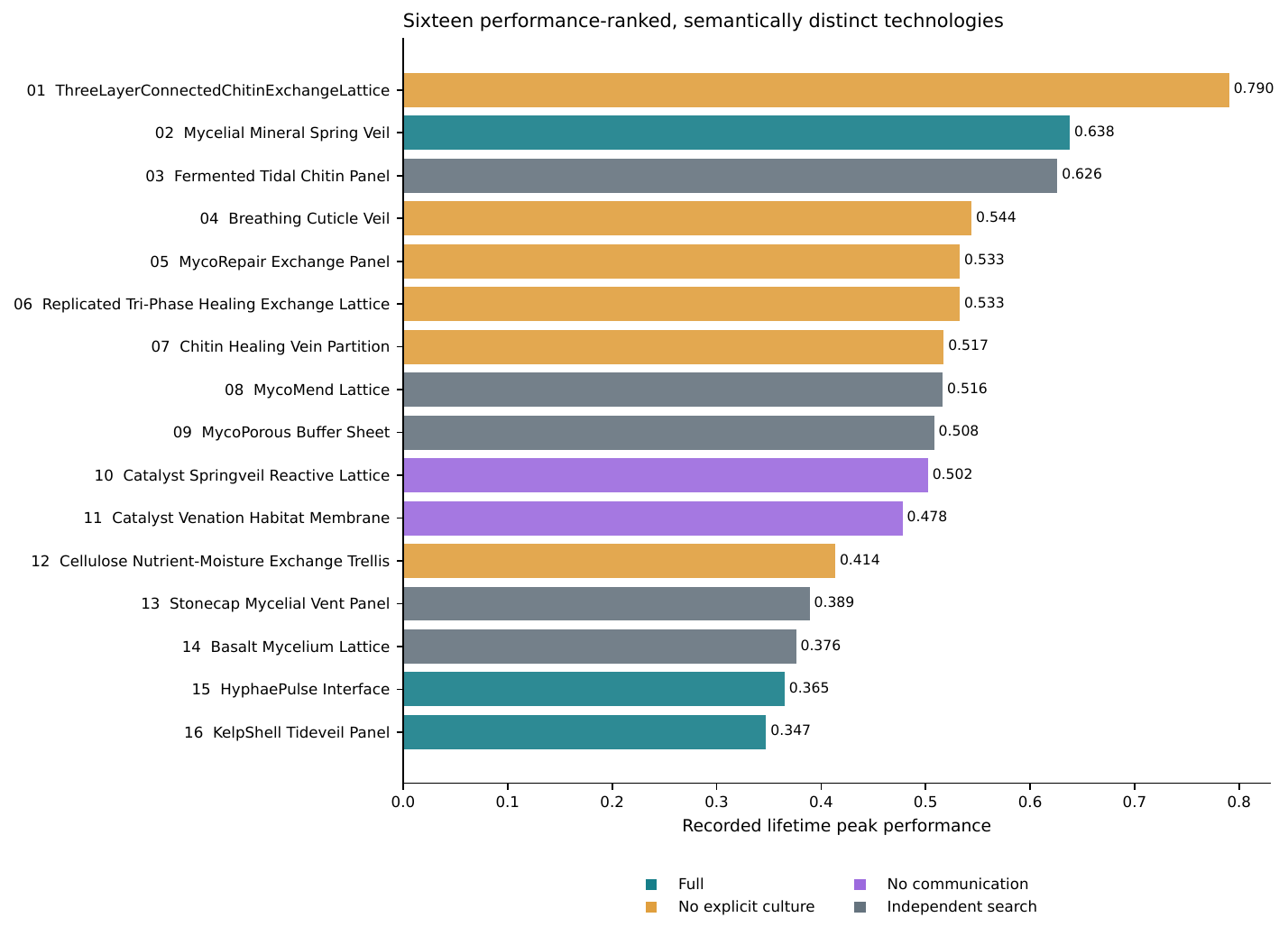}
    \end{minipage}
    \caption[Semantic selection of representative technologies]{
    Semantic selection of representative agent-invented technologies.
    (A) UMAP visualization of 1,718 technologies extracted from 48
    condition--population--seed episodes across the four experimental conditions. Each
    small point represents one complete trace-derived technology and is
    colored by its originating condition. Technology descriptions were
    encoded locally as normalized 768-dimensional vectors using
    \texttt{google/embeddinggemma-300m}. Large outlined points numbered
    1--16 identify the technologies retained for the representative
    gallery in the main text. Nearby points have
    similar recorded architectures, functions, materials, fabrication
    processes, biological inspirations, and controller operations.
    UMAP used cosine distance, 30 neighbors, a minimum distance of 0.12,
    and random seed 42. Its axes have no direct physical meaning, and
    selection was performed in the original embedding space rather than
    in this two-dimensional projection.
    (B) Recorded lifetime-peak simulator performance of the 16 retained
    technologies, ranked from highest to lowest and colored by
    experimental condition. Candidates were considered in descending
    performance order. Exact textual duplicates were removed, no
    selected pair was permitted to have cosine similarity greater than
    0.82, and no more than four technologies could be drawn from any of
    the 11 data-derived semantic clusters. Consequently, the featured
    set includes global performance ranks 1, 2, 3, 14, 16, 17, 22, 24,
    26, 35, 54, 154, 232, 275, 300, and 366. the
    procedure selected strong inventions while preventing the gallery
    from being dominated by many nearly identical versions of the same
    design. Neither the UMAP coordinates nor the generated technology
    illustrations entered simulator scoring or statistical inference. 
    The detailed procedure is given in Algorithm~\ref{alg:technology-selection}.
    }
    \label{fig:technology-semantic-selection}
\end{figure}


\begin{figure}[htbp]
    \centering
    \includegraphics[width=0.97\textwidth,height=0.64\textheight,keepaspectratio]
    {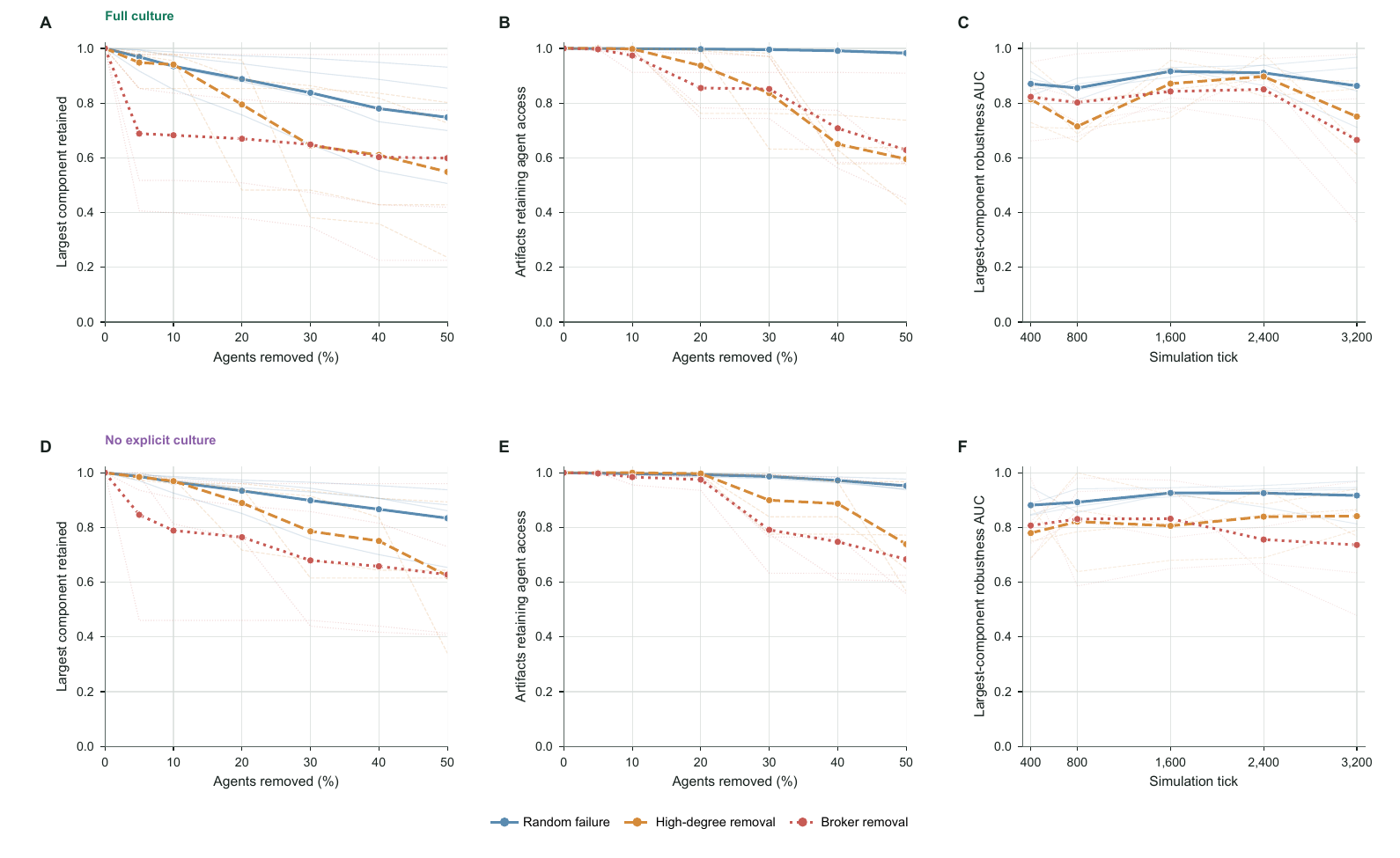}
    \caption[Structural robustness to agent removal]{Structural robustness of evolved agent-artifact networks. Rows compare full culture (A-C) and no explicit culture (D-F). At each checkpoint and seed, agents are removed randomly over 64 deterministic permutations, from highest degree downward, or from highest exact betweenness downward; the graph is not rewired. (A, D) Largest connected component retained relative to the intact network at tick 3,200 as 0 to 50\% of agents are removed. Random loss is gradual, while targeted high-degree and broker removal causes greater fragmentation. (B, E) Fraction of artifacts retaining at least one surviving agent connection. After random 50\% removal, access remains 0.983 under full culture and 0.952 without explicit culture; high-degree removal lowers these values to 0.596 and 0.739, and broker removal to 0.629 and 0.684. (C, F) Largest-component robustness AUC across the full removal curve at five checkpoints. Thin lines are seeds and heavy lines are means. The societies have enough overlapping participation to survive indiscriminate dropout, but a small set of highly connected or bridging agents remains disproportionately important. The assay measures graph structure only, not functional performance or adaptive recovery after knockout.}
    \label{fig:structural-robustness}
\end{figure}


\begin{figure}[htbp]
    \centering
    \includegraphics[width=0.97\textwidth,height=0.80\textheight,keepaspectratio]
    {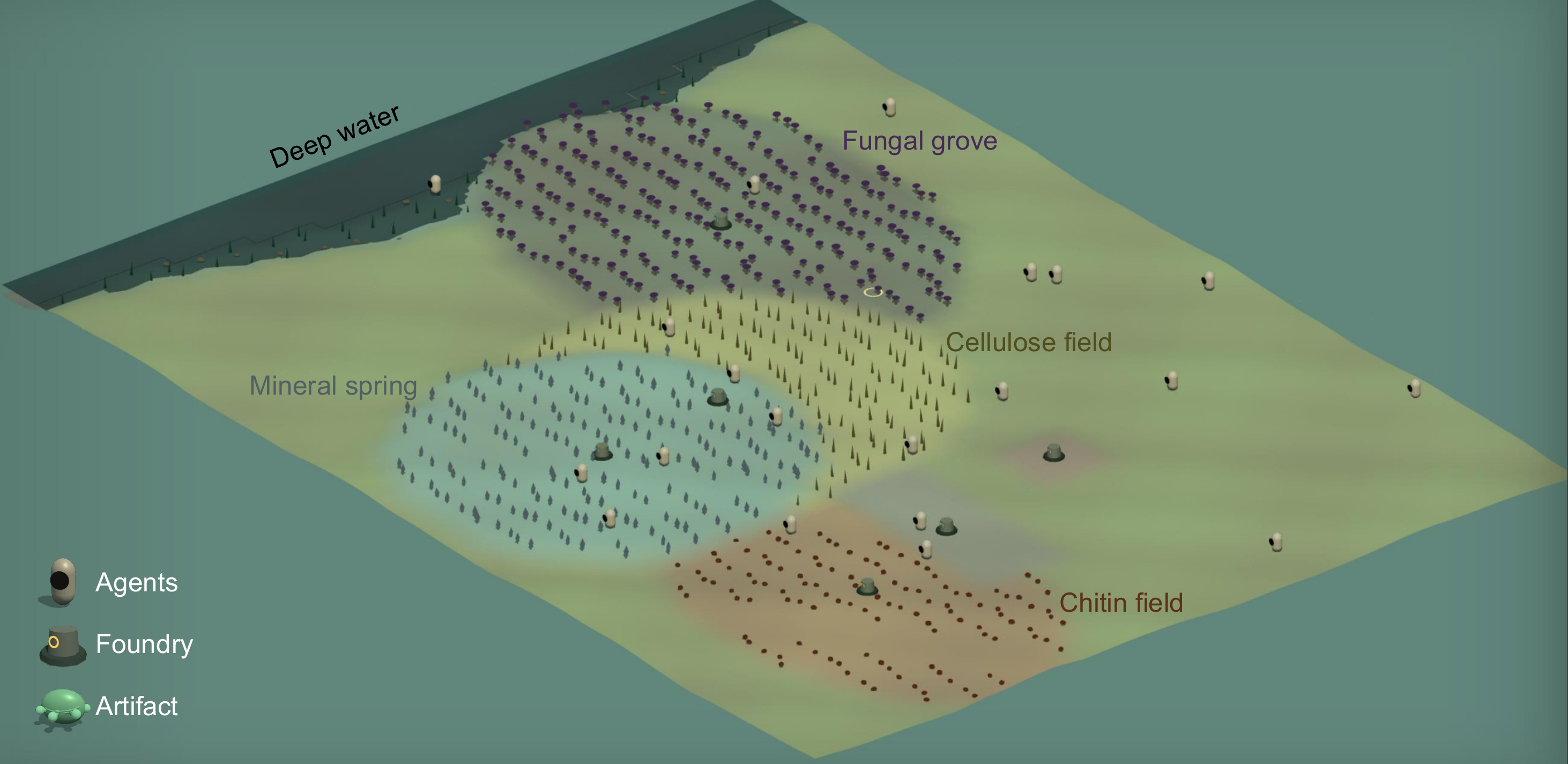}
    \caption[Zoomed version of the field]{Zoomed-in view of a representative SwarmWorld BioFoundry environment. The rendering shows the $72\times54$-cell world for seed 3202, highlighting the spatial substrate through which agents explore, gather resources, fabricate materials, and construct persistent technologies. Distinct environmental regions include deep water and resource biomes such as the fungal grove, cellulose field, mineral spring, and chitin field. Individual LLM agents move through these regions and interact locally with resources and infrastructure, while fixed foundries provide processing locations and agent-created artifacts persist in the environment after construction. The spatial separation of resources, processing sites, agents, and durable artifacts creates localized constraints on discovery and enables indirect coordination through repeated encounters with the modified world.}
    \label{fig:biofoundry_field}
\end{figure}



\newpage

\begin{figure}[htbp]
    \centering
    \includegraphics[width=0.97\textwidth,height=0.80\textheight,keepaspectratio]
    {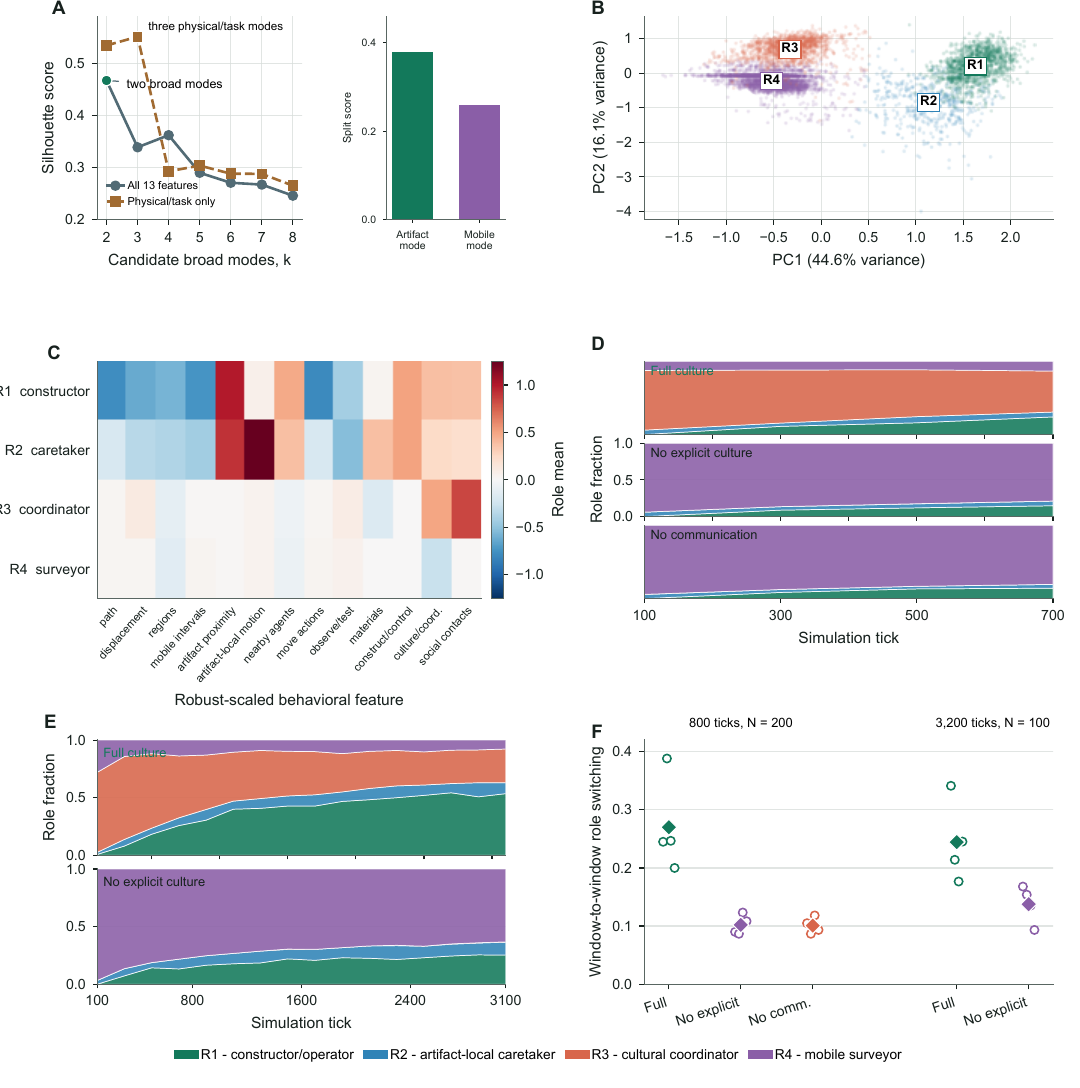}
    \caption[Dynamic behavioral roles and role succession]{Behavioral specialization appears as recurring states and changes over the lifetime of the society. The analysis uses nonoverlapping 200-tick windows from 20 shared-world episodes, giving 22,400 agent-windows, with an episode-balanced sample of 16,000 used for fitting. Thirteen movement, spatial-context, task-action, and cultural-interaction features enter the model; condition, study, population, seed, identity, and time do not. (A) Silhouette audit of candidate broad mode counts. All 13 features select two broad modes with score 0.467, while an anti-circularity fit using only 11 physical/task features selects three modes with score 0.551. The adjacent bars are the accepted two-way conditional split scores inside the artifact-centered and mobile parents, 0.379 and 0.259, with a minimum 5\% child occupancy. (B) PCA display of a deterministic 6,000-window sample colored by final role. PC1 and PC2 explain 44.6\% and 16.1\% of variance; clustering occurs in the original 13-dimensional space. (C) Mean robust-scaled feature signatures. R1 is constructor/operator, combining artifact proximity with construction and control; R2 is artifact-local caretaker, marked by artifact-local motion, processing, and maintenance-context activity; R3 is cultural coordinator, enriched in cultural and social activity; and R4 is mobile surveyor, characterized by movement and observation/testing away from persistent artifacts. (D) Mean state fractions across four N=200 seeds in the 800-tick study. (E) Mean state fractions across four N=100 seeds in the 3,200-tick study. Full-culture societies in both panels reallocate activity toward construction as they mature, and the longer run shows a pronounced coordinator-to-constructor succession. (F) Window-to-window state switching. Open circles are seed means and diamonds are condition means; full culture switches more often in both studies. Leave-replicate-out adjusted Rand indices are 0.999 for the broad modes and 0.921 and 0.980 for their conditional subdivisions. Agents do not acquire fixed assigned professions; they repeatedly enter a small, stable vocabulary of activities as the society changes. The R3 comparison is partly intervention-defined because cultural features are disabled in the ablations, whereas the three-mode physical/task-only sensitivity provides the noncircular evidence for finer differentiation. Agent windows are descriptive, seed is the inferential unit, and the two studies are not a controlled comparison of horizon alone.}
    \label{fig:dynamic-roles}
\end{figure}

\begin{figure}[htbp]
    \centering
    \includegraphics[width=0.97\textwidth,height=0.80\textheight,keepaspectratio]
    {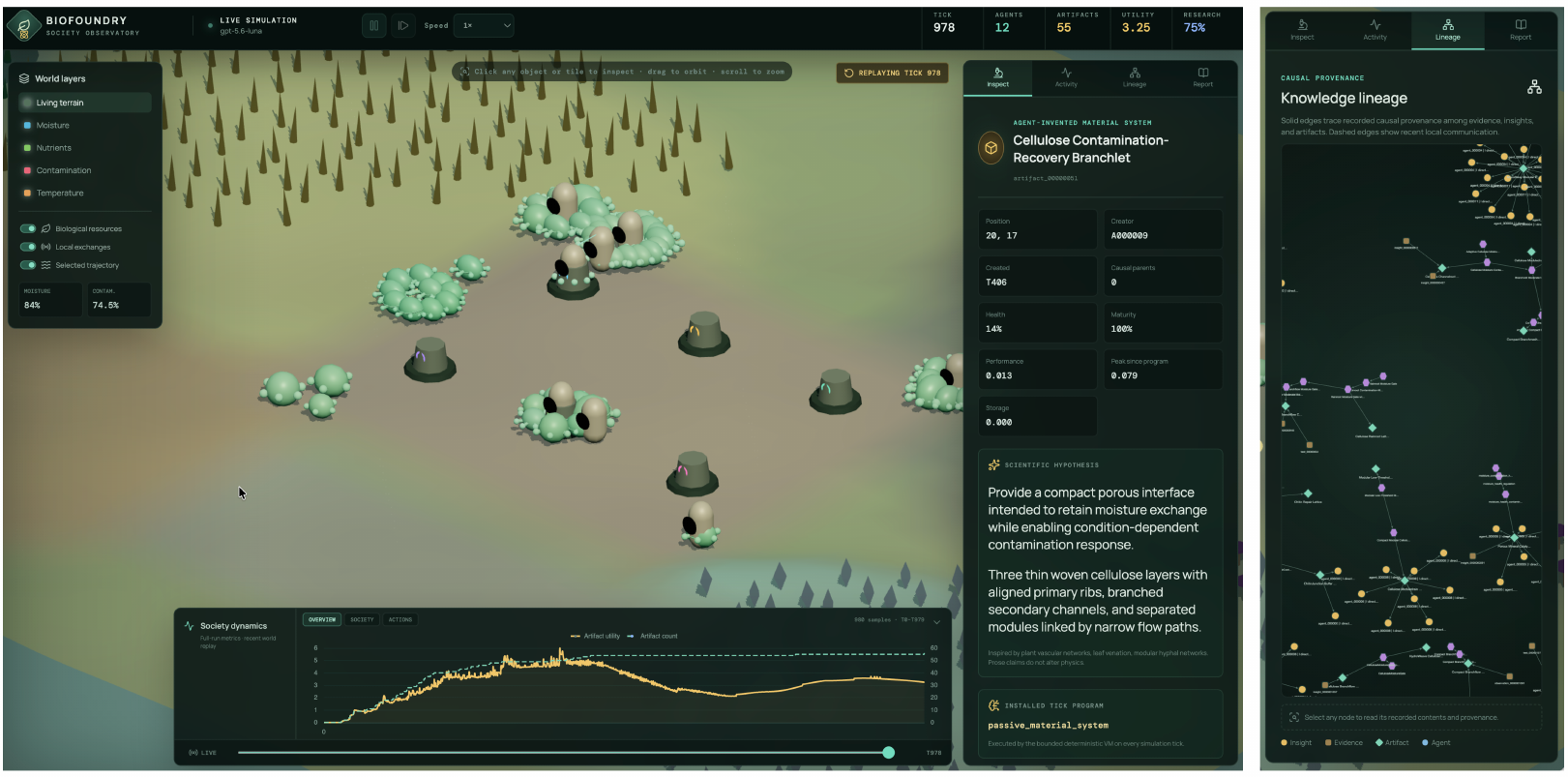}
    \caption{Interactive SwarmWorld interface for visualizing collective technological discovery and knowledge evolution. The central panel provides a three-dimensional view of the shared simulation environment, in which LLM agents move through a spatially heterogeneous world, encounter resources and processing stations, and construct persistent technological artifacts. The interface reports the evolving simulation state, including the current world tick and active agent population, while selecting an agent or artifact exposes its local state, observations, and associated actions in the side panel. Persistent artifacts remain accessible to subsequent agents, allowing earlier discoveries, constructions, and executable controllers to modify the environment encountered by later agents and thereby support stigmergic coordination. Society-level trajectories displayed along the lower panel track the temporal development of collective activity and technological accumulation. The interface additionally exposes the knowledge-lineage graph, which reconstructs how scientific information propagates through the society by linking agent observations and evidence to authored insights, executable programs, precursor artifacts, and downstream technologies. Typed lineage edges record grounded events including authorship or observation, causal dependence, construction, program installation, and cross-agent program forking, allowing complex inventions to be traced back to the agents, evidence, and reusable technological components from which they emerged. In this way, the interface visualizes not only where agents act, but also how discoveries are accumulated, inherited, recombined, and transformed into persistent technologies over time. All functional outcomes are determined by the underlying deterministic simulator rather than by the visualization or agents' textual claims.}
    \label{fig:biofoundry_ui}
\end{figure}

\newpage

\begin{figure}[]
    \centering
    \includegraphics[width=0.97\textwidth,height=0.80\textheight,keepaspectratio]
    {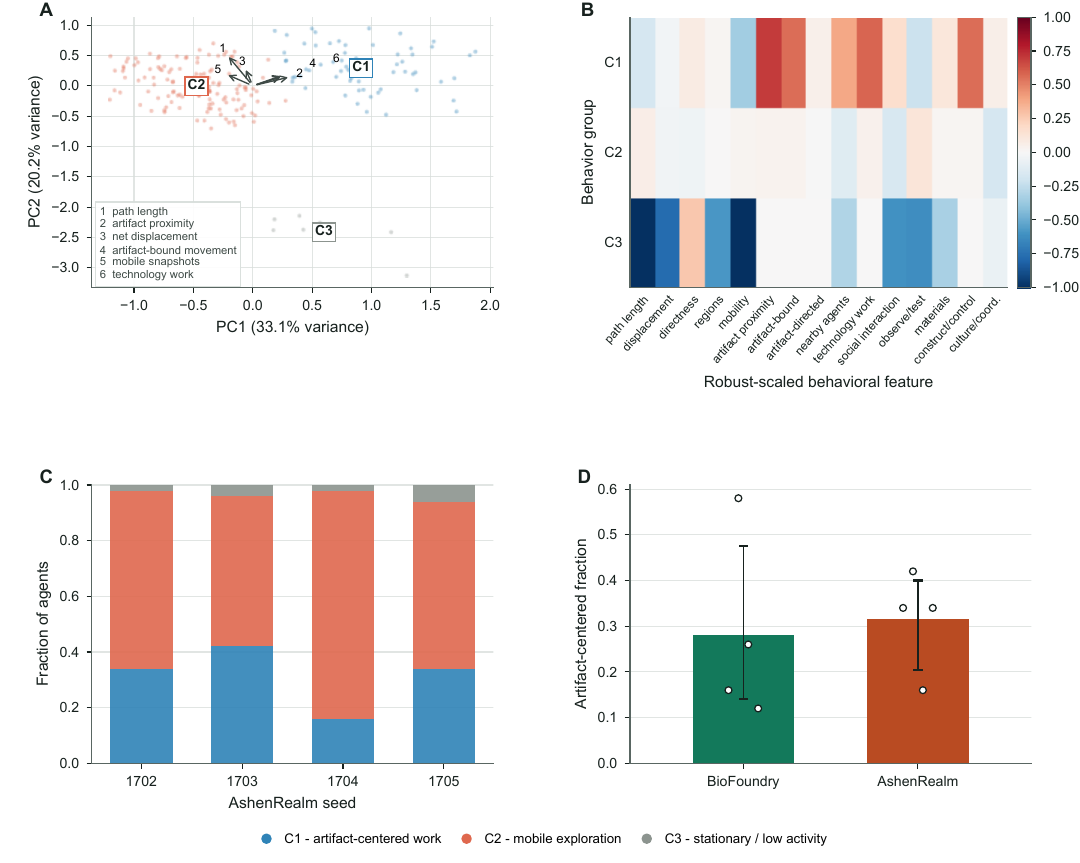}
    \caption{Data-driven behavioral phenotypes in AshenRealm with full communication and N=50. (A) PCA projection of complete agent trajectories represented by 15 robust-scaled behavioral features. PC1 and PC2 explain 33.1\% and 20.2\% of the variance, respectively; numbered vectors indicate the six strongest displayed loading directions, including path length, artifact proximity, net displacement, artifact-bound movement, mobile activity, and technology work. Clustering is performed in the full behavioral feature space rather than in the two-dimensional projection. Each (B) Cluster-average standardized signatures identify C1 as artifact-centered work, C2 as mobile exploration, and C3 as stationary/low activity. (C) Behavioral composition across AshenRealm seeds 1702--1705 shows that all three modes recur across independent worlds. (D) Comparison of the artifact-centered fraction between BioFoundry and AshenRealm. The recurrence of artifact-centered and exploratory behavior under a distinct geography and materials task indicates that behavioral differentiation is not restricted to the original BioFoundry environment.}
    \label{fig:behavior_phenotype_ashen_world}
\end{figure}

\newpage

\begin{figure}[]
    \centering
    \includegraphics[width=0.97\textwidth,height=0.80\textheight,keepaspectratio]
    {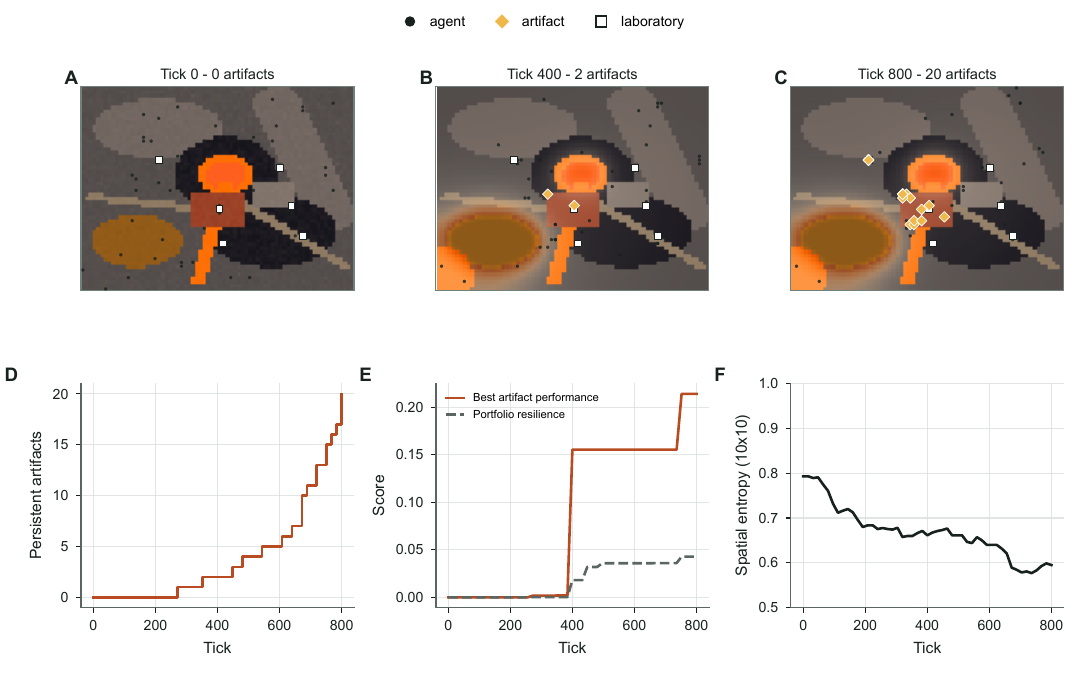}
    \caption{Within-run technological accumulation in a representative AshenRealm society with full communication with N=50 (seed 1703). (A--C) Authoritative world states at ticks 0, 400, and 800 show the progressive modification of the volcanic landscape by persistent technologies. Agents are black circles, artifacts are gold diamonds, and processing laboratories are open squares. The number of constructed artifacts increases from 0 to 2 to 20 as the world changes. (D) Persistent-artifact count through the 800-tick episode. (E) Running best-artifact performance and portfolio resilience. Best-object performance rises through discrete discovery events, whereas portfolio resilience develops as additional technologies accumulate. (F) Normalized spatial entropy of agent occupancy on a fixed $10\times10$ grid declines over the episode, indicating increasing localization of activity as persistent technological sites emerge. This single trajectory is illustrative rather than inferential, but shows how agents progressively transform a materially distinct volcanic environment into a persistent technological landscape.}
    \label{fig:world_evolution_ashen_world}
\end{figure}

\begin{figure}[]
    \centering
    \includegraphics[width=0.97\textwidth,height=0.80\textheight,keepaspectratio]
    {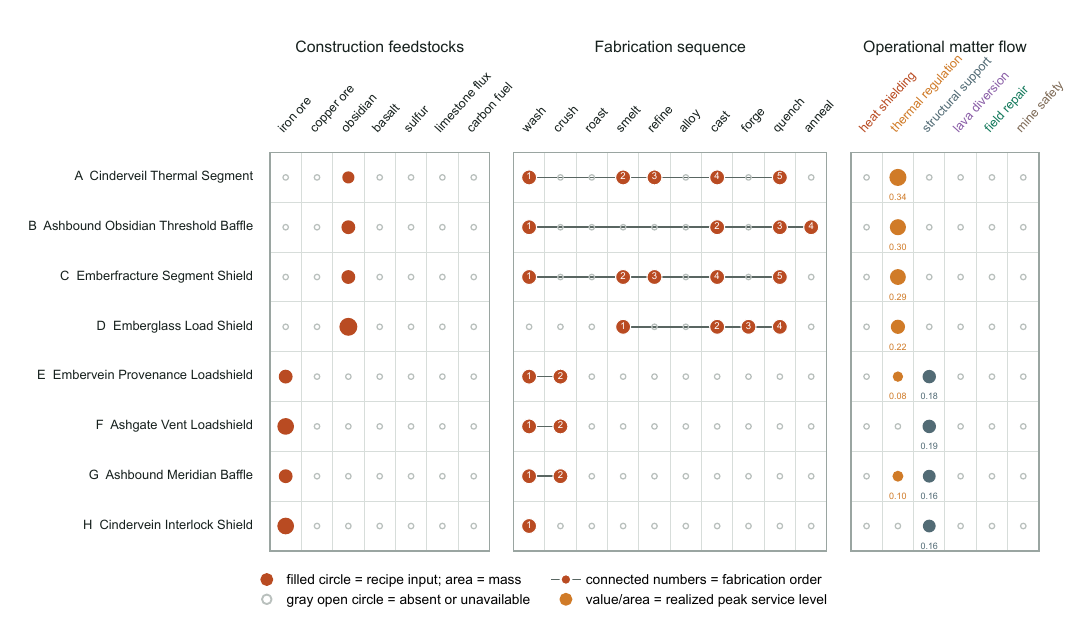}
    \caption{Material inputs, fabrication pathways, and realized functions of representative AshenRealm technologies. Rows A--H correspond to the eight performance-ranked technologies shown in Figure~\ref{fig:ashenrealm_tech}. \textbf{Left:} construction feedstocks recorded when each artifact was built. Filled circles identify recipe inputs and circle area scales with consumed mass; gray open circles indicate absent or unavailable feedstocks. The four highest-ranked designs (A--D) are predominantly obsidian-based, whereas E--H use iron ore. \textbf{Center:} ordered fabrication pathways. Filled numbered nodes identify executed processing operations and connecting lines show their sequence; open circles mark operations not used. Obsidian technologies generally employ longer multistep routes involving washing, smelting, refining, casting, forging, quenching, or annealing, while the iron-based designs use shorter wash--crush pathways. \textbf{Right:} realized lifetime-peak functional service recorded by the simulator. Circle area and printed values encode the realized service magnitude; open circles indicate no realized service in that dimension. Obsidian designs are dominated by thermal regulation, reaching 0.34 for the Cinderveil Thermal Segment, whereas the iron-based technologies primarily provide structural support, with several designs also contributing thermal regulation. The figure separates what matter agents actually consumed, how it was processed, and what functions the resulting artifacts actually realized, preventing agent-authored design claims from being conflated with executed simulator behavior.}
    \label{fig:matter_pathway_ashen_world}
\end{figure}

\newpage

\section{Agent communication and simulator-validated consequences}
\label{sec:si-agent-communication}

This section exposes the model-to-world interface at the level of individual messages and actions. Text inside the rounded agent bubbles is reproduced verbatim from the retained language-model trace; only line wrapping and the display abbreviations A000, A013, and so forth are introduced. A000 denotes \texttt{agent\_000000}. Pale green and blue bubbles distinguish conversational participants, while pale orange bubbles contain verbatim natural-language intent attached to a structured \texttt{TEACH} or \texttt{TRADE} action. Gray strips are compact human-readable decodings of authoritative simulator events. They are deliberately separated from agent speech because a message is a claim, whereas a simulator record states what was delivered, measured, transferred, installed, or constructed.

All examples come from the representative full-culture, $N=100$, seed-3301 trajectory run for 3,200 ticks. The complete trace contains 2,914 delivered messages, 40 formal teaching events, 52 physical resource trades, 457 artifact-program installations, and 389 constructed artifacts. The episodes below were selected to illustrate distinct communication mechanisms rather than to estimate their frequency or average effect.

\begin{center}
\footnotesize
\setlength{\tabcolsep}{5pt}
\renewcommand{\arraystretch}{1.15}
\begin{tabular}{@{}>{\ttfamily\raggedright\arraybackslash}p{0.17\linewidth}>{\raggedright\arraybackslash}p{0.30\linewidth}>{\raggedright\arraybackslash}p{0.43\linewidth}@{}}
\toprule
Display label & Agent-side object & Authoritative consequence \\
\midrule
MESSAGE / REPLY & Natural-language payload generated inside a structured \texttt{COMMUNICATE} action & A \texttt{message\_delivered} event records sender, recipients, location, message identifier, and any replied-to message. \\
TEACH & Structured transfer containing a natural-language explanation and cited record or program identifiers & A \texttt{knowledge\_taught} event adds a durable teaching record to the addressed agent and can fulfill a tracked request. \\
TRADE & Structured addressed action specifying a resource and amount & A \texttt{resource\_traded} event changes the two grounded inventories transactionally. \\
SIMULATOR RECORD & Not agent dialogue & Accepted world event used to distinguish an executed consequence from an agent's unverified statement. \\
\bottomrule
\end{tabular}
\end{center}

\FloatBarrier
\subsection{Vignette S1: local discovery and independent confirmation}

\begin{swchatleft}{Tick 6 - A000 to A013 - MESSAGE - \texttt{message\_0000000057}}
``I am surveying nearby fabrication workspaces and cellulose material behavior. Have you observed a station, tested recipe, or useful local landmark to share?''
\end{swchatleft}

\begin{swworldrecord}{ticks 6 and 37}
The message is delivered within the local communication radius. At tick 37, A013 harvests 0.30 mass of resource enum 5 (CELLULOSE) at $(45,15)$ and marks the request as fulfilled.
\end{swworldrecord}

\begin{swchatright}{Tick 38 - A013 to A000 - REPLY to \texttt{message\_0000000057}}
``I directly observed CELLULOSE at [45,15], approximately 2.6 mass across five nearby cellulose-field sites. No fabrication station is present locally; I am exploring for one.''
\end{swchatright}

\begin{swworldrecord}{tick 39}
A013's subsequent \texttt{INSPECT} creates \texttt{observation\_0000000587}: terrain \texttt{CELLULOSE\_FIELD}, resource \texttt{CELLULOSE}, remaining resource mass 2.303663, station \texttt{NONE}, at $(45,15)$. The measurement is stored independently of the prose message.
\end{swworldrecord}

\Needspace{0.20\textheight}
\begin{swchatleft}{Tick 51 - A000 to A013 - REPLY to \texttt{message\_0000000568}}
``I have confirmed CELLULOSE at [45,16] and harvested a sample. Have you found a verified fabrication station, landmark, or tested cellulose behavior? Please share coordinates or evidence if available.''
\end{swchatleft}

\begin{swworldrecord}{tick 50}
Immediately before sending the reply, A000 successfully harvests 0.30 CELLULOSE at $(45,16)$. The follow-up message therefore reports an independently executed local action rather than simply repeating A013's claim.
\end{swworldrecord}

\begin{swinterpretation}
The conversation turns a local observation into a navigational lead, but the receiving agent still verifies and harvests the material for itself. The sequence illustrates grounded information sharing rather than unrestricted copying of another agent's belief.
\end{swinterpretation}

\subsection{Vignette S2: measurement exchange, executable inheritance, and replication}

\begin{swchatleft}{Tick 277 - A075 to A011 - REPLY - \texttt{message\_0000006055}}
``My independent CHITIN microbatch was fabricated and TESTed: utility 0.475047, quality 0.6766, permeability 0.1838, toughness 0.214246, healing 0.23336, responsiveness 0.15782. I am now attempting a small persistent modular lattice build at the confirmed FERMENTER.''
\end{swchatleft}

\begin{swchatright}{Tick 291 - A011 to A075 - REPLY to \texttt{message\_0000006055}}
``I independently measured my 0.3-mass chitin batch at utility 0.470496, quality 0.6588, permeability 0.184889, toughness 0.214246, healing 0.23048, and responsiveness 0.14912. I am using this evidence to build a distinct moisture-gated porous membrane at the confirmed FERMENTER.''
\end{swchatright}

\begin{swworldrecord}{ticks 289 and 324}
A011 constructs \texttt{artifact\_00000003}, ``Moisture-Gated Chitin Membrane,'' containing program \texttt{program\_88ab6e937138ab56c8b8c12f}. At tick 324, A027 constructs \texttt{artifact\_00000006}, ``Moisture-Gated Chitin Interface,'' deploying the same content-addressed program. The controller senses local moisture, compares it with 0.35, and actuates aperture state.
\end{swworldrecord}

\begin{swworldrecord}{tick 325}
A075 executes \texttt{FORK\_PROGRAM} on \texttt{artifact\_00000006}. The accepted child, \texttt{program\_bc99a75d96d63e1f35dcbbdc}, records \texttt{program\_88ab6e937138ab56c8b8c12f} as its parent and changes the moisture threshold from 0.35 to 0.50.
\end{swworldrecord}

\begin{swchatleft}{Tick 327 - A075 to A011 - MESSAGE - \texttt{message\_0000007110}}
``I am testing a fork of the observed moisture-gated chitin program on artifact\_00000006, raising the aperture threshold from 0.35 to 0.5. Local moisture is 0.435; I will report inspected service changes without claiming unmeasured field effects.''
\end{swchatleft}

\begin{swworldrecord}{tick 370}
A030 constructs \texttt{artifact\_00000009}, ``Adaptive Chitin Exchange Interface,'' with independent material evidence and the original parent program. Its causal parents include the previously observed \texttt{artifact\_00000006}.
\end{swworldrecord}

\begin{swchatleft}{Tick 376 - A075 to A011 - MESSAGE - \texttt{message\_0000008097}}
``The locally observed high-threshold fork on artifact\_00000006 currently reports performance 0.0969, adaptive regulation 0.0986, and structural support 0.0926; the parent-program artifact\_00000009 reports performance 0.0851, adaptive regulation 0.0515, and structural support 0.0982. This is an observed program comparison under current local conditions, not proof of broader field performance.''
\end{swchatleft}

\begin{swchatright}{Tick 390 - A011 to A075 - REPLY to \texttt{message\_0000008097}}
``I inspected the locally observed high-threshold fork comparison. The fork on artifact\_00000006 reports performance 0.0969, adaptive regulation 0.0986, and structural support 0.0926; the parent-program artifact\_00000009 reports performance 0.0851, adaptive regulation 0.0515, and structural support 0.0982. This is an observed local comparison, not proof of broader field performance.''
\end{swchatright}

\begin{swinterpretation}
The agents exchange measurements, encounter a persistent controller in the world, modify it through an explicit parent-child program edge, and compare the descendant with a parent-program artifact. The caveats in the raw messages are important: the agents describe matched local measurements without treating one comparison as universal proof. This is a compact example of executable cultural evolution rather than language-only brainstorming.
\end{swinterpretation}

\subsection{Vignette S3: a distributed request becomes durable teaching}

\begin{swchatleft}{Tick 747 - A005 to 18 nearby agents - MESSAGE - \texttt{message\_0000015114}}
``I published direct evidence for my MINERAL-CATALYST microbatch and local contextual-versus-passive field observations. Please use the new records as causal parents and report only matched measurements, especially under elevated contamination.''
\end{swchatleft}

\begin{swchatright}{Tick 749 - A002 to A005 - REPLY to \texttt{message\_0000015114}}
``I am performing a fresh direct inspection of artifact\_00000029 under the current mineral-spring condition and will share the matched measurement record; existing evidence indicates contextual remediation is conditional and near zero at contamination 0.012.''
\end{swchatright}

\begin{swchatright}{Tick 793 - A008 to A005 - REPLY to \texttt{message\_0000015114}}
``My independent current inspection of artifact\_00000029 running program\_3494cd0a9e51636ce656c8cd at the mineral spring measured performance 0.0347, structural support 0.0481, remediation 0.0, self-maintenance 0.0, health 0.7981, and contamination 0.012. This is matched local evidence only; causal record insight\_0000015016.''
\end{swchatright}

\begin{swchataction}{Tick 834 - A017 to A005 - TEACH action - reply to \texttt{message\_0000015114}}
``Direct comparison from my controlled tests: test\_00000080 cellulose-mineral-catalyst utility 0.417955, toughness 0.145664, permeability 0.142931, healing 0.06355; test\_00000052 mineral-catalyst control utility 0.426554, toughness 0.066569, permeability 0.153, healing 0.074146. I have built artifact\_00000040 from the cellulose-containing batch and will inspect it for matched current field services; these results do not establish field effects.''
\end{swchataction}

\begin{swworldrecord}{tick 834}
The \texttt{TEACH} action succeeds with cited causal parents \texttt{test\_00000080} and \texttt{test\_00000052}. The simulator creates \texttt{teaching\_0000016669} in A005's durable records and marks the original request as fulfilled.
\end{swworldrecord}

\begin{swinterpretation}
One agent makes an open research request, several peers undertake local follow-up, and another agent transfers a controlled comparison through the stronger \texttt{TEACH} channel. Unlike an ordinary message, teaching creates a durable record with explicit evidence identifiers, making the informational handoff auditable and retrievable later.
\end{swinterpretation}

\subsection{Vignette S4: a conversational request becomes a physical transaction}

\begin{swchatleft}{Tick 1691 - A010 to A006 - MESSAGE - \texttt{message\_0000031166}}
``I need a small grounded CHITIN transfer for a persistent composite build. Can you trade approximately 0.20 CHITIN at the TESTER? My direct composite evidence is insight\_0000030344.''
\end{swchatleft}

\begin{swchataction}{Tick 1694 - A006 to A010 - TRADE action - reply to \texttt{message\_0000031166}}
``Transferring 0.20 grounded CHITIN for your persistent composite build; please return direct build or field evidence when available.''
\end{swchataction}

\begin{swworldrecord}{tick 1694}
The transaction succeeds: 0.20 CHITIN is removed from A006's grounded inventory and added to A010's inventory. The simulator creates \texttt{fulfillment\_0000031213} and marks the request as fulfilled.
\end{swworldrecord}

\begin{swworldrecord}{tick 1742}
Forty-eight ticks later, A010 constructs \texttt{artifact\_00000138}, ``Fungal-Chitin Healing Exchange Interface,'' at the TESTER. The accepted recipe consumes 0.40 CHITIN and 0.15 FUNGUS and records a three-layer aligned porous composite with a passive installed program.
\end{swworldrecord}

\begin{swinterpretation}
The episode shows that communication can invoke a simulator-enforced transfer of physical matter rather than merely exchanging text. The later build is temporally consistent with the request and uses the requested material. However, its declared causal parents are prior experimental records rather than the trade message, so this vignette should be interpreted as a transaction followed by construction, not as definitive proof that the transferred 0.20 mass alone caused the artifact.
\end{swinterpretation}

\subsection{Raw structured output corresponding to a displayed message}

The following is the complete accepted action object stored for A000 at tick 6. Numeric enum \texttt{verb: 11} denotes \texttt{COMMUNICATE}; the zero-valued direction, resource, and artifact fields are the required inactive defaults. The text in the \texttt{message} field is the text rendered in the first chat bubble above.

\begin{lstlisting}[style=swjson]
{
  "amount": 0.0,
  "artifact": 0,
  "artifact_spec": null,
  "causal_parents": [],
  "direction": 0,
  "insight": null,
  "message": "I am surveying nearby fabrication workspaces and cellulose material behavior. Have you observed a station, tested recipe, or useful local landmark to share?",
  "program": null,
  "recipe": null,
  "reply_to": "",
  "resource": 0,
  "target_agent_id": "agent_000013",
  "target_artifact_id": "",
  "target_x": -1,
  "target_y": -1,
  "verb": 11
}
\end{lstlisting}

\subsection{Trace provenance and interpretation limits}

The source is generated with engine revision 9 and model \texttt{gpt-5.6-luna}. Displayed message text, action intent, identifiers, coordinates, values, ticks, and recipient relationships are taken from the retained trace. Simulator-record strips condense the corresponding event payloads into readable prose but do not introduce additional outcomes. These examples establish that the mechanism can support grounded communication, executable inheritance, durable teaching, and transactional exchange. They are illustrative process evidence; comparative claims about prevalence or performance remain based on the paired seed-level analyses in the main Results.

\end{document}